\pdfoutput=1

\documentclass[11pt]{article}

\usepackage{EMNLP2023}

\usepackage{times}
\usepackage{latexsym}
\usepackage[T1]{fontenc}
\usepackage[utf8]{inputenc}
\usepackage{microtype}
\usepackage{inconsolata}

\usepackage{xcolor}

\newcommand{\myflushright}[1]{%
  \unskip\hspace*{1em plus 1fill}%
  \nolinebreak[3]\hspace*{\fill}\mbox{\upshape #1}
}
\usepackage{pifont}

\usepackage{hyperref}       
\usepackage{url}            
\usepackage{booktabs}       
\usepackage{amsfonts}       
\usepackage{nicefrac}       
\usepackage{xcolor}         
\usepackage{amsmath}
\usepackage{soul}
\usepackage{caption}
\usepackage{algpseudocode}
\usepackage{algorithm}
\usepackage{tikz}
\usepackage{pgfplots}
\usepackage{graphicx}
\usepackage{subcaption}
\usepackage{cleveref}
\usepackage{multirow}
\usepackage{amssymb}

\usepackage{nccmath}

\definecolor{codegreen}{rgb}{0,0.6,0}
\definecolor{codegray}{rgb}{0.5,0.5,0.5}
\definecolor{codepurple}{rgb}{0.58,0,0.82}
\definecolor{backcolour}{rgb}{0.95,0.95,0.92}

\definecolor{czredd}{HTML}{720026}
\definecolor{czred}{HTML}{D90429}
\definecolor{czredl}{HTML}{EF9595}
\definecolor{czredl}{HTML}{F4998D}
\definecolor{czoranged}{HTML}{CC5803}
\definecolor{czorange}{HTML}{FF8600}
\definecolor{czorangel}{HTML}{FFD89C}
\definecolor{czyellowd}{HTML}{FDC43F}
\definecolor{czyellow}{HTML}{FBE134}
\definecolor{czyellowl}{HTML}{FCEFB4}
\definecolor{czgreend}{HTML}{003E1F}
\definecolor{czgreen}{HTML}{08A045}
\definecolor{czgreenl}{HTML}{B5E48C}
\definecolor{czblued}{HTML}{01497C}
\definecolor{czblue}{HTML}{0466c8}
\definecolor{czbluel}{HTML}{78C1F3}
\definecolor{czpurpled}{HTML}{4A148C}
\definecolor{czpurple}{HTML}{7B2CBF}
\definecolor{czpurplel}{HTML}{B1B2FF}
\definecolor{czpinkd}{HTML}{EA638C}
\definecolor{czpink}{HTML}{FF8FA3}
\definecolor{czpinkl}{HTML}{FFCCD5}
\definecolor{czbrownd}{HTML}{774936}
\definecolor{czbrown}{HTML}{9D6B53}
\definecolor{czbrownl}{HTML}{DEAB90}
\definecolor{czgray}{HTML}{D1D1D1}
\definecolor{czgrayl}{HTML}{EEEEEE}

\newcommand\capped[1]{\textcolor{czblue}{\boldsymbol{#1}}}
\newcommand\varied[1]{\textcolor{czred}{\boldsymbol{#1}}}

\newcommand\good[1]{\textcolor{codegreen}{\bf #1}}

\title{Revisiting Block-based Quantisation:\\What is Important for Sub-8-bit LLM Inference?}

\author{Cheng Zhang$^1$, Jianyi Cheng$^1$, Ilia Shumailov$^2$, George A. Constantinides$^1$, Yiren Zhao$^1$ \\
  $^1$Imperial College London, $^2$University of Oxford \\
  \texttt{\{cheng.zhang122, jianyi.cheng17, g.constantinides, a.zhao\}@imperial.ac.uk} \\
  \texttt{ilia.shumailov@chch.ox.ac.uk}
}

\begin{document}

\maketitle

\begin{abstract}

The inference of Large language models (LLMs) requires immense computation and memory resources. To curtail these costs, quantisation has emerged as a promising solution, but existing LLM quantisation mainly focuses on 8-bit. In this work, we explore the statistical and learning properties of the LLM layer and attribute the bottleneck of LLM quantisation to \emph{numerical scaling offsets}. To address this, we adapt block quantisations for LLMs, a family of methods that share scaling factors across packed numbers. Block quantisations efficiently reduce the numerical scaling offsets solely from an arithmetic perspective, without additional treatments in the computational path. Our nearly-lossless quantised 6-bit LLMs achieve a $19\times$ higher arithmetic density and $5\times$ memory density than the \texttt{float32} baseline, surpassing the prior art 8-bit quantisation by $2.5\times$ in arithmetic density and $1.2\times$ in memory density, without requiring any data calibration or re-training. We also share our insights into sub-8-bit LLM quantisation, including the mismatch between activation and weight distributions, optimal fine-tuning strategies, and a lower quantisation granularity inherent in the statistical properties of LLMs. The latter two tricks enable nearly-lossless 4-bit LLMs on downstream tasks. Our code is open-sourced \footnote{\url{https://github.com/ChengZhang-98/llm-mixed-q}}.

\end{abstract}

\section{Introduction}
Pre-trained Large Language Models (LLMs)~\cite{brown2020language,gpt-neo,zhang2022opt} have demonstrated impressive performance on a range of Natural Language Processing (NLP) tasks. However, their underlying computational and memory costs are a critical bottleneck to their usability. For instance, the larger variants in the GPT family scale up to hundreds of billions of parameters, requiring at least 300GB of memory to store these parameters in a float16 format~\cite{brown2020language}. Quantisation serves as a natural solution for reducing the cost of running inference on these LLMs
~\cite{yao2022zeroquant,xiao2022smoothquant,dettmers2022llm},
as a low-precision format enables cost savings across all relevant efficiency metrics: reduced on-chip memory, increased arithmetic intensity for matrix multiplies, and decreased DRAM bandwidth requirement.
On the other hand, the growing popularity of running services such as ChatGPT \cite{chatgpt}
provides an impetus for exploring the use of custom silicon to support LLM inference. This raises the question: \textit{What would a low-precision number system look like in these near-future LLM hardware accelerators (ASICs)?}

LLM quantisation is challenging because of the activations with large absolute magnitudes, also known as activation outliers~\cite{bondarenko2021understanding, xiao2022smoothquant}. Previous approaches have proposed various techniques to address such outliers. However, these either require additional treatments in the integer quantisation domain (\texttt{LLM.int8()} and SmoothQuant) or yield unsatisfactory performance (ZeroQuant); and prior work has primarily focused on arithmetics that can be ported to GPUs. We observe that the presence of outliers necessitates different scaling factors at a finer granularity than per-tensor or per-token level
~\cite{yao2022zeroquant,xiao2022smoothquant}. This insight naturally leads us to revisit arithmetic systems with small exponents, such as MiniFloat
~\cite{sun2019hybrid}, Block Minifloat~\cite{fox2021block}, Block Logarithm ~\cite{miyashita2016convolutional}, and Block Floating Point
~\cite{kalliojarvi1996roundoff}, as they can effectively represent outliers in Transformer models. To the best of our knowledge, our work is the first to systemically investigate short-exponent arithmetics for LLM quantisation.



\begin{table*}[!t]
\vskip 0.15in
\begin{center}
\begin{small}
\begin{sc}
\begin{tabular}{lcccc}
\toprule
Method   & (QW, QAct)  & Bitwidth  & PTQ or TAQ & $\#$ Quantised GEMMs\\
\midrule
ZeroQuant \cite{yao2022zeroquant}
& ($\surd$, $\surd$)
& W4A8
& TAQ
& 8/8 \\

LLM.int8() \cite{dettmers2022llm}
& ($\surd$, $\surd$)
& W8A8$^*$
& PTQ
& 6/8 \\

GPTQ \cite{frantar2022gptq}
& ($\surd$, $\times$)
& W4
& PTQ + DC
& 6/8 \\

SmoothQuant \cite{xiao2022smoothquant}
& ($\surd$, $\surd$)
& W8A8
& PTQ + DC
& 6/8 \\
Ours
& ($\surd$, $\surd$)
& W6A6/W4A4
& PTQ/TAQ
& 8/8 \\
\bottomrule
\end{tabular}
\end{sc}
\end{small}
\end{center}
\vskip -0.1in
\caption{A comparison of different LLM quantisation methods. (QW, QAct) shows whether quantisations are applied to weights or activations, W$x$A$y$ means $x$-bit quantisation for weights and $y$-bit quantisation for activation. PTQ and TAQ represents Post Training Quantisation and Training After Quantisation respectively. DC means data calibration. There are eight general matrix multiplications (GEMMs) per transformer layer (\textcircled{\small{1}}-\textcircled{\small{8}} in \Cref{alg:transformer}). Only ZeroQuant and ours quantise all of them. Other approaches leave \textcircled{\small{4}} and \textcircled{\small{5}} in \texttt{float32}/\texttt{float16} format, which take up 20.6\% floating-point operations in OPT-6.7B's self-attention. $^*$ means outliers in \texttt{LLM.INT8()} is computed in \texttt{float16}; this improves arithmetic density but memory density is kept identical to canonical \texttt{float16}.
}
\label{tab:introduction:comparison}
\end{table*}

\begin{figure*}[!t]
\centering
\begin{minipage}[b]{.4\linewidth}
    \input{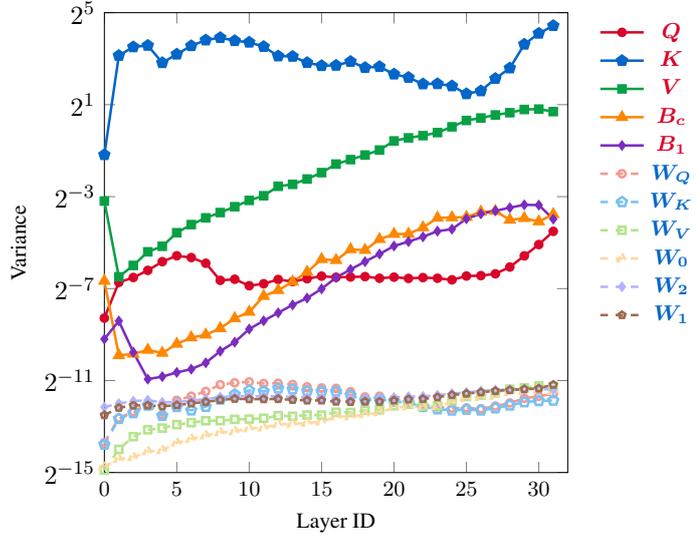}

\end{algorithm}
\end{minipage}
\begin{minipage}[b]{.55\linewidth}
    \begin{tikzpicture}[thick,scale=0.95, every node/.style={scale=0.95}]
\pgfplotsset{every x tick label/.append style={font=\footnotesize}, compat=1.3}
\begin{semilogyaxis}[
    height=80mm,
    width=80mm,
    xlabel={\footnotesize Layer ID},
    ylabel={\footnotesize Variance},
    ylabel style = {align=center},
    xmin=0, xmax=32,
    ymax = 32, ymin = 3.0517578125e-05,
    xtick={0, 5, 10, 15, 20, 25, 30},
    ymode=log, log ticks with fixed point,
    log basis y={2},
    ytick={0.0000306, 0.00048828125, 0.0078125, 0.125, 2, 32},
    yticklabels={$2^{-15}$, $2^{-11}$, $2^{-7}$, $2^{-3}$, $2^{1}$, $2^5$},
    scaled x ticks = false,
    scaled y ticks = false,
    legend columns = 1,
    legend style={at={(1.05,1)},anchor=north west, draw=none, fill=none, font=\footnotesize},
    ]

\addplot[czred, draw=czred, mark=*, line width=1pt, mark options={scale=0.7}] table
[x=layer, y=q, col sep=space] {opt6.7b.dat};
\addlegendentry{$\varied{Q}$}

\addplot[czblue, draw=czblue, mark=pentagon*, line width=1pt] table
[x=layer, y=k, col sep=space] {opt6.7b.dat};
\addlegendentry{$\varied{K}$}

\addplot[czgreen, draw=czgreen, mark=square*, line width=1pt, mark options={scale=0.7}] table
[x=layer, y=v, col sep=space] {opt6.7b.dat};
\addlegendentry{$\varied{V}$}

\addplot[czorange, draw=czorange, mark=triangle*, line width=1pt] table
[x=layer, y=b0, col sep=space] {opt6.7b.dat};
\addlegendentry{$\varied{B_c}$}

\addplot[czpurple, draw=czpurple, mark=diamond*, line width=1pt, mark options={scale=0.7}] table
[x=layer, y=b1, col sep=space] {opt6.7b.dat};
\addlegendentry{$\varied{B_1}$}


\addplot[dashed, czredl, draw=czredl, mark=o, line width=1pt, mark options={solid, scale=0.7}] table
[x=layer, y=wq, col sep=space] {opt6.7b.dat};
\addlegendentry{$\capped{W_Q}$}

\addplot[dashed, czbluel, draw=czbluel, mark=pentagon, line width=1pt, mark options={solid}] table
[x=layer, y=wk, col sep=space] {opt6.7b.dat};
\addlegendentry{$\capped{W_K}$}

\addplot[dashed, czgreenl, draw=czgreenl, mark=square, line width=1pt, mark options={solid, scale=0.7}] table
[x=layer, y=wv, col sep=space] {opt6.7b.dat};
\addlegendentry{$\capped{W_V}$}

\addplot[dashed, czorangel, draw=czorangel, mark=triangle, line width=1pt] table
[x=layer, y=w0, col sep=space] {opt6.7b.dat};
\addlegendentry{$\capped{W_0}$}

\addplot[dashed, czpurplel, draw=czpurplel, mark=diamond, line width=1pt, mark options={solid, scale=0.7}] table
[x=layer, y=w1, col sep=space] {opt6.7b.dat};
\addlegendentry{$\capped{W_2}$}

\addplot[dashed, czbrown, draw=czbrown, mark=pentagon, line width=1pt, mark options={solid, scale=0.7}] table
[x=layer, y=w2, col sep=space] {opt6.7b.dat};
\addlegendentry{$\capped{W_1}$}

\end{semilogyaxis}
\end{tikzpicture}
\end{minipage}
\caption{The algorithm on the left is the forward pass computation of a single Transformer layer \cite{vaswani2017attention} in mainstream LLMs, wherein values in \textcolor{czblue}{blue} (\textit{e.g.}~\textcolor{czblue}{$X_n$}) represent tensors with predetermined min-max values, such as the outputs of a normalisation layer or softmax. Values in \textcolor{czred}{red} have unbounded min-max, and are plotted on the upper right for different layers of OPT-6.7B \cite{zhang2022opt}.
We show that for almost all activation tensors, their variances increase at deeper layers, resulting in \emph{scaling offsets} in their quantisation, while weight tensors on the lower right have smaller variances. This statistical trend enlightens our LLM quantisation study. }
\label{fig:introduction:motivation}
\end{figure*}

\Cref{fig:introduction:motivation} illustrates the variance
of the tensors joining the GEMMs in an OPT-6.7B \cite{zhang2022opt}. After feeding 128 samples from Wikitext2 to the pretrained \texttt{float32} model, we make three interesting observations. 1) The variance of most activations in \Cref{fig:introduction:motivation} increases with the depth of the layer; 2) Certain tensors (\textit{e.g.} $\mathbf{K}$) consistently have a greater variance compared to others; 3) All the weight variance is smaller than activations. {Similar trends can be observed in other LLMs. We provide a variance plot of Vicuna-7B~\cite{zheng2023judging} in Appendix (\Cref{fig:appendix:motivation_vicuna}).}

The presence of varying numerical ranges across layers and tensors poses a challenge to the efficacy of a single quantisation configuration for the entire network. From an arithmetic perspective, we refer to this phenomenon as \emph{numerical scaling offsets}, as it requires different numerical ranges and granularities for quantisation. To ensure optimal performance, these layers should be subjected to fine-grained non-linear quantisation strategies.

\Cref{tab:introduction:comparison} provides a comparison between our work and existing LLM quantisation methods. Our quantisation considers all GEMMs (8/8) in transformer layers and both Post-Training-Quantisation (PTQ) and Training-After-Quatisation (TAQ) scenarios. In this work, we also explore suitable places to perform TAQ and quantisation search within the entire NLP pipeline.
We make the following contributions:
\begin{itemize}
    \item We address the LLM quantisation problem with activation outliers and examine it as a \emph{scaling offsets} problem from an arithmetic design perspective. We demonstrate the efficacy of a family of arithmetic systems with short exponents shared across a block of numbers.
    \item We propose a novel quantisation framework based on block arithmetic, and demonstrate its effectiveness in performing W6A6 inference for various tasks. Our nearly-lossless W6A6 outperforms prior work in terms of arithmetic density and memory density, without requiring data calibration or fine-tuning.
    \item We present two methods to achieve 4-bit quantisation on downstream tasks: one is fine-tuning-based, and the other is mixed-precision search. The latter further  demonstrates the potential advantage of shifting LLM inference to cost-effective ASICs.
\end{itemize}

\section{Related Work}

While quantisation of earlier Machine learning (ML) models has been extensively studied, effective quantisation of LLMs still remains an
open problem.
In this section, we review the previous works on block-based
quantisation and compare to the existing
LLM quantisation techniques.

\subsection{Block-based Quantisation}

Block-based quantisation is a technique that quantises a block of values into a compact format, where the elements within each block share common digits. This technique offers a significant memory footprint reduction while maintaining a minor round-off error.
A number of previous works rely on this method to  quantise Convolutional Neural Networks (CNNs). Lin \textit{et al.} utilised a linear combination of multiple binary bases,
equivalent to each binary matrix having a scaling factor~\cite{lin2017accurate}.
Subsequently, Zhang~\textit{et al.} introduced LQ-Nets that rely on a form of block quantisation with a shared scaling factor at the vector level~\cite{zhang2018lqnets}.
Further investigations explored grouping numbers at various granularities, including layer-wise~\cite{wu2018training}, channel-wise~\cite{krishnamoorthi2018quantizing}, and vector-wise quantisation~\cite{dai2021vs}.

It is worth noting that sharing a scaling factor is similar to, but not necessarily the same as, sharing the exponent \cite{darvish2020pushing}. This distinction arises because scaling factors can be arbitrary \texttt{float32} values, whereas exponent values must be integers represented by the assigned number of bits. Our work focuses on sharing the exponent or exponent bias.
When the block size of the shared exponent is 1, we fall back to the minifloat representation such as \texttt{FP8} \cite{sun2019hybrid}. These approaches showed promising results primarily for vision models or relatively small Transformer-based models, while we shift the focus to quantising LLMs with a significantly larger parameter count.

\subsection{LLM Quantisation}

Efficient quantisation techniques for language models have been explored in previous works.
Zafrir \textit{et al.} proposed an approach for quantising BERT~\cite{shen2019qbert} into 8-bit integers~\cite{Zafrir_2019}, while
Shen \textit{et al.}~\cite{shen2019qbert} proposed Hessian-based
ultra-low precision quantisation for the same model.
Zhang \textit{et al.}~\cite{zhang2020ternarybert} quantised BERT to ternary values leveraging layer-wise
knowledge distillation, and
Bai \textit{et al.}~\cite{bai2021binarybert} further pushed the quantisation of BERT weights
to binary values.

The recent surge of interest in quantising LLMs has presented a unique challenge distinct from the prior art summarised above. This challenge stems from the increased model sizes of LLMs.
Yao \textit{et al.} proposed ZeroQuant, which quantises both weights and activations of large transformers into small integers with shared scaling factors~\cite{yao2022zeroquant}. However, as mentioned by \citet{xiao2022smoothquant}, ZeroQuant suffers from a severe accuracy loss. Dettmers \textit{et al.} introduced {\tt LLM.int8()}, a method that computes outlier GEMMs in \texttt{float16} and the rest in 8-bit integer~\cite{dettmers2022llm}. Xiao \textit{et al.} extended 8-bit LLM quantisation with their PTQ technique named SmoothQuant,
Xiao \textit{et al.} proposed SmoothQuant
which scales down activations by row and scales up weights by column proportionally before 8-bit fixed-point quantisation~\cite{xiao2022smoothquant}. Frantar \textit{et al.}
proposed GPTQ, which quantises the weights of LLMs to 3 or 4-bit integers while keeping the activations in \texttt{float32}. Most LLM quantisation methods, directly or indirectly, reserve LLM activation outliers.
\begin{figure*}[t]
    \centering
    \includegraphics[width=1\textwidth]{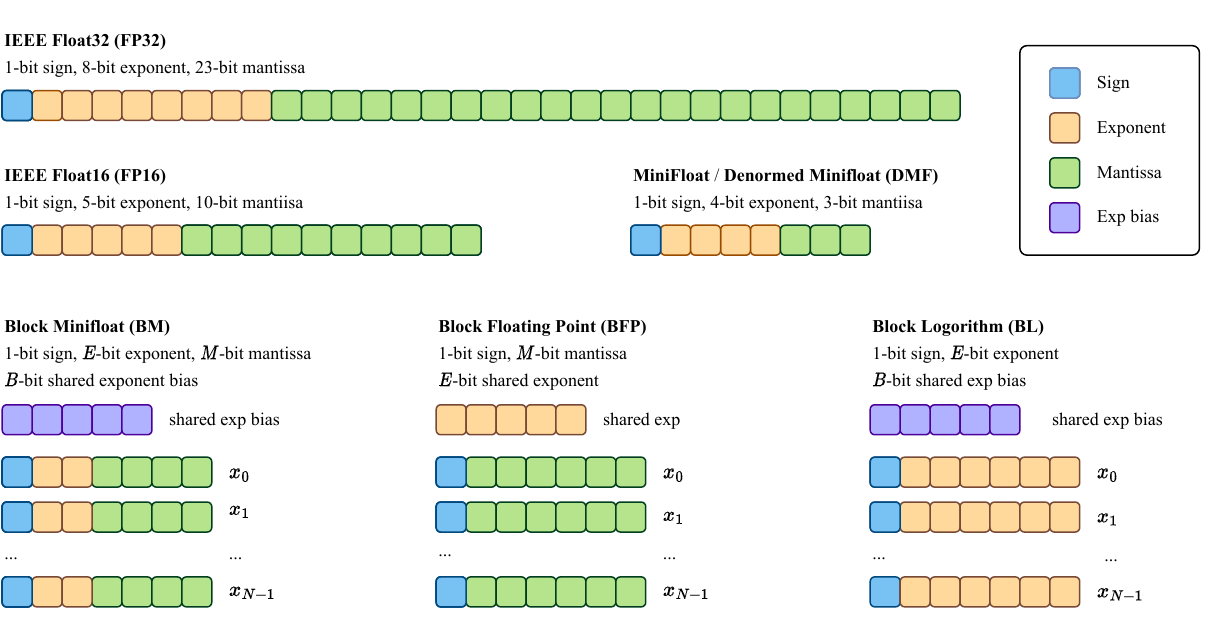}
    \caption{An illustration of different quantisation methods considered in this work: MiniFloat~\cite{sun2019hybrid} and Denormed MiniFloat (DMF), Block MiniFloat (BM)~\cite{fox2021block}, Block Floating-Point (BFP)~\cite{darvish2020pushing} and Block Logarithm (BL).
    }
    \label{fig:method:arithmetics}
\end{figure*}

\section{Method}

In this section, we outline our quantisation strategy for LLMs. We first define block-based quantisation and then describe the metrics we use for evaluating quantisation methods. Finally, we detail a precision search that lowers the quantisation granularity down to the tensor level, effectively accommodating the statistical distribution inherent in LLMs.

\subsection{Block-based Arithmetic}

\Cref{fig:method:arithmetics} illustrates the data representation we explore to address LLM quantisation as well as the standard \texttt{float32}/\texttt{float16}. We outline the specifications for traditional floating-point numbers and extend them to block-based quantisation. Detailed definitions can be found in \Cref{appendix:quant_arith}.

\paragraph{Standard floating-point}
A standard IEEE floating-point number is defined as a 4-tuple, $(s, e, m, b)$~\cite{kahan1996ieee}. $s \in \{0, 1\}$ is the sign bit, $e \in \mathbb{N}$ is the exponent field; $b \in \mathbb{N}$ is the exponent bias; and $m \in \mathbb{N}$ is the mantissa. Let the bit widths of the exponent and the mantissa be $E$ and $M$, respectively. The IEEE standard \texttt{float32} (FP32) number has $E=8$ and $M=23$, where the other bit is used as the sign bit.
Note that the exponent bias depends on $E$: $b = 2^{E-1}-1$, separating the exponent field symmetrically. Similarly, \texttt{float16} (FP16) has $E=5$ and $M=10$.

\paragraph{MiniFloat and Denormalised MiniFloat}
MiniFloat is an efficient floating-point representation that requires fewer bits than traditional floating-point numbers. Traditionally, an 8-bit MiniFloat inherits the definition of FP32 by assigning $E=4$ and $M=3$. We saturate MiniFloat when $e=2^E-1$ and thus no $\pm \inf$ is included.

In this paper, we also introduce a Denormalised MiniFloat (DMF) with zero as the implicit leading bit in the mantissa.
Similar to MiniFloat, we saturate the infinity to a maximum finite
value. DMF provides a higher precision than MiniFloat for small values at the expense of narrowing down the value range. We investigate this trade-off in the context of quantising LLMs.

\paragraph{Block MiniFloat, Block Floating-Point and Block Logarithm}
As shown in \Cref{fig:method:arithmetics}, Block quantisation packs values in a block in which a common scaling factor is shared across $N$ values where $N$ is the block size, reducing the computation in vector inner products. This work mainly explores three block quantisation arithmetics on LLMs: BM, BFP and BL.

Block Minifloat (BM) shares a $B$-bit exponent bias~\cite{fox2021block}. This representation achieves high precision and high range at the same time, at the cost of a larger quantisation error at medium value than standard floating point. This is potentially amenable to values in a multimodal distribution,
where values close to a peak can be efficiently represented in a block.
Block Floating-Point (BFP) shares an $E$-bit exponent. This shared exponent
bounds the range in the block and is amenable to values with small block variances.
Block Logarithm (BL) sets the mantissa in BM to 1 and shares a $B$-bit exponent bias, resulting in values that are powers-of-twos. This contrasts with BFP and is amenable to values with large dynamic ranges.

All these quantisation methods are non-linear and thus can be useful tools to address the \emph{scaling offsets} phenomenon depicted in \Cref{fig:introduction:motivation}. Moreover, the hyper-parameter block size allows for flexible quantisation granularity, ranging from layer-wise, tensor-wise, and channel-wise, to slice-wise (a slice along the token/channel vector).

\subsection{Arithmetic and Memory Densities}

Reducing model size is not the only advantage of quantisation; it also simplifies the computation, thereby accelerating inference. We evaluate quantisation arithmetics using adopted memory and arithmetic densities~\cite{darvish2020pushing}. We define memory density as the reciprocal of the size of the activation and weight data in a model, and the arithmetic density as the reciprocal of the area/the number of Look-Up-Tables (LUTs) to synthesise a multiply-accumulate (MAC) unit, which serves as the basic cell for matrix multiplication in custom inference circuits.
An efficient quantisation method should make a good trade-off among task accuracy, memory density, and arithmetic density. We implemented MAC units with different above-mentioned arithmetics in FPGAs to obtain the number of LUTs.
A detailed description of this procedure can be found in \Cref{appendix:arith_density}.

\subsection{Quantisation Search}
\label{subsec:quantisation_search}

Previous works~\cite{dong2019hawq, habi2020hmq}  observed that the layers in CNNs exhibit varying tolerance, or ``sensitivity'', to quantisation -- we also notice this phenomenon in LLMs. The crucial aspect is identifying the layers that are sensitive and determining tailored quantisation configurations. To achieve this, we apply Tree-structured Parzen Estimator (TPE)~\cite{bergstra2011algorithms} to conduct a fine-grained search for quantisation precision multiple times and analyse the statistics inherent in the quantised models that recover more accuracy.
Our search space is constructed on a per-tensor basis, allowing each input tensor or weight tensor in \textcircled{\small{1}}-\textcircled{\small{8}} (See \Cref{alg:transformer}) to have its own precision. The search space increase exponentially as the layer count increases. We leverage accuracy and memory density to design the objective function: $O_f = acc + \alpha \cdot mem$.
Here $O_f$, $acc$, $mem$ represent the objective function, accuracy, and memory density of the searched quantised models, respectively.
The constant $\alpha$ is used to balance $acc$ and $mem$. To determine the $\alpha$ for a specific search, we initially set $\alpha$ to 1.0 and perform the search while recording the values of $(acc, mem)$ until convergence. The final value of $\alpha$ is determined as $\frac{acc_c}{mem_c}$, where $(acc_c, mem_c)$ represents the converged values. Detailed search parameters are in \Cref{appendix:exp_setup}.

\section{Evaluation}
\label{sec:evaluation}

We conducted a comprehensive set of experiments to identify the key factors influencing the performance of sub-8-bit LLMs. We begin with a language modelling task to eliminate less promising quantisation methods (\Cref{sec:ptq_on_wikitext2}), and then run the promising ones on downstream tasks. For the tasks that proved challenging even for FP32 models, we resort to fine-tuning. Additionally, we conducted a mixed-precision search on two tasks where the quantised 4-bit model struggle. The results of this search provide insights into how to further refine quantisation at the tensor level.

\begin{table}
    \centering
    \begin{tabular}{lcrrr}
\toprule
Method & Config & $E$ & $M$ & $B$ \\
\midrule
Fixed-point & W8A8 & - & 7 & - \\
MiniFloat & W8A8 & 4 & 3 & - \\
DMF  & W8A8 & 4 & 3 & - \\
BFP  & W8A8 & 8 & 7 & - \\
BFP  & W6A6 & 8 & 5 & - \\
BFP  & W4A4 & 8 & 3 & - \\
BM   & W8A8 & 4 & 3 & 8 \\
BL   & W8A8 & 7 & - & 8 \\
\bottomrule
    \end{tabular}
    \caption{The quantisation configuration used in the following sections, where $E$, $M$, and $B$ are the bit-width of exponent (shared exponent), mantissa, and bias (shared bias) respectively. 
    }
    \label{tab:results:quant_config}
\end{table}
\begin{table*}
\centering
\begin{tabular}{lcrrrrrrr}
\toprule
\multicolumn{1}{c}{\multirow{2}{*}{Method}}
& \multicolumn{1}{c}{\multirow{2}{*}{Config}}
& \multicolumn{5}{c}{\multirow{1}{*}{Perplexity ($\downarrow$)}}
& \multicolumn{2}{c}{\textit{Hardware metrics}}
\\
\cmidrule(lr){3-7}
\cmidrule(lr){8-9}
&
& 125M  & 350M  & 1.3B  & 2.7B  & 6.7B
& \multicolumn{1}{c}{Mem $\uparrow$}
& Arith $\uparrow$ \\ \midrule

FP32
& -
& 27.65     & 22.00      & 14.62      &  12.47     & 10.86
& 1$\times$
& 1$\times$
\\ \midrule
\tt{LLM.int8()}
& W8A8$^{\dag}$
& {27.72}     & {22.03}    & 14.64    & {12.49}     & 10.86
& 2$\times$
& $<\text{7.7}\times$
\\
GPTQ
& W4$^{*}$
& 31.12     & 24.24     & 15.47     & 12.87     & 11.39
& $<\text{1.6}\times$
& -
\\
SmoothQuant
& W8A8
& -$^{\ddag}$     & -$^{\ddag}$     & {14.62}     & 12.50     & {10.85}
& {$<\text{4}\times $}
& {$<\text{7.7}\times$}
\\
SmoothQuant-c
& W8A8
& -$^{\ddag}$     & -$^{\ddag}$     & 17.97     & 26.88     & 42.90
& {4$\times $}
& {7.7$\times$}
\\
\midrule
Fixed-point
& W8A8
& 275    & 117
& 1.78E4    & 7.81E3  & 3.77E3
& 4$\times$
& 7.7$\times$
\\
MiniFloat
& W8A8
& {28.16}     & 22.24      & {15.03}      & 12.73  & 10.99
& 4$ \times$
& 17.4$\times$
\\
DMF
& W8A8
& 30.41     & 23.89      & 18.08     &  14.55  & 11.95
& 4$\times$
& 17.4$\times$
\\
BFP
& W6A6
& \good{28.27}    &  \good{22.22}    &  \good{15.08}   &  \good{12.54}    &   \good{10.90}
& \good{4.9$\boldsymbol{\times}$}
&  \good{19.2$\times$}
\\
BFP
& W4A4
&  41.94     &  33.98    &  24.70   &  19.34  & 13.59
& 7.1$\times$
& 37.3$\times$
\\
BM
& W8A8
&  5.6E3    &  2.7E4    &  1.17E4   &  1.33E4  &  8.61E3
&  3.8$\times$
&  14.4$\times$
\\
BL
& W8A8
&  780  &  1.26E3    &  323    &   950   &  289
& 3.8$\times$
& 16.1$\times$
\\

\bottomrule
\end{tabular}
\caption{Perplexity ($\downarrow$) values with zero-shot Post-Training-Quantisation (PTQ) on WikiText2, this means we directly quantise the pre-trained model and apply on WikiText2. Mem and Airth represent Memory and Arithmetic density accordingly. DMF, BM, BFP and BL represent Denormalised MiniFloat, Block Minifloat, Block Floating Point and Block Logarithm respectively. SmoothQuant-c is our improved implementation where the two activation matrix multiplications are now also quantised. $^{\dag}$ means the inliner matrix multiplications are calculated in 8-bit fixed-point, and outliers are calculated in FP16. $^*$ means the weights of GPTQ are kept in FP32. $^{\ddag}$ means SmoothQuant repository does not include the weight scaling matrices for 125M and 350M. We \good{highlight} the best block-based quantisation arithmetic, 6-bit BFP, considering perplexity, memory density, and arithmetic density together.
}
\label{tab:results:opt_ptq_wikitext2}
\end{table*}

\begin{table}
\centering
\resizebox{0.5\textwidth}{!}{%
\begin{tabular}{lccl}
\toprule
Model & FP32 & \texttt{LLM.int8()} & W6A6 BFP \\
\midrule
LLaMA-7B & 5.79 & 5.83 (+0.04) & 5.83 (+0.04) \\
Vicuna-7B & 7.06 & \good{7.07 (+0.01)} & 7.08 (+0.02) \\
Alpaca-7B & 7.01 & 7.02 (+0.01) & 7.02 (+0.01) \\
LLaMA-13B & 5.17 & 5.22 (+0.05) & \good{5.20 (+0.03)} \\
Vicuna-v1.5-13B & 6.13 & 6.16 (+0.03) & 6.16 (+0.03) \\\bottomrule
\end{tabular}
}
\caption{
Perplexity ($\downarrow$) values of LLM family quantized by W6A6 BFP. We compare our method with FP32 and \texttt{LLM.int8()} and find that our method achieves nearly lossless perplexity on Wikitext2. We exclude GPTQ and SmoothQuant-c in this table because they have obvious perplexity increase larger than 0.2 and 5.0 respectively.
}
\label{tab:results:other_llms}
\end{table}

\begin{table*}[t]
\centering
\begin{tabular}{lcrrrrrrr}
\toprule
\multicolumn{1}{c}{\multirow{2}{*}{Method}}
& \multicolumn{1}{c}{\multirow{2}{*}{Config}}
& \multicolumn{5}{c}{\multirow{1}{*}{Mean accuracy ($\uparrow$,\%)}}
\\
\cmidrule(lr){3-7}
\cmidrule(lr){8-9}
&
& 125M  & 350M  & 1.3B  & 2.7B  & 6.7B
\\ \midrule

Float32
& -
& 52.7   & 57.5   & 69.6   & 65.4  & 73.4
\\ \midrule

\texttt{LLM.int8()}
& W8A8
& 52.5 (-0.2)     & 58.3 (+0.8)      & 69.2 (-0.4)     & 65.3 (-0.1)      & 73.5 (+0.1)
\\
\texttt{LLM.int4()}
& W4A4
& 50.8 (-1.9)   & 55.8 (-1.7)   & 67.0 (-2.6)   & 64.5 (-0.9)   & 72.5 (-0.9)
\\
SmoothQuant-c
& W8A8
& -     & -     & 67.2 (-2.4)     & 65.2 (-0.2)      & 72.2 (-1.2)
\\ \midrule
MiniFloat
& W8A8
& 52.1(-0.6)        & 55.1(-2.4)   & 64.7(-4.9)     & 65.7(+0.3)     & 70.5(-2.9)
\\

BFP
& W4A4
& 47.8 (-4.9)   & 51.7 (-5.8)    & 57.2 (-12.4)     & 55.7 (-9.7)     & 67.2 (-6.2)
\\
BFP
& W5A5
& 51.1 (-1.6)   & 56.8 (-0.7)   & 65.5 (-4.1)   & 64.6 (-0.8)    & 72.0 (-1.4)
\\
BFP
& W6A6
& \good{52.6 (-0.1)}       & \good{57.6 (+0.1)}  &  \good{67.8 (-1.8)}     &  \good{65.5 (+0.1)}    &  \good{72.9 (-0.5)}
\\
BFP
& W8A8
& 52.8 (+0.1)    & 57.6 (+0.2)    & 69.1 (-0.5)   & 65.2 (-0.2)   & 73.1 (-0.3)
\\

\bottomrule
\end{tabular}
\caption{Mean accuracy ($\uparrow, \%$) values with zero-shot prompting PTQ on ARC (easy), COPA, LAMBADA, PIQA, and SST2, this means we directly quantise the pre-trained model and benchmark on these downstream tasks using zero-shot prompting. We \good{highlight} 6-bit BFP which also achieves an accuracy close to FP32 on these tasks.
}
\label{tab:results:opt_ptq_downstream}
\end{table*}

\subsection{Experiment setup}
\label{subsec:experiment_setup}

\paragraph{Baselines}
We compare our approach with four baselines: 8-bit plain fixed-point quantisation,
\texttt{LLM.int8()}~\cite{dettmers2022llm}, GPTQ~\cite{frantar2022gptq}, and SmoothQuant~\cite{xiao2022smoothquant}.
We amend SmoothQuant's source code to ensure its consistency with their paper
(See \Cref{appendix:exp_setup}) and add this amended version (referred to as ``SmoothQuant-c'') to the result table.

\paragraph{Quantisation configuration}
\Cref{tab:results:quant_config} clarifies the quantisation configuration used in the following sections, where $E$, $M$, and $B$ are the bit-width of exponent (shared exponent), mantissa, and bias (shared bias) respectively. All these representations include a 1-bit sign bit. The block size of block-based methods is set to $[1,16]$ for both the weight and activation matrix (a slice along matrix row in \Cref{alg:transformer}) unless otherwise specified.

\paragraph{Models and datasets}
We choose the representative OPT~\cite{zhang2022opt} family, and evaluate on Wikitext2~\cite{merity2016pointer}, ARC(easy)~\cite{clark2018think}, LAMBADA~\cite{paperno-etal-2016-lambada}, PIQA~\cite{Bisk2020}, COPA~\cite{roemmele2011choice}, QNLI~\cite{wang-etal-2018-glue}, SST2~\cite{socher-etal-2013-recursive}, MRPC~\cite{dolan-brockett-2005-automatically}, and COLA~\cite{warstadt-etal-2019-neural}.  {To demonstrate the generalizability of our method, we also report the Wikitext2 perplexity of quantized LLaMA models~\cite{touvron2023llama, vicuna2023, alpaca}.} Following prior work~\cite{zhang2022opt, xiao2022smoothquant}, we use \texttt{lm-eval-harness}~\cite{eval-harness} to evaluate models on downstream tasks in the context of zero-shot prompting.

\subsection{Zero-shot PTQ on Wikitext2 and downstream tasks}
\label{sec:ptq_on_wikitext2}

In this section we present our results in a setup we call zero-shot Post-Training-Quantisation (PTQ), which was also adopted by prior work on LLM quantisation~\cite{dettmers2022llm,frantar2022gptq,xiao2022smoothquant}. In this approach, we take a pre-trained OPT model from Huggingface, quantise it, and apply it on Wikitext2 to calculate perplexity, and the eight downstream tasks shortlisted in \Cref{subsec:experiment_setup} to calculate accuracy.
The zero-shot PTQ setup is particularly advantageous in scenarios where LLMs lack prior knowledge, as it eliminates the need for downstream task fine-tuning and Training-After-Quantisation (TAQ).

\paragraph{Perplexity on Wikitext2}
\Cref{tab:results:opt_ptq_wikitext2} compares our results with the baselines in terms of perplexity, memory density, and arithmetic density.
Similar to prior work \cite{dettmers2022llm,xiao2022smoothquant}, plain fixed-point quantisation performs poorly. In contrast, non-linear arithmetic, such as MiniFloat, yields a significantly better perplexity at a similar memory density.
MiniFloat yields slightly better results than DMF, indicating the 2$\times$ higher range is more important than precision in this context.

Block-based quantisation exhibits inconsistent performance on Wikitext2. A noteworthy result is that our 6-bit BFP achieves higher memory density, higher arithmetic density, and lower perplexity than the prior art GPTQ and SmoothQuant-c without requiring data calibration. BM and BL perform poorly compared to BFP. BM was originally proposed in the context of Quantisation-Aware-Training (QAT), whereas our evaluation is based on PTQ. Without retraining, the 3-bit mantissa of BM and the 1-bit mantissa of BL may be the reason for the poor perplexity.

{\Cref{tab:results:other_llms} shows the perplexity of W6A6 BFP on LLaMA family, including LLaMA-7B/-13B~\cite{touvron2023llama}, Vicuna-7B~\cite{zheng2023judging}, Alpaca-7B~\cite{vicuna2023}, and Vicuna-v1.5-13B~\cite{vicuna2023}, with FP32 and \texttt{LLM.int8()} as baselines. We observe that 6-bit BFP still achieves nearly lossless perplexity on these models, verifying the efficacy of our method across model architectures.}

\begin{figure*}[ht]
\begin{tikzpicture}
\pgfplotsset{every x tick label/.append style={font=\footnotesize}}
\begin{axis}[
    width=\textwidth,
    ybar stacked, height=5cm,
    legend style={at={(1.0,1.0)},anchor=south east, draw=none},
    ymin=0, ymax=1,
    ytick={0,0.5,1},
    yticklabels={0,0.5, 1},
    xmin=0.3, xmax=32.7,
    ylabel={Percentage of occurrence},
    xlabel={Layer ID},
    legend style={font=\footnotesize},
    legend columns = 5,
    bar width=10,
    ]

\addplot[czbluel, draw=czbluel, fill=czbluel] table
[x=layers, y=freq4, col sep=space] {
layers freq4 freq5 freq6 freq7
1	0.633333333	0.366666667	0	0	0
2	0.766666667	0.233333333	0	0	0
3	0.3	0.366666667	0	0.333333333	0.333333333
4	0.466666667	0.233333333	0	0.3	0.3
5	0.433333333	0.266666667	0	0.3	0.3
6	0.866666667	0.133333333	0	0	0
7	0.266666667	0.266666667	0	0.466666667	0.466666667
8	0.533333333	0.166666667	0	0.3	0.3
9	0.4	0.3	0	0.3	0.3
10	0.533333333	0.166666667	0.3	0	0.3
11	0.433333333	0.233333333	0	0.333333333	0.333333333
12	0.7	0.3	0	0	0
13	0.666666667	0.333333333	0	0	0
14	0.733333333	0.266666667	0	0	0
15	0.466666667	0.2	0	0.333333333	0.333333333
16	0.8	0.2	0	0	0
17	0.7	0.3	0	0	0
18	0.133333333	0.2	0.3	0.366666667	0.666666667
19	0.366666667	0.266666667	0.366666667	0	0.366666667
20	0.7	0.3	0	0	0
21	0.433333333	0.233333333	0	0.333333333	0.333333333
22	0.633333333	0.366666667	0	0	0
23	0.7	0.3	0	0	0
24	0.8	0.2	0	0	0
25	0.066666667	0.233333333	0.366666667	0.333333333	0.7
26	0.4	0.3	0.3	0	0.3
27	0.666666667	0.333333333	0	0	0
28	0.4	0.3	0	0.3	0.3
29	0.5	0.2	0.3	0	0.3
30	0.133333333	0.166666667	0.366666667	0.333333333	0.7
31	0.666666667	0.333333333	0	0	0
32	0.366666667	0.166666667	0.466666667	0	0.466666667

};
\addlegendentry{$\leq$ 4-bit}

\addplot[czgreenl, draw=czgreenl, fill=czgreenl] table
[x=layers, y=freq5, col sep=space] {
layers freq4 freq5 freq6 freq7
1	0.633333333	0.366666667	0	0	0
2	0.766666667	0.233333333	0	0	0
3	0.3	0.366666667	0	0.333333333	0.333333333
4	0.466666667	0.233333333	0	0.3	0.3
5	0.433333333	0.266666667	0	0.3	0.3
6	0.866666667	0.133333333	0	0	0
7	0.266666667	0.266666667	0	0.466666667	0.466666667
8	0.533333333	0.166666667	0	0.3	0.3
9	0.4	0.3	0	0.3	0.3
10	0.533333333	0.166666667	0.3	0	0.3
11	0.433333333	0.233333333	0	0.333333333	0.333333333
12	0.7	0.3	0	0	0
13	0.666666667	0.333333333	0	0	0
14	0.733333333	0.266666667	0	0	0
15	0.466666667	0.2	0	0.333333333	0.333333333
16	0.8	0.2	0	0	0
17	0.7	0.3	0	0	0
18	0.133333333	0.2	0.3	0.366666667	0.666666667
19	0.366666667	0.266666667	0.366666667	0	0.366666667
20	0.7	0.3	0	0	0
21	0.433333333	0.233333333	0	0.333333333	0.333333333
22	0.633333333	0.366666667	0	0	0
23	0.7	0.3	0	0	0
24	0.8	0.2	0	0	0
25	0.066666667	0.233333333	0.366666667	0.333333333	0.7
26	0.4	0.3	0.3	0	0.3
27	0.666666667	0.333333333	0	0	0
28	0.4	0.3	0	0.3	0.3
29	0.5	0.2	0.3	0	0.3
30	0.133333333	0.166666667	0.366666667	0.333333333	0.7
31	0.666666667	0.333333333	0	0	0
32	0.366666667	0.166666667	0.466666667	0	0.466666667

};
\addlegendentry{5-bit}

\addplot[czorangel, draw=czorangel, fill=czorangel] table
[x=layers, y=freq6, col sep=space] {
layers freq4 freq5 freq6 freq7
1	0.633333333	0.366666667	0	0	0
2	0.766666667	0.233333333	0	0	0
3	0.3	0.366666667	0	0.333333333	0.333333333
4	0.466666667	0.233333333	0	0.3	0.3
5	0.433333333	0.266666667	0	0.3	0.3
6	0.866666667	0.133333333	0	0	0
7	0.266666667	0.266666667	0	0.466666667	0.466666667
8	0.533333333	0.166666667	0	0.3	0.3
9	0.4	0.3	0	0.3	0.3
10	0.533333333	0.166666667	0.3	0	0.3
11	0.433333333	0.233333333	0	0.333333333	0.333333333
12	0.7	0.3	0	0	0
13	0.666666667	0.333333333	0	0	0
14	0.733333333	0.266666667	0	0	0
15	0.466666667	0.2	0	0.333333333	0.333333333
16	0.8	0.2	0	0	0
17	0.7	0.3	0	0	0
18	0.133333333	0.2	0.3	0.366666667	0.666666667
19	0.366666667	0.266666667	0.366666667	0	0.366666667
20	0.7	0.3	0	0	0
21	0.433333333	0.233333333	0	0.333333333	0.333333333
22	0.633333333	0.366666667	0	0	0
23	0.7	0.3	0	0	0
24	0.8	0.2	0	0	0
25	0.066666667	0.233333333	0.366666667	0.333333333	0.7
26	0.4	0.3	0.3	0	0.3
27	0.666666667	0.333333333	0	0	0
28	0.4	0.3	0	0.3	0.3
29	0.5	0.2	0.3	0	0.3
30	0.133333333	0.166666667	0.366666667	0.333333333	0.7
31	0.666666667	0.333333333	0	0	0
32	0.366666667	0.166666667	0.466666667	0	0.466666667

};
\addlegendentry{6-bit}

\addplot[czredl, draw=czredl, fill=czredl] table
[x=layers, y=freq7, col sep=space] {
layers freq4 freq5 freq6 freq7
1	0.633333333	0.366666667	0	0	0
2	0.766666667	0.233333333	0	0	0
3	0.3	0.366666667	0	0.333333333	0.333333333
4	0.466666667	0.233333333	0	0.3	0.3
5	0.433333333	0.266666667	0	0.3	0.3
6	0.866666667	0.133333333	0	0	0
7	0.266666667	0.266666667	0	0.466666667	0.466666667
8	0.533333333	0.166666667	0	0.3	0.3
9	0.4	0.3	0	0.3	0.3
10	0.533333333	0.166666667	0.3	0	0.3
11	0.433333333	0.233333333	0	0.333333333	0.333333333
12	0.7	0.3	0	0	0
13	0.666666667	0.333333333	0	0	0
14	0.733333333	0.266666667	0	0	0
15	0.466666667	0.2	0	0.333333333	0.333333333
16	0.8	0.2	0	0	0
17	0.7	0.3	0	0	0
18	0.133333333	0.2	0.3	0.366666667	0.666666667
19	0.366666667	0.266666667	0.366666667	0	0.366666667
20	0.7	0.3	0	0	0
21	0.433333333	0.233333333	0	0.333333333	0.333333333
22	0.633333333	0.366666667	0	0	0
23	0.7	0.3	0	0	0
24	0.8	0.2	0	0	0
25	0.066666667	0.233333333	0.366666667	0.333333333	0.7
26	0.4	0.3	0.3	0	0.3
27	0.666666667	0.333333333	0	0	0
28	0.4	0.3	0	0.3	0.3
29	0.5	0.2	0.3	0	0.3
30	0.133333333	0.166666667	0.366666667	0.333333333	0.7
31	0.666666667	0.333333333	0	0	0
32	0.366666667	0.166666667	0.466666667	0	0.466666667

};
\addlegendentry{7-bit}

\end{axis}
\end{tikzpicture}
\caption{The bit width distribution of $\bf{Q}$ in Line 6, \Cref{alg:transformer} from 2688 searches. We identify the layers less tolerant to aggressive quantisation in OPT-2.7B. For example, layers 18, 25 and 30 often need more bits than other layers. Keeping these layers in relatively high precision recovers the accuracy from 36.2\% to 61.3\% without decreasing the memory density, equivalent to a 4.3-bit OPT-2.7B on average.}
\label{fig:results:searched_bar_chart}
\end{figure*}
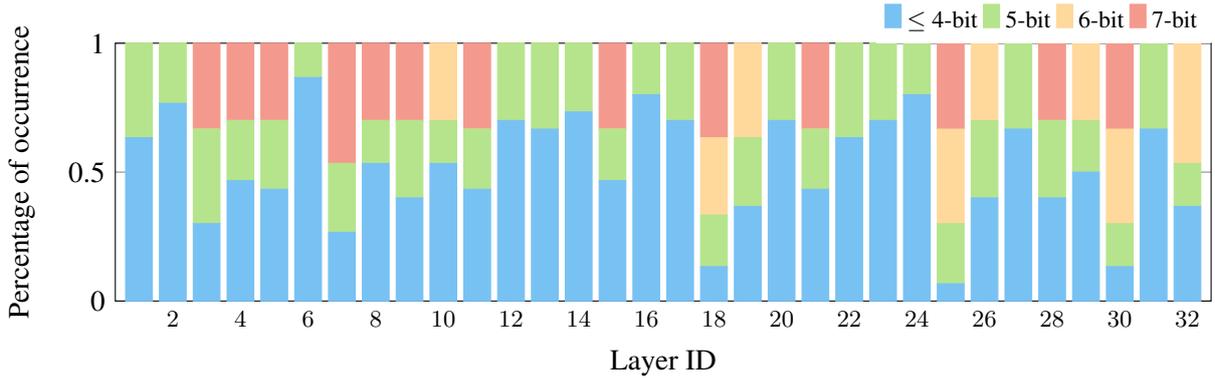

\paragraph{Accuracy on downstream tasks}
We exclude fixed-point, DMF, BM, and BL from downstream task evaluation due to their poor language modelling performance. \Cref{tab:results:opt_ptq_downstream} represents the mean accuracy on ARC (easy), COPA, LAMBADA, PIQA, and SST2. The results of QNLI, MRPC, and COLA are not included in this table as even FP32 LLMs exhibited poor accuracy close to random guess. A plot depicting how these methods match FP32 accuracy as the model scales up and a complete result table are in \Cref{appendix:ptq_on_downstream}.

Besides \texttt{LLM.int8()} and SmoothQuant-c, we also report a 4-bit version \texttt{LLM.int8()} (referred to as \texttt{LLM.int4()}) reported by \citet{2023-dettmers-bitsandbytes} on downstream tasks. We observe that 6-bit BFP achieve nearly lossless accuracy, below FP32 and \texttt{LLM.int8()}, and above SmoothQuant-c and  \texttt{LLM.int4()}. Note that 6-bit BFP has the highest memory density and arithmetic density among these methods. The 4-bit BFP suffers severe accuracy degradation because its shared exponent and 3-bit mantissa cause large quantisation errors.

Overall, we make the following observations:
\begin{itemize}
    \item Fixed-point representation performs inadequately due to unability of linear quantisation to address the scaling offset issue caused by varying variances.
    \item LLMs have different tolerance to block-based quantisations. BM and BL exhibit subpar performance compared to BFP, indicating that non-linear quantisation still needs sufficient mantissa length to capture the learned weight distribution, or retraining may be required.
    \item BFP strikes a good balance in the trade-off between range and resolution. Our nearly-lossless 6-bit LLMs, without data calibration/re-training, outperform prior art methods in terms of perplexity (accuracy), memory density, and arithmetic density.

\end{itemize}

We also observe that sub-6-bit BFP has a severe accuracy drop. To address this problem, we further investigate two approaches for improving the accuracy of 4-bit LLMs.

\subsection{4-bit LLMs via fine-tuning}

Previous study~\cite{brown2020language, zhang2022opt} reported FP32 LLMs' low accuracy  on several downstream tasks in the context of zero-shot prompting. In our experiments, OPTs also exhibit poor accuracy on QNLI, MRPC, and COLA. Fine-tuning language models on downstream tasks has proven to be helpful for improving accuracy \cite{devlin2019bert}. We explore the fine-tuning and quantisation of LLMs on downstream tasks.

There are two stages where quantisation can be applied. LLMs are typically pre-trained in FP32. The first option is to continue fine-tuning the FP32 model on downstream tasks and subsequently quantise this fine-tuned FP32 model. We refer to this setup as \textit{PTQ on fine-tuned FP32}. The second option is to quantise the pre-trained FP32 model and retrain this quantised model on downstream tasks, which we refer to as \textit{TAQ on downstream tasks}.

We compare these two cases on four downstream tasks (SST2, QNLI, MRPC, and COLA) that zero-shot prompting struggles to handle. The result table is in \Cref{appendix:ptq_vs_taq}.
We observe that:
\begin{itemize}
    \item Both options effectively improve accuracy, enabling nearly lossless downstream accuracy even if 4-bit BFP is applied.
    \item TAQ on downstream tasks reaches a slightly better accuracy (a gain of 0.2\% on average) than PTQ on fine-tuned \texttt{FP32} given the same bit-width. However, the former is harder to optimize through backpropagation because of the forward quantisation error and the Straight-Through Estimator (STE) \cite{bengio2013estimating} used in backpropagation.
\end{itemize}


\subsection{4-bit LLMs via mixed precision}
\label{sec:mixed-precision}

Currently, our block-based quantisation uses a uniform configuration, where the block size and bit-width remain constant across the entire model. What if we push the barrier further? Existing works on CNN compression have explored mixed-precision quantisation~\cite{wu2018mixed, wang2019haq}, thereby increasing memory density. This subsection lowers the block size granularity and the bit-width granularity to the tensor level to demonstrate uncharted possibilities of aggressive LLM quantisation.

\paragraph{Variation-aware block size}
By comparing the activation variance and weight variance in \Cref{fig:introduction:motivation}, we observe that the weight variance remains stable and much smaller, suggesting that we can increase the weight block size while decreasing the activation block size. This approach enhances accuracy while maintaining memory density.

\paragraph{Mixed-precision}
We repeat the quantisation search described in \Cref{subsec:quantisation_search} on downstream tasks and filter out less promising quantisation configurations using an accuracy threshold and a memory density threshold.  Each time we start TPE search with a different random seed, so the distribution of filtered quantisation configurations exposed the sensitivity of the searched tensors in LLMs.
An example of a mixed-precision search result is presented in \Cref{fig:results:searched_bar_chart}. We find \textit{certain layers were consistently assigned with higher precision, while others tended to have lower bit widths}. By preserving high precision for these sensitive layers, we recovered the 4-bit LLM accuracy \textit{from 36.2\% to 61.3\%} on LAMBADA without compromising memory density. {The memory density of the searched OPT-2.7B is 7.42$\times$, which is slightly better than the uniform 4-bit BFP's 7.11$\times$. \Cref{fig:appendix:mixed_precision_vs_uniform} in~\Cref{appendix:mixed_precision_search} compares uniform 4-bit BFP and mixed-precision 4-bit BFP on LAMBADA and ARC (easy), highlighting the effectiveness of our mixed-precision quantisation.} We include more tasks and model sizes in~\Cref{appendix:mixed_precision_search}. In conclusion, variance-aware block size and mixed precision allow aggressive quantisation beyond 6-bit without fine-tuning.


\section{Conclusion}
This study focuses on addressing the scaling offset issue in LLMs and provides valuable insights into the quantisation of LLMs. Through extensive experimentation, we identify key factors that significantly impact LLM quantisation. When aiming for quantisation above or equal to 6-bit, BFP surpasses previous methods in terms of accuracy, memory density, and arithmetic density, without requiring for data calibration or training. Moreover, we demonstrate that fine-tuning or mixed precision techniques enable 4-bit LLMs on downstream tasks. Fine-tuning is suitable for GPUs, and mixed precision has the potential to shift the inference platform from GPUs to cost-effective ASICs. Our findings contribute to advancing the field of LLM quantisation and provide practical guidance for achieving good quantisation performance.

\section*{Limitations}

Different from many prior arts in LLM quantisation that focus on integers, our work puts particular emphasis on minifloat variants. However, the potential gains of our work have not manifested in GPU systems due to a lack of CUDA kernel implementation. The implementation of some proposed quantisation methods in this paper requires specialised kernels and hardware, however, a major focus of our work is to \emph{explore potential designs for next-generation hardware to run LLM inference}.
Another limitation is that our search algorithm does not include arithmetic density due to a lack of hardware models {for LLMs. We ran a mixed-precision search with hardware models on a small transformer. The result included in~\Cref{appendix:mixed_precision_with_dse} is promising. We leave sufficient study on hardware-aware LLM quantization as a future work.}

\bibliography{references}
\bibliographystyle{acl_natbib}

\appendix

\section{Tensor variance in LLMs}
\begin{figure*}[h]
\centering
\begin{minipage}[b]{.4\linewidth}
\input{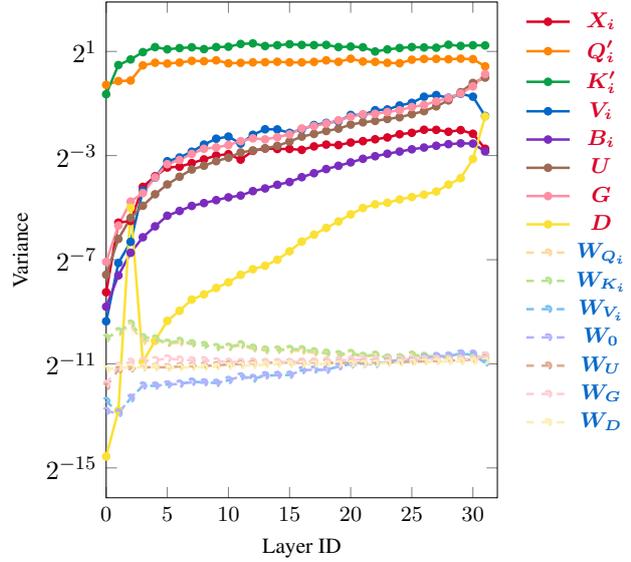}
\end{minipage}
\begin{minipage}[b]{0.45\linewidth}
    \begin{tikzpicture}[thick,scale=0.95, every node/.style={scale=0.95}]

\pgfplotsset{every x tick label/.append style={font=\footnotesize}, compat=1.3}
\begin{semilogyaxis}[
    height=85mm,
    width=70mm,
    xlabel={\footnotesize Layer ID},
    ylabel={\footnotesize Variance},
    ylabel style = {align=center},
    xmin=0, xmax=32,
    xtick={0, 5, 10, 15, 20, 25, 30},
    ymode=log, log ticks with fixed point,
    log basis y={2},
    ytick={0.0000306, 0.00048828125, 0.0078125, 0.125, 2, 32},
    yticklabels={$2^{-15}$, $2^{-11}$, $2^{-7}$, $2^{-3}$, $2^{1}$, $2^5$},
    scaled x ticks = false,
    scaled y ticks = false,
    legend columns = 1,
    legend style={at={(1.05,1)},anchor=north west, draw=none, fill=none, font=\footnotesize},
    ]

\addplot[czred, draw=czred, mark=*, line width=1pt, mark options={scale=0.6}] table
[x=block_id, y=attn_in, col sep=space] {vicuna7b.dat};
\addlegendentry{$\varied{X_i}$}

\addplot[czorange, draw=czorange, mark=*, line width=1pt, mark options={scale=0.6}] table
[x=block_id, y=q_out, col sep=space] {vicuna7b.dat};
\addlegendentry{$\varied{Q_i'}$}

\addplot[czgreen, draw=czgreen, mark=*, line width=1pt, mark options={scale=0.6}] table
[x=block_id, y=k_out, col sep=space] {vicuna7b.dat};
\addlegendentry{$\varied{K_i'}$}

\addplot[czblue, draw=czblue, mark=*, line width=1pt, mark options={scale=0.6}] table
[x=block_id, y=v_out, col sep=space] {vicuna7b.dat};
\addlegendentry{$\varied{V_i}$}

\addplot[czpurple, draw=czpurple, mark=*, line width=1pt, mark options={scale=0.6}] table
[x=block_id, y=mlp_in, col sep=space] {vicuna7b.dat};
\addlegendentry{$\varied{B_i}$}

\addplot[czbrown, draw=czbrown, mark=*, line width=1pt, mark options={scale=0.6}] table
[x=block_id, y=up_out, col sep=space] {vicuna7b.dat};
\addlegendentry{$\varied{U}$}

\addplot[czpink, draw=czpink, mark=*, line width=1pt, mark options={scale=0.6}] table
[x=block_id, y=gate_out, col sep=space] {vicuna7b.dat};
\addlegendentry{$\varied{G}$}

\addplot[czyellow, draw=czyellow, mark=*, line width=1pt, mark options={scale=0.6}] table
[x=block_id, y=down_in, col sep=space] {vicuna7b.dat};
\addlegendentry{$\varied{D}$}


\addplot[dashed, czorangel, draw=czorangel, mark=o, line width=1pt, mark options={scale=0.6}] table
[x=block_id, y=q_proj_weight, col sep=space] {vicuna7b.dat};
\addlegendentry{$\capped{W_{Q_i}}$}

\addplot[dashed, czgreenl, draw=czgreenl, mark=o, line width=1pt, mark options={scale=0.6}] table
[x=block_id, y=k_proj_weight, col sep=space] {vicuna7b.dat};
\addlegendentry{$\capped{W_{K_i}}$}

\addplot[dashed, czbluel, draw=czbluel, mark=o, line width=1pt, mark options={scale=0.6}] table
[x=block_id, y=v_proj_weight, col sep=space] {vicuna7b.dat};
\addlegendentry{$\capped{W_{V_i}}$}

\addplot[dashed, czpurplel, draw=czpurplel, mark=o, line width=1pt, mark options={scale=0.6}] table
[x=block_id, y=o_proj_weight, col sep=space] {vicuna7b.dat};
\addlegendentry{$\capped{W_{0}}$}

\addplot[dashed, czbrownl, draw=czbrownl, mark=o, line width=1pt, mark options={scale=0.6}] table
[x=block_id, y=up_proj_weight, col sep=space] {vicuna7b.dat};
\addlegendentry{$\capped{W_{U}}$}

\addplot[dashed, czpinkl, draw=czpinkl, mark=o, line width=1pt, mark options={scale=0.6}] table
[x=block_id, y=gate_proj_weight, col sep=space] {vicuna7b.dat};
\addlegendentry{$\capped{W_{G}}$}

\addplot[dashed, czyellowl, draw=czyellowl, mark=o, line width=1pt, mark options={scale=0.6}] table
[x=block_id, y=down_proj_weight, col sep=space] {vicuna7b.dat};
\addlegendentry{$\capped{W_{D}}$}

\end{semilogyaxis}
\end{tikzpicture}
\end{minipage}
\caption{
    {The algorithm on the left is the forward pass of Vicuna-7B. The graph on the right depicts how the tensor variances change with layer number. We observe the same trend of increasing activation variances in Vicuna-7B, which is similar to OPT-6.7B. interestingly, the variances of self-attention input activations ($\bf{Q_i'}$ and $\bf{K_i'}$ in line 6 of the algorithm above) are consistently high from the first Transformer layer to the last in Vicuna-7B. We assume this is because of the Rotary Positional Encoding (RoPE)~\cite{su2021roformer} layers.}
}
\label{fig:appendix:motivation_vicuna}
\end{figure*}
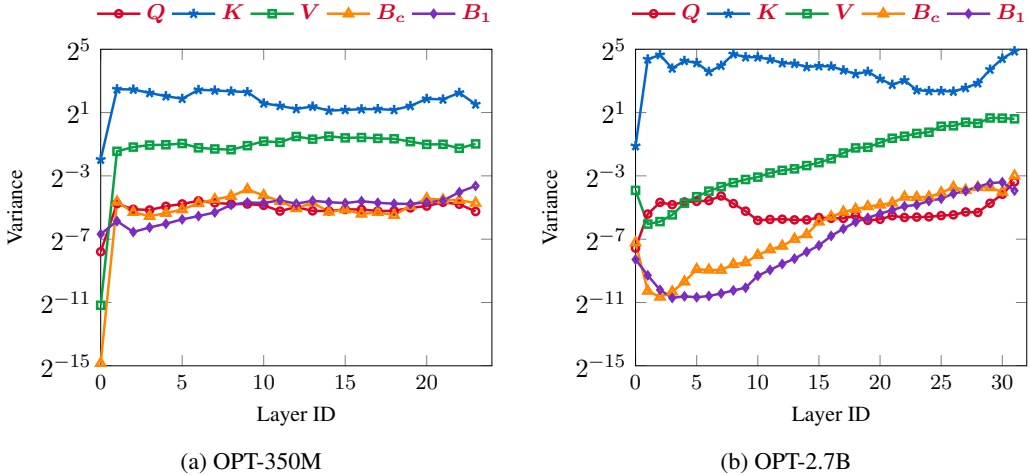
\begin{figure*}[ht]
    \centering
    \subcaptionbox{OPT-350M\label{fig:var_opt_350m}}{
        \begin{tikzpicture}[thick,scale=0.95, every node/.style={scale=0.95}]
\pgfplotsset{every x tick label/.append style={font=\footnotesize}, compat=1.3}
\begin{axis}[
    height=60mm,
    width=70mm,
    xlabel={\footnotesize Layer ID},
    ylabel={\footnotesize Variance},
    ylabel style = {align=center},
    xmin=0, xmax=24,
    ymax = 32, ymin = 3.0517578125e-05,
    xtick={0, 5, 10, 15, 20, 25, 30},
    ymode=log, log ticks with fixed point,
    log basis y={2},
    ytick={0.0000306, 0.00048828125, 0.0078125, 0.125, 2, 32},
    yticklabels={$2^{-15}$, $2^{-11}$, $2^{-7}$, $2^{-3}$, $2^{1}$, $2^5$},
    scaled x ticks = false,
    scaled y ticks = false,
    legend columns = 5,
    legend style={at={(0.5, 1.05)},anchor=south, draw=none, fill=none, font=\footnotesize},
    ]

\addplot[czred, draw=czred, mark=o, line width=1pt, mark options={scale=0.7}] table
[x=layer, y=q, col sep=space] {opt350m.dat};
\addlegendentry{$\varied{Q}$}

\addplot[czblue, draw=czblue, mark=star, line width=1pt] table
[x=layer, y=k, col sep=space] {opt350m.dat};
\addlegendentry{$\varied{K}$}

\addplot[czgreen, draw=czgreen, mark=square, line width=1pt, mark options={scale=0.7}] table
[x=layer, y=v, col sep=space] {opt350m.dat};
\addlegendentry{$\varied{V}$}

\addplot[czorange, draw=czorange, mark=triangle, line width=1pt] table
[x=layer, y=b0, col sep=space] {opt350m.dat};
\addlegendentry{$\varied{B_c}$}

\addplot[czpurple, draw=czpurple, mark=diamond, line width=1pt, mark options={scale=0.7}] table
[x=layer, y=b1, col sep=space] {opt350m.dat};
\addlegendentry{$\varied{B_1}$}

\end{axis}
\end{tikzpicture}
    }
    \subcaptionbox{OPT-2.7B\label{fig:var_opt_2.7b}}{
        \begin{tikzpicture}[thick,scale=0.95, every node/.style={scale=0.95}]
\pgfplotsset{every x tick label/.append style={font=\footnotesize}, compat=1.3}
\begin{axis}[
    height=60mm,
    width=70mm,
    xlabel={\footnotesize Layer ID},
    ylabel={\footnotesize Variance},
    ylabel style = {align=center},
    xmin=0, xmax=32,
    ymax = 32, ymin = 3.0517578125e-05,
    xtick={0, 5, 10, 15, 20, 25, 30},
    ymode=log, log ticks with fixed point,
    log basis y={2},
    ytick={0.0000306, 0.00048828125, 0.0078125, 0.125, 2, 32},
    yticklabels={$2^{-15}$, $2^{-11}$, $2^{-7}$, $2^{-3}$, $2^{1}$, $2^5$},
    scaled x ticks = false,
    scaled y ticks = false,
    legend columns = 5,
    legend style={at={(0.5, 1.05)},anchor=south, draw=none, fill=none, font=\footnotesize},
    ]

\addplot[czred, draw=czred, mark=o, line width=1pt, mark options={scale=0.7}] table
[x=layer, y=q, col sep=space] {opt2.7b.dat};
\addlegendentry{$\varied{Q}$}

\addplot[czblue, draw=czblue, mark=star, line width=1pt] table
[x=layer, y=k, col sep=space] {opt2.7b.dat};
\addlegendentry{$\varied{K}$}

\addplot[czgreen, draw=czgreen, mark=square, line width=1pt, mark options={scale=0.7}] table
[x=layer, y=v, col sep=space] {opt2.7b.dat};
\addlegendentry{$\varied{V}$}

\addplot[czorange, draw=czorange, mark=triangle, line width=1pt] table
[x=layer, y=b0, col sep=space] {opt2.7b.dat};
\addlegendentry{$\varied{B_c}$}

\addplot[czpurple, draw=czpurple, mark=diamond, line width=1pt, mark options={scale=0.7}] table
[x=layer, y=b1, col sep=space] {opt2.7b.dat};
\addlegendentry{$\varied{B_1}$}

\end{axis}
\end{tikzpicture}
    }
    \caption{We demonstrate a similar analysis to \Cref{fig:introduction:motivation}, where on the left we have OPT-350M variance vs layer ID and OPT-2.7B variance vs layer ID on the right. The trend of increasing activation variance is more obvious on larger models.}
    \label{fig:appendix:tensor_variances}
\end{figure*}

We investigate the variance trend of Vicuna-7B~\cite{vicuna2023} and observe the same trend of increasing activation variances as OPT-6.7B. \Cref{fig:appendix:motivation_vicuna} depicts the forward pass of Vicuna-7B and the variance trend of its tensors. Interestingly, the variances of self-attention input activations ($\bf{Q_i'}$ and $\bf{K_i'}$ in line 6 of the algorithm above) are consistently high from the first Transformer layer to the last in Vicuna-7B. We assume this is because of the Rotary Positional Encoding (RoPE)~\cite{su2021roformer} layers.

We additionally analyse the trend of increasing activation variance when the model size increases,
In \Cref{fig:introduction:motivation}, for OPT-6.7B, we plotted the variances of all tensors that have unbounded input ranges and that are taken as input operands to matrix multiplications in the Transformer layer. {\Cref{fig:appendix:tensor_variances}} further illustrates the results for OPT-350M and OPT-2.7B. We observe that:
\begin{itemize}
    \item If we consider $\mathbf{V}$, $\mathbf{B_c}$ and $\mathbf{B_1}$ as the \emph{main information path} \footnote{The computation of $\mathbf{Q}$ and $\mathbf{K}$ yields attention factors (post-softmax) that are applied to $\mathbf{V}$.}, these components have much smaller variances than $\mathbf{K}$ and $\mathbf{Q}$.
    \item Bigger models tend to have small variances at shallow layers and larger variances at deep layers.
\end{itemize}

These observations explain why linear quantisation, such as integer quantisation, is effective for smaller models but struggles with larger ones. This increasing activation variance trend can be considered into variance-aware block size. Since a higher variance implies a higher possibility of extreme outliers, we can apply larger block sizes to those tensors with smaller variance and smaller block sizes to those with higher variance. Limited by time, we leave this exploration as well as the combination of fine-tuning, variance-aware block size, and mixed precision in future work.

\section{Experiment details}
\label{appendix:exp_setup}

\subsection{Setup and Implementation}
\paragraph{Hardware resources}
We run the experiments using four NVIDIA RTX3090s, three A100s, and eight V100s with 64GB, 192GB, and 128GB RAM respectively. The evaluation of PTQ perplexity on Wikitext2 takes around 64 GPU hours in total; the zero-shot prompting evaluation on downstream tasks takes around 160 GPU hours in total; the fine-tuning of FP32 models on SST2, QNLI, MRPC and COLA takes around 30 GPU hours in total; the fine-tuning of quantised BFP models takes around 70 GPU hours in total; the evaluation of fine-tuned models takes around 6 GPU hours in total; the mixed-precision search takes around 120 GPU hours in total.

\paragraph{Implementation}
We download the model codes and pre-trained weights from HuggingFace Transformers \footnote{\url{https://github.com/huggingface/transformers}} and implement the quantisation arithmetics using PyTorch \footnote{\url{https://github.com/pytorch/pytorch}}. We use Vivado to report arithmetic density and Optuna\footnote{\url{https://optuna.readthedocs.io/en/stable/index.html}} to perform the mixed-precision search.

\paragraph{Evaluation}
We follow the code base of GPTQ~\cite{frantar2022gptq}\footnote{\url{https://github.com/IST-DASLab/gptq}} to estimate LLM's perplexity on Wikitext2. We chop Wikitext2's test set into sequences of 2000 tokens, feed the sequences to LLMs, and normalise the cross entropy loss by the sequence length and batch size. To evaluate LLM accuracy on downstream tasks, we follow OPT~\cite{zhang2022opt} and SmoothQuant~\cite{xiao2022smoothquant} to use \texttt{lm-eval-harness} in the zero-shot prompting setup.

\subsection{Comparison with SmoothQuant}
\label{appendix:smoothquant}

The SmoothQuant paper~\cite{xiao2022smoothquant} declares all the eight GEMMs (\textcircled{\small{1}}-\textcircled{\small{8}} in \ref{alg:transformer}) are quantised. However, their codes\footnote{\url{https://github.com/mit-han-lab/smoothquant}} do not support quantising \textcircled{\small{5}} and \textcircled{\small{6}}, which takes up 19.6\% floating-point operations (FLOPs) in OPT-6.7B's self-attention. We amend their code and refer to the amended version as ``SmoothQuant-c'', which should be the same as SmoothQuant-O2 in the paper. We observe that SmoothQuant-c has much higher perplexity and slightly lower accuracy on downstream tasks than SmoothQuant. Besides, the SmoothQuant repository does not include the scaling factor files of OPT-125m and OPT-350m, so the perplexity/accuracy for these two models is missing in our result table.

\subsection{Comparison with \texttt{LLM.int8()}}

We give a short comparison between \texttt{LLM.int8()} and our method. \texttt{LLM.int8()} is different from plain 8-bit fixed-point quantization. In \texttt{LLM.int8()} all the tensors are stored as FP16 numbers, which is the reason why \texttt{LLM.int8()} has 2$\times$ memory density while plain 8-bit fixed-point has 4$\times$ memory density in~\Cref{tab:results:opt_ptq_wikitext2}. \texttt{LLM.int8()} targets GPUs while ours targets ASICs. \texttt{LLM.int8()} is not as friendly as uniform BFP to ASICs, because \texttt{LLM.int8()} separates one matrix multiply into two (one for inliers the other for outliers), casts inliers to 8-bit, performs an 8-bit matrix multiplication for inliers and an FP16 matrix multiplication for outliers respectively. This separation is performed on the fly. In comparison, 6-bit uniform BFP does not require a runtime separation or an FP16 matrix multiply engine. All tensors are stored and calculated in 6-bit BFP.

\subsection{Quantisaion search}

The specific search configuration depends on model size, task, and FP32 performance.
We use the accuracy threshold and memory density threshold to sort out promising mixed-precision configs. Given a model and a task, the accuracy threshold is 2\% below the FP32 values. The memory density is set to 7.1\% in most search configs.

Note that to estimate the memory density of quantisation config candidates, we need the model architecture information including input sizes and weight sizes for all the GEMMs in \Cref{alg:transformer} across all layers. We implement a FLOP profiler to collect this information and feed it as input to the search algorithm. The numeric values of these parameters can be found in the bash scripts of our source code.

\begin{table*}[ht]
\centering
\begin{tabular}{lcrrrrr}
\toprule
Method & Config & Block size & \#DSPs & \#LUTs & Area factor & Arithmetic density \\
\midrule
FP32 & - & 1 & 5 & 335 & 835 & 1$\times$ \\
Integer & W8A8 & 1 & 1 & 9 & 109 & 7.7$\times$ \\
MiniFloat & W8A8 & 1 & 0 & 48 & 48 & 17.4$\times$ \\
BM & W8A8 & 1 & 0 & 27 & 51 & 16.4$\times$ \\
BFP & W8A8 & 16 & 0 & 544 & 58 & 14.4$\times$ \\
BL & W8A8 & 1 & 0 & 28 & 52 & 16.1$\times$ \\
BFP & W6A6 & 16 & 0 & 313 & 43.6 & 19.2$\times$ \\
BFP & W4A4 & 16 & 0 & 358 & 22.4 & 37.3$\times$ \\
\bottomrule
\end{tabular}
\caption{
The arithmetic density of various quantisation configurations explored in this paper. To calculate the area factor, we convert the Digital Signal Processing units (DSPs) to equivalent LUTs to get the area factor, and then divide the quantisation arithmetic's area factor density by FP32.
}
\label{tab:appendix:arith_density}
\end{table*}

\section{Definition of quantisation arithmetics}
\label{appendix:quant_arith}
\paragraph{FP32, FP16 and MiniFloat}
A traditional floating-point representation follows IEEE floating-point standard~\cite{kahan1996ieee}, which can define a floating-point number as a 4-tuple, $(s, e, m, b)$, where
\begin{itemize}
    \item $s \in \{0, 1\}$ is the sign bit of the number;
    \item $e \in \mathbb{N}$ is the exponent field;
    \item $b \in \mathbb{N}$ is the exponent bias; and
    \item $m \in \mathbb{N}$ is the mantissa.
\end{itemize}
Given the bit widths of the exponent and the mantissa be $E$ and $M$, the value $x$ of a floating-point number can be obtained via:

\begin{equation}
\begin{medsize}
        x =
        \begin{cases}
        (-1)^s \times 2^{1-b} \times \frac{m}{2^M} & e = 0\\
        (-1)^s \times 2^{e-b} \times (1+\frac{m}{2^M}) & 0 < e < 2^E-1 \\
        (-1)^s \times \infty & e = 2^E-1, m = 0 \\
        \text{NaN} & \text{others} \\
        \end{cases}
\end{medsize}
\label{eqn:float32}
\end{equation}

where $e$ is the unsigned integer value represented by the exponent bits, and $m$ is the unsigned integer value represented by mantissa bits. The exponent bias ($b$) is a constant depending on $E$: $b=2^{E-1}-1$. FP32, FP16, and MiniFloat have $E=8, M=23$, $E=5, M=10$ and $E=4, M=3$, respectively. Note that the ``1'' in the fraction term of Line 2, \Cref{eqn:float32} comes from the implicit leading bit in the mantissa.

We additionally saturate MiniFloat when $e=2^E-1$, thus the value of a MiniFloat is

\begin{equation}
\begin{medsize}
        x =
        \begin{cases}
        (-1)^s \times 2^{1-b} \times \frac{m}{2^M} & e = 0\\
        (-1)^s \times 2^{e-b} \times (1+\frac{m}{2^M}) & 0 < e \leq 2^E-1 \\
        \text{NaN} & \text{others} \\
        \end{cases} 
\end{medsize}
\label{eqn:sat_float}
\end{equation}

\paragraph{DMF}
The definition of DMF is the same as MiniFloat except that there is no implicit leading bit in the mantissa:
\begin{equation}
\begin{medsize} 
    x =
    \begin{cases}
    (-1)^s \times 2^{e-b} \times \frac{m}{2^M} & e \leq 2^E-1\\
    \text{NaN} & \text{others} \\
    \end{cases} 
\end{medsize}
\label{eqn:denorm}
\end{equation}

\paragraph{BM, BL, and BFP}

BM~\cite{fox2021block} shares the exponent bias and was proposed in the context of Quantisaion-Aware-Training (QAT). When an FP32 value is cast to BM, the exponent bias is determined by the maximum value in the block.

BFP~\cite{darvish2020pushing} shares the exponent and was proposed in the context of PTQ. Similar to BM, the shared exponent bias is also determined by the maximum FP32 values when casted from FP32.

Logarithm quantisation was proposed by~\citet{miyashita2016convolutional} to perform QAT on CNNs. Block Logarithm (BL) was used as a baseline to compare with BM in~\cite{fox2021block}. BL shares the exponent bias and does not have mantissa bits (mantissa is always 1).

Basically, block-based quantisation facilitates the vector's inner product by simplifying the accumulation after multiplication. For example, the inner product between two BFP vectors $\mathbf{x}$ and $\mathbf{y}$ is,
\begin{equation} 
\begin{medsize} 
        \begin{split}
           & \mathbf{x}\cdot\mathbf{y} \\
           = & (-1)^{s_x} e^{e_x} \left[ x_1, \dots, x_{B-1} \right] \cdot \\
             & (-1)^{s_y} e^{e_y} \left[ y_1, \dots, y_{B-1} \right] \\
           = & (-1)^{s_x+s_y} e^{e_x+e_y} \left[ x_1y_1 + \dots +x_{B-1}y_{B-1} \right]
        \end{split}
\end{medsize}
\end{equation}
where $B$ is the block size. Since exponents are shared across vectors, the element products can be accumulated without shifting. The block sizes of the two vectors are not necessarily the same.

\begin{table*}[]
\centering

\resizebox{\textwidth}{!}{%
\begin{tabular}{lllcccccll}
\toprule
\multicolumn{1}{r}{Method} & Config & Model size & \multicolumn{1}{l}{ARC (easy)} & \multicolumn{1}{l}{COPA} & \multicolumn{1}{l}{LAMBADA} & \multicolumn{1}{l}{PIQA} & \multicolumn{1}{l}{QNLI} & SST2 & Average \\ \midrule
\multicolumn{1}{c}{\multirow{5}{*}{FP32}} & \multirow{5}{*}{-} & 125M & 43.5\% & 66.0\% & 37.9\% & 63.0\% & 49.4\% & 53.3\% & 52.7\% \\
\multicolumn{1}{c}{} &  & 350M & 44.0\% & 72.0\% & 45.2\% & 64.4\% & 49.5\% & 61.8\% & 57.5\% \\
\multicolumn{1}{c}{} &  & 1.3B & 57.0\% & 79.0\% & 57.9\% & 71.7\% & 51.3\% & 82.2\% & 69.6\% \\
\multicolumn{1}{c}{} &  & 2.7B & 60.8\% & 77.0\% & 63.6\% & 73.9\% & 51.1\% & 51.7\% & 65.4\% \\
\multicolumn{1}{c}{} &  & 6.7B & 65.6\% & 81.0\% & 67.7\% & 76.3\% & 50.9\% & 76.5\% & 73.4\% \\ \midrule
\multirow{5}{*}{\texttt{LLM.int8()}} & \multirow{5}{*}{W8A8} & 125M & 43.6\% & 66.0\% & 37.7\% & 63.0\% & 49.5\% & 52.1\% & 52.5\% \\
 &  & 350M & 43.8\% & 72.0\% & 45.3\% & 64.2\% & 49.5\% & 66.1\% & 58.3\% \\
 &  & 1.3B & 57.5\% & 79.0\% & 57.7\% & 71.6\% & 51.1\% & 80.1\% & 69.2\% \\
 &  & 2.7B & 60.6\% & 78.0\% & 62.9\% & 72.9\% & 51.2\% & 52.1\% & 65.3\% \\
 &  & 6.7B & 65.5\% & 83.0\% & 66.6\% & 76.1\% & 50.7\% & 76.4\% & 73.5\% \\ \midrule
\multirow{5}{*}{\texttt{LLM.int4()}} & \multirow{5}{*}{W8A8} & 125M & {41.5\%} & {65.0\%} & {34.3\%} & {62.1\%} & {49.5\%} & 51.1\% & 50.8\% \\
 &  & 350M & {41.6\%} & {68.0\%} & {44.6\%} & {64.0\%} & {49.5\%} & 60.8\% & 55.8\% \\
 &  & 1.3B & {55.9\%} & {78.0\%} & {54.5\%} & {70.0\%} & {51.7\%} & 76.6\% & 67.0\% \\
 &  & 2.7B & {58.7\%} & {77.0\%} & {62.2\%} & {73.7\%} & {51.3\%} & 50.8\% & 64.5\% \\
 &  & 6.7B & {64.5\%} & {81.0\%} & {66.2\%} & {74.7\%} & {51.6\%} & 76.4\% & 72.5\% \\ \midrule
\multirow{5}{*}{SmoothQuant-c} & \multirow{5}{*}{W8A8} & 125M & {-} & {-} & {-} & {-} & {-} & - & - \\
 &  & 350M & {-} & {-} & {-} & {-} & {-} & - & - \\
 &  & 1.3B & 55.8\% & 78.0\% & 55.3\% & 71.2\% & 51.1\% & 75.9\% & 67.2\% \\
 &  & 2.7B & 60.4\% & 77.0\% & 64.0\% & 72.6\% & 51.3\% & 51.7\% & 65.2\% \\
 &  & 6.7B & 65.3\% & 81.0\% & 68.5\% & 74.7\% & 51.1\% & 71.6\% & 72.2\% \\ \midrule
\multirow{5}{*}{BFP} & \multirow{5}{*}{W6A6} & 125M & 42.6\% & 69.0\% & 37.1\% & 62.6\% & 49.4\% & 51.6\% & 52.6\% \\
 &  & 350M & 43.7\% & 72.0\% & 42.8\% & 65.1\% & 49.6\% & 64.6\% & 57.6\% \\
 &  & 1.3B & 57.0\% & 79.0\% & 51.8\% & 71.6\% & 51.5\% & 79.7\% & 67.8\% \\
 &  & 2.7B & 60.9\% & 76.0\% & 64.1\% & 73.3\% & 49.9\% & 53.1\% & 65.5\% \\
 &  & 6.7B & 65.2\% & 80.0\% & 67.2\% & 75.8\% & 50.9\% & 76.1\% & 72.9\% \\ \midrule
\multirow{5}{*}{BFP} & \multirow{5}{*}{W4A4} & 125M & 37.0\% & 65.0\% & 28.7\% & 58.9\% & 49.1\% & 49.5\% & 47.8\% \\
 &  & 350M & 39.9\% & 65.0\% & 38.7\% & 58.9\% & 49.3\% & 55.9\% & 51.7\% \\
 &  & 1.3B & 50.0\% & 71.0\% & 41.4\% & 65.7\% & 50.2\% & 58.1\% & 57.2\% \\
 &  & 2.7B & 52.4\% & 70.0\% & 36.2\% & 68.0\% & 51.4\% & 51.7\% & 55.7\% \\
 &  & 6.7B & 61.5\% & 84.0\% & 56.0\% & 73.1\% & 52.3\% & 61.1\% & 67.2\% \\ \midrule
\multirow{5}{*}{MiniFloat} & \multirow{5}{*}{W4A4} & 125M & 42.9\% & 66.0\% & 38.3\% & 62.7\% & 49.6\% & 50.8\% & 52.1\% \\
 &  & 350M & 44.0\% & 69.0\% & 44.9\% & 64.2\% & 49.8\% & 53.3\% & 55.1\% \\
 &  & 1.3B & 57.2\% & 80.0\% & 54.6\% & 71.4\% & 51.5\% & 60.2\% & 64.7\% \\
 &  & 2.7B & 59.9\% & 75.0\% & 63.4\% & 73.7\% & 49.8\% & 56.7\% & 65.7\% \\
 &  & 6.7B & 64.9\% & 82.0\% & 67.3\% & 76.0\% & 51.8\% & 62.2\% & 70.5\% \\ \bottomrule
\end{tabular}}
\caption{A complete comparison of LLM quantisation methods on downstream tasks in the zero-shot prompting setup. QNLI, MRPC, and COLA results are not included because even FP32 LLMs yield an accuracy close to random prediction.}
\label{tab:appendix:ptq_on_downstream_complete}
\end{table*}
\section{Estimate arithmetic density via logic synthesis}
\label{appendix:arith_density}
We implemented the hardware designs of the corresponding modules and measured their arithmetic density using hardware synthesis tools. Each design contains versions for Int8, Float32, MiniFloat and all the block arithmetic types above. All these designs are functionally verified in AMD Xilinx simulator on a set of test vectors. The hardware arithmetic density is obtained using the same formulation by Darvish Rouhani~\textit{et al.}~\cite{darvish2020pushing} with area in FPGAs. The area results were obtained from the post-Place \& Synthesis report in AMD Xilinx Vivado~\cite{vivado}. We estimate the total circuit area in LUTs, and a DSP is considered to be equivalent to 100 LUTs. We used the UltraScale+ family of FPGA devices for experiments, and the version of AMD Xilinx software is 2020.2. The arithmetic densities of various quantisation methods are present in \Cref{tab:appendix:arith_density}.

\section{PTQ on downstream tasks}
\label{appendix:ptq_on_downstream}
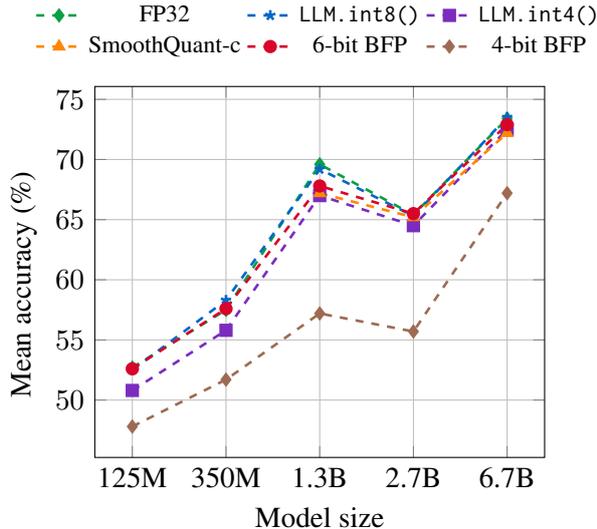
\begin{figure}[t]
\begin{tikzpicture}
\begin{axis}[
    height=65mm,
    width=75mm,
    grid=both,
    xlabel={Model size},
    ylabel={Mean accuracy (\%)},
    ytick={45, 50, 55, 60, 65, 70, 75},
    xtick={1, 2, 3, 4, 5},
    xticklabels={125M, 350M, 1.3B, 2.7B, 6.7B},
    legend columns = 3,
    y label style={at={(axis description cs:0.05,0.8)}, anchor=east},
    legend style={at={(0.5,1.05)},anchor=south, draw=none, fill=none, font=\footnotesize},
    ]

\addplot[dashed, czgreen, draw=czgreen, mark=diamond*, line width=1pt, mark options={solid, scale=1.0}] table
[x=model_idx, y=fp32] {
    model_idx	model_size	fp32
    1	125m	52.7
    2	350m	57.5
    3	1.3b	69.6
    4	2.7b	65.4
    5	6.7b	73.4
        };
\addlegendentry{FP32}

\addplot[dashed, czblue, draw=czblue, mark=star, line width=1pt, mark options={solid}] table
[x=model_idx, y=llm_int8] {
    model_idx	model_size	fp32	llm_int8
    1	125m	52.7	52.50
    2	350m	57.5	58.30
    3	1.3b	69.6	69.20
    4	2.7b	65.4	65.30
    5	6.7b	73.4	73.50
};
\addlegendentry{\texttt{LLM.int8()}}

\addplot[dashed, czpurple, draw=czpurple, mark=square*, line width=1pt, mark options={solid, scale=1.0}] table
[x=model_idx, y=llm_int4, col sep=space] {
model_idx	llm_int4
1	50.80
2	55.80
3	67.00
4	64.50
5	72.50
};
\addlegendentry{\texttt{LLM.int4()}}

\addplot[dashed, czorange, draw=czorange, mark=triangle*, line width=1pt, mark options={solid}] table
[x=model_idx, y=smoothquant, col sep=space] {
model_idx	smoothquant
3	67.20
4	65.20
5	72.20

};
\addlegendentry{SmoothQuant-c}

\addplot[dashed, czred, draw=czred, mark=*, line width=1pt, mark options={solid, scale=1.0}] table
[x=model_idx, y=bfp_6, col sep=space] {
model_idx	bfp_6
1	52.60
2	57.60
3	67.80
4	65.50
5	72.90
};
\label{fig:appendix:ptq_on_downstream:6bit_bfp}
\addlegendentry{6-bit BFP}

\addplot[dashed, czbrown, draw=czbrown, mark=diamond*, line width=1pt, mark options={solid, scale=1.0}] table
[x=model_idx, y=bfp_4, col sep=space] {
model_idx	bfp_4
1	47.80
2	51.70
3	57.20
4	55.70
5	67.20

};
\label{fig:appendix:ptq_on_downstream:4bit_bfp}
\addlegendentry{4-bit BFP}

\end{axis}
\end{tikzpicture}
\caption{Mean accuracy ($\uparrow$, \%) of various quantisation methods on downstream tasks. We observe that 6-bit BFP align with FP32 as model size scales, above SmoothQuant-c and \texttt{LLM.int4()}, below FP32 and \texttt{LLM.int8()}. Note that 6-bit BFP (\ref{fig:appendix:ptq_on_downstream:6bit_bfp}) achieves the highest memory density and arithmetic density among these methods. 4-bit BFP (\ref{fig:appendix:ptq_on_downstream:4bit_bfp}) has a severe accuracy drop.}
\label{fig:appendix:ptq_on_downstream}
\end{figure}

We quantised the pre-trained model and apply it to the downstream tasks in the zero-shot prompting setup. \Cref{fig:appendix:ptq_on_downstream} depicts how the performance of quantised models scale with model sizes. Our 6-bit BFP align with FP32 at various model sizes. \Cref{tab:appendix:ptq_on_downstream_complete} presents the detailed accuracy of each task. Note that QNLI, MRPC, and COLA results are not included in this table because even FP32 LLMs yield an accuracy close to random prediction.

\begin{table*}[]
\centering
\resizebox{\textwidth}{!}{%
\begin{tabular}{lclcrrrr}
\toprule
Task & Fine-tuning style & Config & Model size & zero-shot prompting & epoch 0 & epoch 1 & epoch 2 \\ \midrule
\multirow{12}{*}{SST2} & \multirow{4}{*}{FP32} & W32A32 & 125m & 53.3\% & 91.2\% & 92.9\% & 92.6\% \\
 &  & W32A32 & 350m & 61.8\% & 92.3\% & 93.1\% & 93.4\% \\
 &  & W32A32 & 1.3b & 82.2\% & 93.9\% & 93.2\% & 94.0\% \\
 &  & W32A32 & 2.7b & 51.7\% & 94.6\% & 94.5\% & 94.7\% \\ \cline{2-8}
 & \multirow{4}{*}{PTQ on downstream} & W5A5 & 125m & 49.5\%  (-3.8\%) & 91.3\%  (0.1\%) & 91.7\%  (-1.2\%) & 92.0\%  (-0.6\%) \\
 &  & W5A5 & 350m & 55.9\%  (-5.9\%) & 92.7\%  (0.4\%) & 92.5\%  (-0.6\%) & 92.2\%  (-1.2\%) \\
 &  & W5A5 & 1.3b & 58.1\%  (-24.1\%) & 93.7\%  (-0.2\%) & 93.2\%  (0.0\%) & 93.6\%  (-0.4\%) \\
 &  & W5A5 & 2.7b & 51.7\%  (0.0\%) & 94.0\%  (-0.6\%) & 95.3\%  (0.8\%) & 94.5\%  (-0.2\%) \\ \cline{2-8}
 & \multirow{4}{*}{TAQ on downstream} & W5A5 & 125m & 49.5\%  (-3.8\%) & 92.0\%  (0.8\%) & 92.0\%  (-0.9\%) & 91.6\%  (-0.9\%) \\
 &  & W5A5 & 350m & 55.9\%  (-6.0\%) & 91.1\%  (-1.3\%) & 91.6\%  (-1.5\%) & 91.6\%  (-1.7\%) \\
 &  & W5A5 & 1.3b & 58.1\%  (-24.1\%) & 94.3\%  (0.3\%) & 94.2\%  (0.9\%) & 94.2\%  (0.1\%) \\
 &  & W5A5 & 2.7b & 51.7\%  (0.0\%) & 94.6\%  (0.0\%) & 95.0\%  (0.4\%) & 94.7\%  (0.0\%) \\ \midrule
\multirow{12}{*}{QNLI} & \multirow{4}{*}{FP32} & W32A32 & 125m & 49.4\% & 87.8\% & 88.3\% & 88.7\% \\
 &  & W32A32 & 350m & 45.9\% & 86.5\% & 88.5\% & 89.1\% \\
 &  & W32A32 & 1.3b & 51.3\% & 89.0\% & 90.6\% & 91.7\% \\
 &  & W32A32 & 2.7b & 51.1\% & 61.2\% & 73.8\% & 85.3\% \\ \cline{2-8}
 & \multirow{4}{*}{PTQ on downstream} & W5A5 & 125m & 49.1\%  (-0.3\%) & 82.0\%  (-5.8\%) & 80.8\%  (-7.5\%) & 85.7\%  (-2.0\%) \\
 &  & W5A5 & 350m & 49.3\%  (-0.2\%) & 85.5\%  (-0.9\%) & 87.3\%  (-1.2\%) & 88.2\%  (-0.9\%) \\
 &  & W5A5 & 1.3b & 50.2\%  (-1.1\%) & 88.6\%  (-0.5\%) & 89.1\%  (-1.5\%) & 90.3\%  (-1.4\%) \\
 &  & W5A5 & 2.7b & 51.4\%  (0.2\%) & 60.9\%  (-0.3\%) & 71.6\%  (2.2\%) & 84.2\%  (-1.1\%) \\ \cline{2-8}
 & \multirow{4}{*}{TAQ on downstream} & W5A5 & 125m & 49.1\%  (-0.3\%) & 86.1\%  (-1.7\%) & 87.4\%  (-0.9\%) & 88.2\%  (-0.5\%) \\
 &  & W5A5 & 350m & 49.3\%  (-0.2\%) & 85.5\%  (-1.0\%) & 88.3\%  (-0.1\%) & 88.6\%  (-0.5\%) \\
 &  & W5A5 & 1.3b & 50.2\%  (-1.1\%) & 86.6\%  (-2.5\%) & 89.5\%  (-1.1\%) & 91.1\%  (0.7\%) \\
 &  & W5A5 & 2.7b & 51.4\%  (0.2\%) & 86.2\%  (25.0\%) & 88.1\%  (14.3\%) & 92.5\%  (4.3\%) \\ \midrule
\multirow{12}{*}{COLA$^{\dag}$} & \multirow{4}{*}{FP32} & W32A32 & 125m & 0.0\% & 41.2\% & 47.3\% & 49.8\% \\
 &  & W32A32 & 350m & 0.0\% & 13.0\% & 39.8\% & 47.2\% \\
 &  & W32A32 & 1.3b & -6.9\% & 29.1\% & 56.3\% & 56.9\% \\
 &  & W32A32 & 2.7b & -3.5\% & 0.0\% & 8.0\% & 25.9\% \\ \cline{2-8}
 & \multirow{4}{*}{PTQ on downstream} & W5A5 & 125m & -1.1\%  (-1.1\%) & 37.9\%  (-3.3\%) & 40.4\%  (-6.9\%) & 49.3\%  (-0.5\%) \\
 &  & W5A5 & 350m & 0.0\%  (0.0\%) & 7.5\%  (-5.5\%) & 32.8\%  (-7.0\%) & 46.0\%  (-1.2\%) \\
 &  & W5A5 & 1.3b & -1.6\%  (5.3\%) & 21.0\%  (-8.1\%) & 51.9\%  (-4.4\%) & 55.8\%  (-1.1\%) \\
 &  & W5A5 & 2.7b & -3.1\%  (0.0\%) & 0.0\%  (0.0\%) & -2.1\%  (-10.1\%) & 26.0\%  (0.1\%) \\ \cline{2-8}
 & \multirow{4}{*}{TAQ on downstream} & W5A5 & 125m & -1.1\%  (-1.1\%) & 43.1\%  (1.9\%) & 41.1\%  (0.7\%) & 43.7\%  (-5.6\%) \\
 &  & W5A5 & 350m & 0.0\%  (0.0\%) & 12.7\%  (-0.3\%) & 31.5\%  (-1.3\%) & 42.1\%  (-3.9\%) \\
 &  & W5A5 & 1.3b & -1.6\%  (5.3\%) & 33.9\%  (4.8\%) & 49.0\%  (-2.9\%) & 54.6\%  (-1.2\%) \\
 &  & W5A5 & 2.7b & -3.1\%  (0.4\%) & 0.7\%  (0.7\%) & 0.0\%  (2.1\%) & 18.6\%  (-7.4\%) \\ \midrule
\multirow{12}{*}{MRPC} & \multirow{4}{*}{FP32} & W32A32 & 125m & 68.4\% & 70.8\% & 76.8\% & 79.7\% \\
 &  & W32A32 & 350m & 68.4\% & 68.4\% & 69.6\% & 70.6\% \\
 &  & W32A32 & 1.3b & 66.4\% & 69.6\% & 70.8\% & 66.7\% \\
 &  & W32A32 & 2.7b & 67.9\% & 70.1\% & 81.6\% & 82.6\% \\ \cline{2-8}
 & \multirow{4}{*}{PTQ on downstream} & W5A5 & 125m & 68.4\%  (0.0\%) & 71.3\%  (0.5\%) & 74.0\%  (-2.8\%) & 78.8\%  (-0.9\%) \\
 &  & W5A5 & 350m & 68.4\%  (0.0\%) & 68.4\%  (0.0\%) & 69.6\%  (0.0\%) & 69.4\%  (-1.2\%) \\
 &  & W5A5 & 1.3b & 57.8\%  (-8.6\%) & 69.1\%  (-0.5\%) & 70.1\%  (-0.7\%) & 67.6\%  (0.9\%) \\
 &  & W5A5 & 2.7b & 60.3\%  (-7.6\%) & 68.6\%  (-1.5\%) & 79.9\%  (-1.7\%) & 78.9\%  (-3.7\%) \\ \cline{2-8}
 & \multirow{4}{*}{TAQ on downstream} & W5A5 & 125m & 68.4\%  (0.0\%) & 70.3\%  (-0.5\%) & 71.6\%  (-5.2\%) & 77.8\%  (-1.9\%) \\
 &  & W5A5 & 350m & 68.4\%  (0.0\%) & 69.4\%  (1.0\%) & 69.7\%  (0.1\%) & 70.6\%  (0.0\%) \\
 &  & W5A5 & 1.3b & 57.8\%  (-8.6\%) & 68.9\%  (-0.7\%) & 71.1\%  (0.3\%) & 71.1\%  (4.4\%) \\
 &  & W5A5 & 2.7b & 60.3\%  (-7.6\%) & 68.4\%  (-1.7\%) & 68.4\%  (-13.2\%) & 81.4\%  (-1.2\%) \\ \bottomrule
\end{tabular}}
\caption{The comparison between PTQ fine-tuned FP32 and TAQ on SST2, QNLI, COLA, and MRPC. Both cases align 4-bit BFP LLMs with FP32 after fine-tuning. The latter may achieve slightly better accuracy. $^{\dag}$ means COLA is evaluated using the Matthews Correlation Coefficient (MCC), while the other tasks are evaluated using accuracy.}
\label{tab:appendix:ptq_vs_taq_complete}
\end{table*}

\section{PTQ on fine-tuned FP32 vs TAQ on downstream tasks}
\label{appendix:ptq_vs_taq}

\Cref{tab:appendix:ptq_vs_taq_complete} compares the two options on four downstream tasks (QNLI, SST2, COLA, MRPC), that FP32 LLM cannot handle. We observe that both align 4-bit BFP LLMs' performance with FP32 on downstream tasks.

\section{Searched mixed-precision LLMs}
\label{appendix:mixed_precision_search}

\begin{figure}[t]
    \centering
    \begin{subfigure}[b]{0.49\textwidth}
         \begin{tikzpicture}
\pgfplotsset{every tick label/.append style={font=\footnotesize}}
\begin{axis}[
    width=\textwidth,
    ybar stacked, height=4cm,
    legend style={at={(1.0,1.0)},anchor=south east, draw=none},
    ymin=0, ymax=1,
    ytick={0,0.5,1},
    yticklabels={0,0.5, 1},
    xmin=0.3, xmax=32.7,
    ylabel={\footnotesize{Percentage of occurrence}},
    xlabel={\footnotesize{Layer ID}},
    legend style={font=\footnotesize},
    legend columns = 4,
    bar width=4,
    ]

\addplot[czbluel, draw=czbluel, fill=czbluel] table
[x=layers, y=freq4, col sep=space]
{
layers	freq4	freq5	freq6	freq7
1	0.784615385	0.215384615	0	0
2	0.723076923	0.276923077	0	0
3	0.4	0.292307692	0	0.307692308
4	0.753846154	0.246153846	0	0
5	0.584615385	0.415384615	0	0
6	0.707692308	0.292307692	0	0
7	0.061538462	0.246153846	0.353846154	0.338461538
8	0.107692308	0.276923077	0.307692308	0.307692308
9	0.661538462	0.338461538	0	0
10	0.676923077	0.323076923	0	0
11	0.476923077	0.246153846	0.276923077	0
12	0.415384615	0.261538462	0	0.323076923
13	0.415384615	0.230769231	0	0.353846154
14	0.661538462	0.338461538	0	0
15	0.753846154	0.246153846	0	0
16	0.815384615	0.184615385	0	0
17	0.4	0.307692308	0	0.292307692
18	0.646153846	0.353846154	0	0
19	0.492307692	0.230769231	0	0.276923077
20	0.476923077	0.230769231	0	0.292307692
21	0.461538462	0.184615385	0.353846154	0
22	0.476923077	0.2	0.323076923	0
23	0.430769231	0.138461538	0	0.430769231
24	0.723076923	0.276923077	0	0
25	0.753846154	0.246153846	0	0
26	0.707692308	0.292307692	0	0
27	0.461538462	0.215384615	0	0.323076923
28	0.461538462	0.246153846	0.292307692	0
29	0.446153846	0.276923077	0	0.276923077
30	0.692307692	0.307692308	0	0
31	0.476923077	0.215384615	0	0.307692308
32	0.461538462	0.246153846	0	0.292307692
};
\addlegendentry{$\leq$ 4-bit}

\addplot[czgreenl, draw=czgreenl, fill=czgreenl] table
[x=layers, y=freq5, col sep=space] {
layers	freq4	freq5	freq6	freq7
1	0.784615385	0.215384615	0	0
2	0.723076923	0.276923077	0	0
3	0.4	0.292307692	0	0.307692308
4	0.753846154	0.246153846	0	0
5	0.584615385	0.415384615	0	0
6	0.707692308	0.292307692	0	0
7	0.061538462	0.246153846	0.353846154	0.338461538
8	0.107692308	0.276923077	0.307692308	0.307692308
9	0.661538462	0.338461538	0	0
10	0.676923077	0.323076923	0	0
11	0.476923077	0.246153846	0.276923077	0
12	0.415384615	0.261538462	0	0.323076923
13	0.415384615	0.230769231	0	0.353846154
14	0.661538462	0.338461538	0	0
15	0.753846154	0.246153846	0	0
16	0.815384615	0.184615385	0	0
17	0.4	0.307692308	0	0.292307692
18	0.646153846	0.353846154	0	0
19	0.492307692	0.230769231	0	0.276923077
20	0.476923077	0.230769231	0	0.292307692
21	0.461538462	0.184615385	0.353846154	0
22	0.476923077	0.2	0.323076923	0
23	0.430769231	0.138461538	0	0.430769231
24	0.723076923	0.276923077	0	0
25	0.753846154	0.246153846	0	0
26	0.707692308	0.292307692	0	0
27	0.461538462	0.215384615	0	0.323076923
28	0.461538462	0.246153846	0.292307692	0
29	0.446153846	0.276923077	0	0.276923077
30	0.692307692	0.307692308	0	0
31	0.476923077	0.215384615	0	0.307692308
32	0.461538462	0.246153846	0	0.292307692
};
\addlegendentry{5-bit}

\addplot[czorangel, draw=czorangel, fill=czorangel] table
[x=layers, y=freq6, col sep=space] {
layers	freq4	freq5	freq6	freq7
1	0.784615385	0.215384615	0	0
2	0.723076923	0.276923077	0	0
3	0.4	0.292307692	0	0.307692308
4	0.753846154	0.246153846	0	0
5	0.584615385	0.415384615	0	0
6	0.707692308	0.292307692	0	0
7	0.061538462	0.246153846	0.353846154	0.338461538
8	0.107692308	0.276923077	0.307692308	0.307692308
9	0.661538462	0.338461538	0	0
10	0.676923077	0.323076923	0	0
11	0.476923077	0.246153846	0.276923077	0
12	0.415384615	0.261538462	0	0.323076923
13	0.415384615	0.230769231	0	0.353846154
14	0.661538462	0.338461538	0	0
15	0.753846154	0.246153846	0	0
16	0.815384615	0.184615385	0	0
17	0.4	0.307692308	0	0.292307692
18	0.646153846	0.353846154	0	0
19	0.492307692	0.230769231	0	0.276923077
20	0.476923077	0.230769231	0	0.292307692
21	0.461538462	0.184615385	0.353846154	0
22	0.476923077	0.2	0.323076923	0
23	0.430769231	0.138461538	0	0.430769231
24	0.723076923	0.276923077	0	0
25	0.753846154	0.246153846	0	0
26	0.707692308	0.292307692	0	0
27	0.461538462	0.215384615	0	0.323076923
28	0.461538462	0.246153846	0.292307692	0
29	0.446153846	0.276923077	0	0.276923077
30	0.692307692	0.307692308	0	0
31	0.476923077	0.215384615	0	0.307692308
32	0.461538462	0.246153846	0	0.292307692
};
\addlegendentry{6-bit}

\addplot[czredl, draw=czredl, fill=czredl] table
[x=layers, y=freq7, col sep=space] {
layers	freq4	freq5	freq6	freq7
1	0.784615385	0.215384615	0	0
2	0.723076923	0.276923077	0	0
3	0.4	0.292307692	0	0.307692308
4	0.753846154	0.246153846	0	0
5	0.584615385	0.415384615	0	0
6	0.707692308	0.292307692	0	0
7	0.061538462	0.246153846	0.353846154	0.338461538
8	0.107692308	0.276923077	0.307692308	0.307692308
9	0.661538462	0.338461538	0	0
10	0.676923077	0.323076923	0	0
11	0.476923077	0.246153846	0.276923077	0
12	0.415384615	0.261538462	0	0.323076923
13	0.415384615	0.230769231	0	0.353846154
14	0.661538462	0.338461538	0	0
15	0.753846154	0.246153846	0	0
16	0.815384615	0.184615385	0	0
17	0.4	0.307692308	0	0.292307692
18	0.646153846	0.353846154	0	0
19	0.492307692	0.230769231	0	0.276923077
20	0.476923077	0.230769231	0	0.292307692
21	0.461538462	0.184615385	0.353846154	0
22	0.476923077	0.2	0.323076923	0
23	0.430769231	0.138461538	0	0.430769231
24	0.723076923	0.276923077	0	0
25	0.753846154	0.246153846	0	0
26	0.707692308	0.292307692	0	0
27	0.461538462	0.215384615	0	0.323076923
28	0.461538462	0.246153846	0.292307692	0
29	0.446153846	0.276923077	0	0.276923077
30	0.692307692	0.307692308	0	0
31	0.476923077	0.215384615	0	0.307692308
32	0.461538462	0.246153846	0	0.292307692
};
\addlegendentry{7-bit}
\end{axis}
\end{tikzpicture}
         \caption{$\bf{X_n}$ in Line 3, \Cref{alg:transformer}. }
    \end{subfigure}
    \begin{subfigure}[b]{0.49\textwidth}
        \begin{tikzpicture}
\pgfplotsset{every tick label/.append style={font=\footnotesize}}
\begin{axis}[
    width=\textwidth,
    ybar stacked, height=4cm,
    legend style={at={(1.0,1.0)},anchor=south east, draw=none},
    ymin=0, ymax=1,
    ytick={0,0.5,1},
    yticklabels={0,0.5, 1},
    xmin=0.3, xmax=32.7,
    ylabel={\footnotesize{Percentage of occurrence}},
    xlabel={\footnotesize{Layer ID}},
    legend style={font=\footnotesize},
    legend columns = 4,
    bar width=4,
    ]

\addplot[czbluel, draw=czbluel, fill=czbluel] table
[x=layers, y=freq4, col sep=space]
{
tensor_name	layers	freq4	freq5	freq6	freq7	high
x_k_proj	1	0.815384615	0.184615385	0	0	0
x_k_proj	2	0.707692308	0.292307692	0	0	0
x_k_proj	3	0.661538462	0.338461538	0	0	0
x_k_proj	4	0.538461538	0.169230769	0.292307692	0	0.292307692
x_k_proj	5	0.8	0.2	0	0	0
x_k_proj	6	0.507692308	0.184615385	0.307692308	0	0.307692308
x_k_proj	7	0.369230769	0.307692308	0.323076923	0	0.323076923
x_k_proj	8	0.723076923	0.276923077	0	0	0
x_k_proj	9	0.446153846	0.2	0.353846154	0	0.353846154
x_k_proj	10	0.153846154	0.2	0.292307692	0.353846154	0.646153846
x_k_proj	11	0.4	0.292307692	0	0.307692308	0.307692308
x_k_proj	12	0.4	0.246153846	0	0.353846154	0.353846154
x_k_proj	13	0.430769231	0.292307692	0.276923077	0	0.276923077
x_k_proj	14	0.476923077	0.230769231	0	0.292307692	0.292307692
x_k_proj	15	0.538461538	0.092307692	0	0.369230769	0.369230769
x_k_proj	16	0.507692308	0.169230769	0	0.323076923	0.323076923
x_k_proj	17	0.184615385	0.246153846	0.276923077	0.292307692	0.569230769
x_k_proj	18	0.476923077	0.215384615	0	0.307692308	0.307692308
x_k_proj	19	0.4	0.276923077	0.323076923	0	0.323076923
x_k_proj	20	0.369230769	0.338461538	0.292307692	0	0.292307692
x_k_proj	21	0.523076923	0.2	0	0.276923077	0.276923077
x_k_proj	22	0.415384615	0.230769231	0	0.353846154	0.353846154
x_k_proj	23	0.692307692	0.307692308	0	0	0
x_k_proj	24	0.384615385	0.246153846	0	0.369230769	0.369230769
x_k_proj	25	0.523076923	0.2	0	0.276923077	0.276923077
x_k_proj	26	0.738461538	0.261538462	0	0	0
x_k_proj	27	0.584615385	0.415384615	0	0	0
x_k_proj	28	0.430769231	0.215384615	0.353846154	0	0.353846154
x_k_proj	29	0.415384615	0.261538462	0.323076923	0	0.323076923
x_k_proj	30	0.430769231	0.261538462	0	0.307692308	0.307692308
x_k_proj	31	0.184615385	0.138461538	0.4	0.276923077	0.676923077
x_k_proj	32	0.4	0.276923077	0.323076923	0	0.323076923

};
\addlegendentry{$\leq$ 4-bit}

\addplot[czgreenl, draw=czgreenl, fill=czgreenl] table
[x=layers, y=freq5, col sep=space] {
tensor_name	layers	freq4	freq5	freq6	freq7	high
x_k_proj	1	0.815384615	0.184615385	0	0	0
x_k_proj	2	0.707692308	0.292307692	0	0	0
x_k_proj	3	0.661538462	0.338461538	0	0	0
x_k_proj	4	0.538461538	0.169230769	0.292307692	0	0.292307692
x_k_proj	5	0.8	0.2	0	0	0
x_k_proj	6	0.507692308	0.184615385	0.307692308	0	0.307692308
x_k_proj	7	0.369230769	0.307692308	0.323076923	0	0.323076923
x_k_proj	8	0.723076923	0.276923077	0	0	0
x_k_proj	9	0.446153846	0.2	0.353846154	0	0.353846154
x_k_proj	10	0.153846154	0.2	0.292307692	0.353846154	0.646153846
x_k_proj	11	0.4	0.292307692	0	0.307692308	0.307692308
x_k_proj	12	0.4	0.246153846	0	0.353846154	0.353846154
x_k_proj	13	0.430769231	0.292307692	0.276923077	0	0.276923077
x_k_proj	14	0.476923077	0.230769231	0	0.292307692	0.292307692
x_k_proj	15	0.538461538	0.092307692	0	0.369230769	0.369230769
x_k_proj	16	0.507692308	0.169230769	0	0.323076923	0.323076923
x_k_proj	17	0.184615385	0.246153846	0.276923077	0.292307692	0.569230769
x_k_proj	18	0.476923077	0.215384615	0	0.307692308	0.307692308
x_k_proj	19	0.4	0.276923077	0.323076923	0	0.323076923
x_k_proj	20	0.369230769	0.338461538	0.292307692	0	0.292307692
x_k_proj	21	0.523076923	0.2	0	0.276923077	0.276923077
x_k_proj	22	0.415384615	0.230769231	0	0.353846154	0.353846154
x_k_proj	23	0.692307692	0.307692308	0	0	0
x_k_proj	24	0.384615385	0.246153846	0	0.369230769	0.369230769
x_k_proj	25	0.523076923	0.2	0	0.276923077	0.276923077
x_k_proj	26	0.738461538	0.261538462	0	0	0
x_k_proj	27	0.584615385	0.415384615	0	0	0
x_k_proj	28	0.430769231	0.215384615	0.353846154	0	0.353846154
x_k_proj	29	0.415384615	0.261538462	0.323076923	0	0.323076923
x_k_proj	30	0.430769231	0.261538462	0	0.307692308	0.307692308
x_k_proj	31	0.184615385	0.138461538	0.4	0.276923077	0.676923077
x_k_proj	32	0.4	0.276923077	0.323076923	0	0.323076923

};
\addlegendentry{5-bit}

\addplot[czorangel, draw=czorangel, fill=czorangel] table
[x=layers, y=freq6, col sep=space] {
tensor_name	layers	freq4	freq5	freq6	freq7	high
x_k_proj	1	0.815384615	0.184615385	0	0	0
x_k_proj	2	0.707692308	0.292307692	0	0	0
x_k_proj	3	0.661538462	0.338461538	0	0	0
x_k_proj	4	0.538461538	0.169230769	0.292307692	0	0.292307692
x_k_proj	5	0.8	0.2	0	0	0
x_k_proj	6	0.507692308	0.184615385	0.307692308	0	0.307692308
x_k_proj	7	0.369230769	0.307692308	0.323076923	0	0.323076923
x_k_proj	8	0.723076923	0.276923077	0	0	0
x_k_proj	9	0.446153846	0.2	0.353846154	0	0.353846154
x_k_proj	10	0.153846154	0.2	0.292307692	0.353846154	0.646153846
x_k_proj	11	0.4	0.292307692	0	0.307692308	0.307692308
x_k_proj	12	0.4	0.246153846	0	0.353846154	0.353846154
x_k_proj	13	0.430769231	0.292307692	0.276923077	0	0.276923077
x_k_proj	14	0.476923077	0.230769231	0	0.292307692	0.292307692
x_k_proj	15	0.538461538	0.092307692	0	0.369230769	0.369230769
x_k_proj	16	0.507692308	0.169230769	0	0.323076923	0.323076923
x_k_proj	17	0.184615385	0.246153846	0.276923077	0.292307692	0.569230769
x_k_proj	18	0.476923077	0.215384615	0	0.307692308	0.307692308
x_k_proj	19	0.4	0.276923077	0.323076923	0	0.323076923
x_k_proj	20	0.369230769	0.338461538	0.292307692	0	0.292307692
x_k_proj	21	0.523076923	0.2	0	0.276923077	0.276923077
x_k_proj	22	0.415384615	0.230769231	0	0.353846154	0.353846154
x_k_proj	23	0.692307692	0.307692308	0	0	0
x_k_proj	24	0.384615385	0.246153846	0	0.369230769	0.369230769
x_k_proj	25	0.523076923	0.2	0	0.276923077	0.276923077
x_k_proj	26	0.738461538	0.261538462	0	0	0
x_k_proj	27	0.584615385	0.415384615	0	0	0
x_k_proj	28	0.430769231	0.215384615	0.353846154	0	0.353846154
x_k_proj	29	0.415384615	0.261538462	0.323076923	0	0.323076923
x_k_proj	30	0.430769231	0.261538462	0	0.307692308	0.307692308
x_k_proj	31	0.184615385	0.138461538	0.4	0.276923077	0.676923077
x_k_proj	32	0.4	0.276923077	0.323076923	0	0.323076923

};
\addlegendentry{6-bit}

\addplot[czredl, draw=czredl, fill=czredl] table
[x=layers, y=freq7, col sep=space] {
tensor_name	layers	freq4	freq5	freq6	freq7	high
x_k_proj	1	0.815384615	0.184615385	0	0	0
x_k_proj	2	0.707692308	0.292307692	0	0	0
x_k_proj	3	0.661538462	0.338461538	0	0	0
x_k_proj	4	0.538461538	0.169230769	0.292307692	0	0.292307692
x_k_proj	5	0.8	0.2	0	0	0
x_k_proj	6	0.507692308	0.184615385	0.307692308	0	0.307692308
x_k_proj	7	0.369230769	0.307692308	0.323076923	0	0.323076923
x_k_proj	8	0.723076923	0.276923077	0	0	0
x_k_proj	9	0.446153846	0.2	0.353846154	0	0.353846154
x_k_proj	10	0.153846154	0.2	0.292307692	0.353846154	0.646153846
x_k_proj	11	0.4	0.292307692	0	0.307692308	0.307692308
x_k_proj	12	0.4	0.246153846	0	0.353846154	0.353846154
x_k_proj	13	0.430769231	0.292307692	0.276923077	0	0.276923077
x_k_proj	14	0.476923077	0.230769231	0	0.292307692	0.292307692
x_k_proj	15	0.538461538	0.092307692	0	0.369230769	0.369230769
x_k_proj	16	0.507692308	0.169230769	0	0.323076923	0.323076923
x_k_proj	17	0.184615385	0.246153846	0.276923077	0.292307692	0.569230769
x_k_proj	18	0.476923077	0.215384615	0	0.307692308	0.307692308
x_k_proj	19	0.4	0.276923077	0.323076923	0	0.323076923
x_k_proj	20	0.369230769	0.338461538	0.292307692	0	0.292307692
x_k_proj	21	0.523076923	0.2	0	0.276923077	0.276923077
x_k_proj	22	0.415384615	0.230769231	0	0.353846154	0.353846154
x_k_proj	23	0.692307692	0.307692308	0	0	0
x_k_proj	24	0.384615385	0.246153846	0	0.369230769	0.369230769
x_k_proj	25	0.523076923	0.2	0	0.276923077	0.276923077
x_k_proj	26	0.738461538	0.261538462	0	0	0
x_k_proj	27	0.584615385	0.415384615	0	0	0
x_k_proj	28	0.430769231	0.215384615	0.353846154	0	0.353846154
x_k_proj	29	0.415384615	0.261538462	0.323076923	0	0.323076923
x_k_proj	30	0.430769231	0.261538462	0	0.307692308	0.307692308
x_k_proj	31	0.184615385	0.138461538	0.4	0.276923077	0.676923077
x_k_proj	32	0.4	0.276923077	0.323076923	0	0.323076923

};
\addlegendentry{7-bit}
\end{axis}
\end{tikzpicture}
        \caption{$\bf{X_n}$ in Line 4, \Cref{alg:transformer}. }
    \end{subfigure}
    \begin{subfigure}[b]{0.49\textwidth}
        \begin{tikzpicture}
\pgfplotsset{every tick label/.append style={font=\footnotesize}}
\begin{axis}[
    width=\textwidth,
    ybar stacked, height=4cm,
    legend style={at={(1.0,1.0)},anchor=south east, draw=none},
    ymin=0, ymax=1,
    ytick={0,0.5,1},
    yticklabels={0,0.5, 1},
    xmin=0.3, xmax=32.7,
    ylabel={\footnotesize{Percentage of occurrence}},
    xlabel={\footnotesize{Layer ID}},
    legend style={font=\footnotesize},
    legend columns = 4,
    bar width=4,
    ]

\addplot[czbluel, draw=czbluel, fill=czbluel] table
[x=layers, y=freq4, col sep=space]
{
tensor_name	layers	freq4	freq5	freq6	freq7	high
x_v_proj	1	0.2	0.215384615	0.307692308	0.276923077	0.584615385
x_v_proj	2	0.707692308	0.292307692	0	0	0
x_v_proj	3	0.523076923	0.2	0.276923077	0	0.276923077
x_v_proj	4	0.769230769	0.230769231	0	0	0
x_v_proj	5	0.430769231	0.276923077	0.292307692	0	0.292307692
x_v_proj	6	0.661538462	0.338461538	0	0	0
x_v_proj	7	0.723076923	0.276923077	0	0	0
x_v_proj	8	0.692307692	0.307692308	0	0	0
x_v_proj	9	0.753846154	0.246153846	0	0	0
x_v_proj	10	0.446153846	0.276923077	0	0.276923077	0.276923077
x_v_proj	11	0.476923077	0.230769231	0	0.292307692	0.292307692
x_v_proj	12	0.569230769	0.138461538	0	0.292307692	0.292307692
x_v_proj	13	0.476923077	0.246153846	0.276923077	0	0.276923077
x_v_proj	14	0.676923077	0.323076923	0	0	0
x_v_proj	15	0.215384615	0.230769231	0.276923077	0.276923077	0.553846154
x_v_proj	16	0.707692308	0.292307692	0	0	0
x_v_proj	17	0.169230769	0.215384615	0.307692308	0.307692308	0.615384615
x_v_proj	18	0.707692308	0.292307692	0	0	0
x_v_proj	19	0.738461538	0.261538462	0	0	0
x_v_proj	20	0.323076923	0.369230769	0.307692308	0	0.307692308
x_v_proj	21	0.661538462	0.338461538	0	0	0
x_v_proj	22	0.707692308	0.292307692	0	0	0
x_v_proj	23	0.492307692	0.230769231	0	0.276923077	0.276923077
x_v_proj	24	0.446153846	0.276923077	0	0.276923077	0.276923077
x_v_proj	25	0.476923077	0.246153846	0.276923077	0	0.276923077
x_v_proj	26	0.815384615	0.184615385	0	0	0
x_v_proj	27	0.738461538	0.261538462	0	0	0
x_v_proj	28	0.8	0.2	0	0	0
x_v_proj	29	0.753846154	0.246153846	0	0	0
x_v_proj	30	0.476923077	0.2	0	0.323076923	0.323076923
x_v_proj	31	0.707692308	0.292307692	0	0	0
x_v_proj	32	0.692307692	0.307692308	0	0	0

};
\addlegendentry{$\leq$ 4-bit}

\addplot[czgreenl, draw=czgreenl, fill=czgreenl] table
[x=layers, y=freq5, col sep=space] {
tensor_name	layers	freq4	freq5	freq6	freq7	high
x_v_proj	1	0.2	0.215384615	0.307692308	0.276923077	0.584615385
x_v_proj	2	0.707692308	0.292307692	0	0	0
x_v_proj	3	0.523076923	0.2	0.276923077	0	0.276923077
x_v_proj	4	0.769230769	0.230769231	0	0	0
x_v_proj	5	0.430769231	0.276923077	0.292307692	0	0.292307692
x_v_proj	6	0.661538462	0.338461538	0	0	0
x_v_proj	7	0.723076923	0.276923077	0	0	0
x_v_proj	8	0.692307692	0.307692308	0	0	0
x_v_proj	9	0.753846154	0.246153846	0	0	0
x_v_proj	10	0.446153846	0.276923077	0	0.276923077	0.276923077
x_v_proj	11	0.476923077	0.230769231	0	0.292307692	0.292307692
x_v_proj	12	0.569230769	0.138461538	0	0.292307692	0.292307692
x_v_proj	13	0.476923077	0.246153846	0.276923077	0	0.276923077
x_v_proj	14	0.676923077	0.323076923	0	0	0
x_v_proj	15	0.215384615	0.230769231	0.276923077	0.276923077	0.553846154
x_v_proj	16	0.707692308	0.292307692	0	0	0
x_v_proj	17	0.169230769	0.215384615	0.307692308	0.307692308	0.615384615
x_v_proj	18	0.707692308	0.292307692	0	0	0
x_v_proj	19	0.738461538	0.261538462	0	0	0
x_v_proj	20	0.323076923	0.369230769	0.307692308	0	0.307692308
x_v_proj	21	0.661538462	0.338461538	0	0	0
x_v_proj	22	0.707692308	0.292307692	0	0	0
x_v_proj	23	0.492307692	0.230769231	0	0.276923077	0.276923077
x_v_proj	24	0.446153846	0.276923077	0	0.276923077	0.276923077
x_v_proj	25	0.476923077	0.246153846	0.276923077	0	0.276923077
x_v_proj	26	0.815384615	0.184615385	0	0	0
x_v_proj	27	0.738461538	0.261538462	0	0	0
x_v_proj	28	0.8	0.2	0	0	0
x_v_proj	29	0.753846154	0.246153846	0	0	0
x_v_proj	30	0.476923077	0.2	0	0.323076923	0.323076923
x_v_proj	31	0.707692308	0.292307692	0	0	0
x_v_proj	32	0.692307692	0.307692308	0	0	0

};
\addlegendentry{5-bit}

\addplot[czorangel, draw=czorangel, fill=czorangel] table
[x=layers, y=freq6, col sep=space] {
tensor_name	layers	freq4	freq5	freq6	freq7	high
x_v_proj	1	0.2	0.215384615	0.307692308	0.276923077	0.584615385
x_v_proj	2	0.707692308	0.292307692	0	0	0
x_v_proj	3	0.523076923	0.2	0.276923077	0	0.276923077
x_v_proj	4	0.769230769	0.230769231	0	0	0
x_v_proj	5	0.430769231	0.276923077	0.292307692	0	0.292307692
x_v_proj	6	0.661538462	0.338461538	0	0	0
x_v_proj	7	0.723076923	0.276923077	0	0	0
x_v_proj	8	0.692307692	0.307692308	0	0	0
x_v_proj	9	0.753846154	0.246153846	0	0	0
x_v_proj	10	0.446153846	0.276923077	0	0.276923077	0.276923077
x_v_proj	11	0.476923077	0.230769231	0	0.292307692	0.292307692
x_v_proj	12	0.569230769	0.138461538	0	0.292307692	0.292307692
x_v_proj	13	0.476923077	0.246153846	0.276923077	0	0.276923077
x_v_proj	14	0.676923077	0.323076923	0	0	0
x_v_proj	15	0.215384615	0.230769231	0.276923077	0.276923077	0.553846154
x_v_proj	16	0.707692308	0.292307692	0	0	0
x_v_proj	17	0.169230769	0.215384615	0.307692308	0.307692308	0.615384615
x_v_proj	18	0.707692308	0.292307692	0	0	0
x_v_proj	19	0.738461538	0.261538462	0	0	0
x_v_proj	20	0.323076923	0.369230769	0.307692308	0	0.307692308
x_v_proj	21	0.661538462	0.338461538	0	0	0
x_v_proj	22	0.707692308	0.292307692	0	0	0
x_v_proj	23	0.492307692	0.230769231	0	0.276923077	0.276923077
x_v_proj	24	0.446153846	0.276923077	0	0.276923077	0.276923077
x_v_proj	25	0.476923077	0.246153846	0.276923077	0	0.276923077
x_v_proj	26	0.815384615	0.184615385	0	0	0
x_v_proj	27	0.738461538	0.261538462	0	0	0
x_v_proj	28	0.8	0.2	0	0	0
x_v_proj	29	0.753846154	0.246153846	0	0	0
x_v_proj	30	0.476923077	0.2	0	0.323076923	0.323076923
x_v_proj	31	0.707692308	0.292307692	0	0	0
x_v_proj	32	0.692307692	0.307692308	0	0	0

};
\addlegendentry{6-bit}

\addplot[czredl, draw=czredl, fill=czredl] table
[x=layers, y=freq7, col sep=space] {
tensor_name	layers	freq4	freq5	freq6	freq7	high
x_v_proj	1	0.2	0.215384615	0.307692308	0.276923077	0.584615385
x_v_proj	2	0.707692308	0.292307692	0	0	0
x_v_proj	3	0.523076923	0.2	0.276923077	0	0.276923077
x_v_proj	4	0.769230769	0.230769231	0	0	0
x_v_proj	5	0.430769231	0.276923077	0.292307692	0	0.292307692
x_v_proj	6	0.661538462	0.338461538	0	0	0
x_v_proj	7	0.723076923	0.276923077	0	0	0
x_v_proj	8	0.692307692	0.307692308	0	0	0
x_v_proj	9	0.753846154	0.246153846	0	0	0
x_v_proj	10	0.446153846	0.276923077	0	0.276923077	0.276923077
x_v_proj	11	0.476923077	0.230769231	0	0.292307692	0.292307692
x_v_proj	12	0.569230769	0.138461538	0	0.292307692	0.292307692
x_v_proj	13	0.476923077	0.246153846	0.276923077	0	0.276923077
x_v_proj	14	0.676923077	0.323076923	0	0	0
x_v_proj	15	0.215384615	0.230769231	0.276923077	0.276923077	0.553846154
x_v_proj	16	0.707692308	0.292307692	0	0	0
x_v_proj	17	0.169230769	0.215384615	0.307692308	0.307692308	0.615384615
x_v_proj	18	0.707692308	0.292307692	0	0	0
x_v_proj	19	0.738461538	0.261538462	0	0	0
x_v_proj	20	0.323076923	0.369230769	0.307692308	0	0.307692308
x_v_proj	21	0.661538462	0.338461538	0	0	0
x_v_proj	22	0.707692308	0.292307692	0	0	0
x_v_proj	23	0.492307692	0.230769231	0	0.276923077	0.276923077
x_v_proj	24	0.446153846	0.276923077	0	0.276923077	0.276923077
x_v_proj	25	0.476923077	0.246153846	0.276923077	0	0.276923077
x_v_proj	26	0.815384615	0.184615385	0	0	0
x_v_proj	27	0.738461538	0.261538462	0	0	0
x_v_proj	28	0.8	0.2	0	0	0
x_v_proj	29	0.753846154	0.246153846	0	0	0
x_v_proj	30	0.476923077	0.2	0	0.323076923	0.323076923
x_v_proj	31	0.707692308	0.292307692	0	0	0
x_v_proj	32	0.692307692	0.307692308	0	0	0

};
\addlegendentry{7-bit}
\end{axis}
\end{tikzpicture}
        \caption{$\bf{X_n}$ in Line 5, \Cref{alg:transformer}. }
    \end{subfigure}

    \begin{subfigure}[b]{0.49\textwidth}
        \begin{tikzpicture}
\pgfplotsset{every tick label/.append style={font=\footnotesize}}
\begin{axis}[
    width=\textwidth,
    ybar stacked, height=4cm,
    legend style={at={(1.0,1.0)},anchor=south east, draw=none},
    ymin=0, ymax=1,
    ytick={0,0.5,1},
    yticklabels={0,0.5, 1},
    xmin=0.3, xmax=32.7,
    ylabel={\footnotesize{Percentage of occurrence}},
    xlabel={\footnotesize{Layer ID}},
    legend style={font=\footnotesize},
    legend columns = 4,
    bar width=4,
    ]

\addplot[czbluel, draw=czbluel, fill=czbluel] table
[x=layers, y=freq4, col sep=space]
{
tensor_name	layers	freq4	freq5	freq6	freq7	high
x_bmm_k	1	0.815384615	0.184615385	0	0	0
x_bmm_k	2	0.769230769	0.230769231	0	0	0
x_bmm_k	3	0.446153846	0.261538462	0	0.292307692	0.292307692
x_bmm_k	4	0.415384615	0.246153846	0	0.338461538	0.338461538
x_bmm_k	5	0.476923077	0.246153846	0	0.276923077	0.276923077
x_bmm_k	6	0.338461538	0.307692308	0	0.353846154	0.353846154
x_bmm_k	7	0.446153846	0.261538462	0.292307692	0	0.292307692
x_bmm_k	8	0.692307692	0.307692308	0	0	0
x_bmm_k	9	0.707692308	0.292307692	0	0	0
x_bmm_k	10	0.753846154	0.246153846	0	0	0
x_bmm_k	11	0.723076923	0.276923077	0	0	0
x_bmm_k	12	0.430769231	0.2	0	0.369230769	0.369230769
x_bmm_k	13	0.476923077	0.246153846	0	0.276923077	0.276923077
x_bmm_k	14	0.476923077	0.246153846	0.276923077	0	0.276923077
x_bmm_k	15	0.769230769	0.230769231	0	0	0
x_bmm_k	16	0.292307692	0.323076923	0	0.384615385	0.384615385
x_bmm_k	17	0.430769231	0.276923077	0	0.292307692	0.292307692
x_bmm_k	18	0.692307692	0.307692308	0	0	0
x_bmm_k	19	0.430769231	0.276923077	0.292307692	0	0.292307692
x_bmm_k	20	0.476923077	0.215384615	0	0.307692308	0.307692308
x_bmm_k	21	0.753846154	0.246153846	0	0	0
x_bmm_k	22	0.723076923	0.276923077	0	0	0
x_bmm_k	23	0.415384615	0.261538462	0.323076923	0	0.323076923
x_bmm_k	24	0.430769231	0.276923077	0.292307692	0	0.292307692
x_bmm_k	25	0.415384615	0.276923077	0	0.307692308	0.307692308
x_bmm_k	26	0.753846154	0.246153846	0	0	0
x_bmm_k	27	0.738461538	0.261538462	0	0	0
x_bmm_k	28	0.676923077	0.323076923	0	0	0
x_bmm_k	29	0.723076923	0.276923077	0	0	0
x_bmm_k	30	0.753846154	0.246153846	0	0	0
x_bmm_k	31	0.476923077	0.230769231	0.292307692	0	0.292307692
x_bmm_k	32	0.384615385	0.323076923	0.292307692	0	0.292307692

};
\addlegendentry{$\leq$ 4-bit}

\addplot[czgreenl, draw=czgreenl, fill=czgreenl] table
[x=layers, y=freq5, col sep=space] {
tensor_name	layers	freq4	freq5	freq6	freq7	high
x_bmm_k	1	0.815384615	0.184615385	0	0	0
x_bmm_k	2	0.769230769	0.230769231	0	0	0
x_bmm_k	3	0.446153846	0.261538462	0	0.292307692	0.292307692
x_bmm_k	4	0.415384615	0.246153846	0	0.338461538	0.338461538
x_bmm_k	5	0.476923077	0.246153846	0	0.276923077	0.276923077
x_bmm_k	6	0.338461538	0.307692308	0	0.353846154	0.353846154
x_bmm_k	7	0.446153846	0.261538462	0.292307692	0	0.292307692
x_bmm_k	8	0.692307692	0.307692308	0	0	0
x_bmm_k	9	0.707692308	0.292307692	0	0	0
x_bmm_k	10	0.753846154	0.246153846	0	0	0
x_bmm_k	11	0.723076923	0.276923077	0	0	0
x_bmm_k	12	0.430769231	0.2	0	0.369230769	0.369230769
x_bmm_k	13	0.476923077	0.246153846	0	0.276923077	0.276923077
x_bmm_k	14	0.476923077	0.246153846	0.276923077	0	0.276923077
x_bmm_k	15	0.769230769	0.230769231	0	0	0
x_bmm_k	16	0.292307692	0.323076923	0	0.384615385	0.384615385
x_bmm_k	17	0.430769231	0.276923077	0	0.292307692	0.292307692
x_bmm_k	18	0.692307692	0.307692308	0	0	0
x_bmm_k	19	0.430769231	0.276923077	0.292307692	0	0.292307692
x_bmm_k	20	0.476923077	0.215384615	0	0.307692308	0.307692308
x_bmm_k	21	0.753846154	0.246153846	0	0	0
x_bmm_k	22	0.723076923	0.276923077	0	0	0
x_bmm_k	23	0.415384615	0.261538462	0.323076923	0	0.323076923
x_bmm_k	24	0.430769231	0.276923077	0.292307692	0	0.292307692
x_bmm_k	25	0.415384615	0.276923077	0	0.307692308	0.307692308
x_bmm_k	26	0.753846154	0.246153846	0	0	0
x_bmm_k	27	0.738461538	0.261538462	0	0	0
x_bmm_k	28	0.676923077	0.323076923	0	0	0
x_bmm_k	29	0.723076923	0.276923077	0	0	0
x_bmm_k	30	0.753846154	0.246153846	0	0	0
x_bmm_k	31	0.476923077	0.230769231	0.292307692	0	0.292307692
x_bmm_k	32	0.384615385	0.323076923	0.292307692	0	0.292307692

};
\addlegendentry{5-bit}

\addplot[czorangel, draw=czorangel, fill=czorangel] table
[x=layers, y=freq6, col sep=space] {
tensor_name	layers	freq4	freq5	freq6	freq7	high
x_bmm_k	1	0.815384615	0.184615385	0	0	0
x_bmm_k	2	0.769230769	0.230769231	0	0	0
x_bmm_k	3	0.446153846	0.261538462	0	0.292307692	0.292307692
x_bmm_k	4	0.415384615	0.246153846	0	0.338461538	0.338461538
x_bmm_k	5	0.476923077	0.246153846	0	0.276923077	0.276923077
x_bmm_k	6	0.338461538	0.307692308	0	0.353846154	0.353846154
x_bmm_k	7	0.446153846	0.261538462	0.292307692	0	0.292307692
x_bmm_k	8	0.692307692	0.307692308	0	0	0
x_bmm_k	9	0.707692308	0.292307692	0	0	0
x_bmm_k	10	0.753846154	0.246153846	0	0	0
x_bmm_k	11	0.723076923	0.276923077	0	0	0
x_bmm_k	12	0.430769231	0.2	0	0.369230769	0.369230769
x_bmm_k	13	0.476923077	0.246153846	0	0.276923077	0.276923077
x_bmm_k	14	0.476923077	0.246153846	0.276923077	0	0.276923077
x_bmm_k	15	0.769230769	0.230769231	0	0	0
x_bmm_k	16	0.292307692	0.323076923	0	0.384615385	0.384615385
x_bmm_k	17	0.430769231	0.276923077	0	0.292307692	0.292307692
x_bmm_k	18	0.692307692	0.307692308	0	0	0
x_bmm_k	19	0.430769231	0.276923077	0.292307692	0	0.292307692
x_bmm_k	20	0.476923077	0.215384615	0	0.307692308	0.307692308
x_bmm_k	21	0.753846154	0.246153846	0	0	0
x_bmm_k	22	0.723076923	0.276923077	0	0	0
x_bmm_k	23	0.415384615	0.261538462	0.323076923	0	0.323076923
x_bmm_k	24	0.430769231	0.276923077	0.292307692	0	0.292307692
x_bmm_k	25	0.415384615	0.276923077	0	0.307692308	0.307692308
x_bmm_k	26	0.753846154	0.246153846	0	0	0
x_bmm_k	27	0.738461538	0.261538462	0	0	0
x_bmm_k	28	0.676923077	0.323076923	0	0	0
x_bmm_k	29	0.723076923	0.276923077	0	0	0
x_bmm_k	30	0.753846154	0.246153846	0	0	0
x_bmm_k	31	0.476923077	0.230769231	0.292307692	0	0.292307692
x_bmm_k	32	0.384615385	0.323076923	0.292307692	0	0.292307692

};
\addlegendentry{6-bit}

\addplot[czredl, draw=czredl, fill=czredl] table
[x=layers, y=freq7, col sep=space] {
tensor_name	layers	freq4	freq5	freq6	freq7	high
x_bmm_k	1	0.815384615	0.184615385	0	0	0
x_bmm_k	2	0.769230769	0.230769231	0	0	0
x_bmm_k	3	0.446153846	0.261538462	0	0.292307692	0.292307692
x_bmm_k	4	0.415384615	0.246153846	0	0.338461538	0.338461538
x_bmm_k	5	0.476923077	0.246153846	0	0.276923077	0.276923077
x_bmm_k	6	0.338461538	0.307692308	0	0.353846154	0.353846154
x_bmm_k	7	0.446153846	0.261538462	0.292307692	0	0.292307692
x_bmm_k	8	0.692307692	0.307692308	0	0	0
x_bmm_k	9	0.707692308	0.292307692	0	0	0
x_bmm_k	10	0.753846154	0.246153846	0	0	0
x_bmm_k	11	0.723076923	0.276923077	0	0	0
x_bmm_k	12	0.430769231	0.2	0	0.369230769	0.369230769
x_bmm_k	13	0.476923077	0.246153846	0	0.276923077	0.276923077
x_bmm_k	14	0.476923077	0.246153846	0.276923077	0	0.276923077
x_bmm_k	15	0.769230769	0.230769231	0	0	0
x_bmm_k	16	0.292307692	0.323076923	0	0.384615385	0.384615385
x_bmm_k	17	0.430769231	0.276923077	0	0.292307692	0.292307692
x_bmm_k	18	0.692307692	0.307692308	0	0	0
x_bmm_k	19	0.430769231	0.276923077	0.292307692	0	0.292307692
x_bmm_k	20	0.476923077	0.215384615	0	0.307692308	0.307692308
x_bmm_k	21	0.753846154	0.246153846	0	0	0
x_bmm_k	22	0.723076923	0.276923077	0	0	0
x_bmm_k	23	0.415384615	0.261538462	0.323076923	0	0.323076923
x_bmm_k	24	0.430769231	0.276923077	0.292307692	0	0.292307692
x_bmm_k	25	0.415384615	0.276923077	0	0.307692308	0.307692308
x_bmm_k	26	0.753846154	0.246153846	0	0	0
x_bmm_k	27	0.738461538	0.261538462	0	0	0
x_bmm_k	28	0.676923077	0.323076923	0	0	0
x_bmm_k	29	0.723076923	0.276923077	0	0	0
x_bmm_k	30	0.753846154	0.246153846	0	0	0
x_bmm_k	31	0.476923077	0.230769231	0.292307692	0	0.292307692
x_bmm_k	32	0.384615385	0.323076923	0.292307692	0	0.292307692
};
\addlegendentry{7-bit}
\end{axis}
\end{tikzpicture}
        \caption{$\bf{K_i}$ in Line 6, \Cref{alg:transformer}. }
    \end{subfigure}

    \begin{subfigure}[b]{0.49\textwidth}
        \begin{tikzpicture}
\pgfplotsset{every tick label/.append style={font=\footnotesize}}
\begin{axis}[
    width=\textwidth,
    ybar stacked, height=4cm,
    legend style={at={(1.0,1.0)},anchor=south east, draw=none},
    ymin=0, ymax=1,
    ytick={0,0.5,1},
    yticklabels={0,0.5, 1},
    xmin=0.3, xmax=32.7,
    ylabel={\footnotesize{Percentage of occurrence}},
    xlabel={\footnotesize{Layer ID}},
    legend style={font=\footnotesize},
    legend columns = 4,
    bar width=4,
    ]

\addplot[czbluel, draw=czbluel, fill=czbluel] table
[x=layers, y=freq4, col sep=space]
{
tensor_name	layers	freq4	freq5	freq6	freq7	high
x_bmm_a	1	0.692307692	0.307692308	0	0	0
x_bmm_a	2	0.707692308	0.292307692	0	0	0
x_bmm_a	3	0.692307692	0.307692308	0	0	0
x_bmm_a	4	0.369230769	0.215384615	0	0.415384615	0.415384615
x_bmm_a	5	0.230769231	0.153846154	0.307692308	0.307692308	0.615384615
x_bmm_a	6	0.338461538	0.107692308	0.276923077	0.276923077	0.553846154
x_bmm_a	7	0.676923077	0.323076923	0	0	0
x_bmm_a	8	0.415384615	0.261538462	0	0.323076923	0.323076923
x_bmm_a	9	0.738461538	0.261538462	0	0	0
x_bmm_a	10	0.661538462	0.338461538	0	0	0
x_bmm_a	11	0.461538462	0.246153846	0.292307692	0	0.292307692
x_bmm_a	12	0.369230769	0.353846154	0.276923077	0	0.276923077
x_bmm_a	13	0.553846154	0.169230769	0	0.276923077	0.276923077
x_bmm_a	14	0.538461538	0.184615385	0.276923077	0	0.276923077
x_bmm_a	15	0.769230769	0.230769231	0	0	0
x_bmm_a	16	0.323076923	0.369230769	0.307692308	0	0.307692308
x_bmm_a	17	0.907692308	0.092307692	0	0	0
x_bmm_a	18	0.630769231	0.369230769	0	0	0
x_bmm_a	19	0.707692308	0.292307692	0	0	0
x_bmm_a	20	0.169230769	0.2	0.307692308	0.323076923	0.630769231
x_bmm_a	21	0.630769231	0.369230769	0	0	0
x_bmm_a	22	0.476923077	0.2	0	0.323076923	0.323076923
x_bmm_a	23	0.523076923	0.169230769	0.307692308	0	0.307692308
x_bmm_a	24	0.784615385	0.215384615	0	0	0
x_bmm_a	25	0.369230769	0.246153846	0.384615385	0	0.384615385
x_bmm_a	26	0.446153846	0.261538462	0.292307692	0	0.292307692
x_bmm_a	27	0.615384615	0.384615385	0	0	0
x_bmm_a	28	0.369230769	0.276923077	0.353846154	0	0.353846154
x_bmm_a	29	0.523076923	0.184615385	0	0.292307692	0.292307692
x_bmm_a	30	0.707692308	0.292307692	0	0	0
x_bmm_a	31	0.707692308	0.292307692	0	0	0
x_bmm_a	32	0.415384615	0.276923077	0.307692308	0	0.307692308

};
\addlegendentry{$\leq$ 4-bit}

\addplot[czgreenl, draw=czgreenl, fill=czgreenl] table
[x=layers, y=freq5, col sep=space] {
tensor_name	layers	freq4	freq5	freq6	freq7	high

x_bmm_a	1	0.692307692	0.307692308	0	0	0
x_bmm_a	2	0.707692308	0.292307692	0	0	0
x_bmm_a	3	0.692307692	0.307692308	0	0	0
x_bmm_a	4	0.369230769	0.215384615	0	0.415384615	0.415384615
x_bmm_a	5	0.230769231	0.153846154	0.307692308	0.307692308	0.615384615
x_bmm_a	6	0.338461538	0.107692308	0.276923077	0.276923077	0.553846154
x_bmm_a	7	0.676923077	0.323076923	0	0	0
x_bmm_a	8	0.415384615	0.261538462	0	0.323076923	0.323076923
x_bmm_a	9	0.738461538	0.261538462	0	0	0
x_bmm_a	10	0.661538462	0.338461538	0	0	0
x_bmm_a	11	0.461538462	0.246153846	0.292307692	0	0.292307692
x_bmm_a	12	0.369230769	0.353846154	0.276923077	0	0.276923077
x_bmm_a	13	0.553846154	0.169230769	0	0.276923077	0.276923077
x_bmm_a	14	0.538461538	0.184615385	0.276923077	0	0.276923077
x_bmm_a	15	0.769230769	0.230769231	0	0	0
x_bmm_a	16	0.323076923	0.369230769	0.307692308	0	0.307692308
x_bmm_a	17	0.907692308	0.092307692	0	0	0
x_bmm_a	18	0.630769231	0.369230769	0	0	0
x_bmm_a	19	0.707692308	0.292307692	0	0	0
x_bmm_a	20	0.169230769	0.2	0.307692308	0.323076923	0.630769231
x_bmm_a	21	0.630769231	0.369230769	0	0	0
x_bmm_a	22	0.476923077	0.2	0	0.323076923	0.323076923
x_bmm_a	23	0.523076923	0.169230769	0.307692308	0	0.307692308
x_bmm_a	24	0.784615385	0.215384615	0	0	0
x_bmm_a	25	0.369230769	0.246153846	0.384615385	0	0.384615385
x_bmm_a	26	0.446153846	0.261538462	0.292307692	0	0.292307692
x_bmm_a	27	0.615384615	0.384615385	0	0	0
x_bmm_a	28	0.369230769	0.276923077	0.353846154	0	0.353846154
x_bmm_a	29	0.523076923	0.184615385	0	0.292307692	0.292307692
x_bmm_a	30	0.707692308	0.292307692	0	0	0
x_bmm_a	31	0.707692308	0.292307692	0	0	0
x_bmm_a	32	0.415384615	0.276923077	0.307692308	0	0.307692308

};
\addlegendentry{5-bit}

\addplot[czorangel, draw=czorangel, fill=czorangel] table
[x=layers, y=freq6, col sep=space] {
tensor_name	layers	freq4	freq5	freq6	freq7	high

x_bmm_a	1	0.692307692	0.307692308	0	0	0
x_bmm_a	2	0.707692308	0.292307692	0	0	0
x_bmm_a	3	0.692307692	0.307692308	0	0	0
x_bmm_a	4	0.369230769	0.215384615	0	0.415384615	0.415384615
x_bmm_a	5	0.230769231	0.153846154	0.307692308	0.307692308	0.615384615
x_bmm_a	6	0.338461538	0.107692308	0.276923077	0.276923077	0.553846154
x_bmm_a	7	0.676923077	0.323076923	0	0	0
x_bmm_a	8	0.415384615	0.261538462	0	0.323076923	0.323076923
x_bmm_a	9	0.738461538	0.261538462	0	0	0
x_bmm_a	10	0.661538462	0.338461538	0	0	0
x_bmm_a	11	0.461538462	0.246153846	0.292307692	0	0.292307692
x_bmm_a	12	0.369230769	0.353846154	0.276923077	0	0.276923077
x_bmm_a	13	0.553846154	0.169230769	0	0.276923077	0.276923077
x_bmm_a	14	0.538461538	0.184615385	0.276923077	0	0.276923077
x_bmm_a	15	0.769230769	0.230769231	0	0	0
x_bmm_a	16	0.323076923	0.369230769	0.307692308	0	0.307692308
x_bmm_a	17	0.907692308	0.092307692	0	0	0
x_bmm_a	18	0.630769231	0.369230769	0	0	0
x_bmm_a	19	0.707692308	0.292307692	0	0	0
x_bmm_a	20	0.169230769	0.2	0.307692308	0.323076923	0.630769231
x_bmm_a	21	0.630769231	0.369230769	0	0	0
x_bmm_a	22	0.476923077	0.2	0	0.323076923	0.323076923
x_bmm_a	23	0.523076923	0.169230769	0.307692308	0	0.307692308
x_bmm_a	24	0.784615385	0.215384615	0	0	0
x_bmm_a	25	0.369230769	0.246153846	0.384615385	0	0.384615385
x_bmm_a	26	0.446153846	0.261538462	0.292307692	0	0.292307692
x_bmm_a	27	0.615384615	0.384615385	0	0	0
x_bmm_a	28	0.369230769	0.276923077	0.353846154	0	0.353846154
x_bmm_a	29	0.523076923	0.184615385	0	0.292307692	0.292307692
x_bmm_a	30	0.707692308	0.292307692	0	0	0
x_bmm_a	31	0.707692308	0.292307692	0	0	0
x_bmm_a	32	0.415384615	0.276923077	0.307692308	0	0.307692308

};
\addlegendentry{6-bit}

\addplot[czredl, draw=czredl, fill=czredl] table
[x=layers, y=freq7, col sep=space] {
tensor_name	layers	freq4	freq5	freq6	freq7	high
x_bmm_a	1	0.692307692	0.307692308	0	0	0
x_bmm_a	2	0.707692308	0.292307692	0	0	0
x_bmm_a	3	0.692307692	0.307692308	0	0	0
x_bmm_a	4	0.369230769	0.215384615	0	0.415384615	0.415384615
x_bmm_a	5	0.230769231	0.153846154	0.307692308	0.307692308	0.615384615
x_bmm_a	6	0.338461538	0.107692308	0.276923077	0.276923077	0.553846154
x_bmm_a	7	0.676923077	0.323076923	0	0	0
x_bmm_a	8	0.415384615	0.261538462	0	0.323076923	0.323076923
x_bmm_a	9	0.738461538	0.261538462	0	0	0
x_bmm_a	10	0.661538462	0.338461538	0	0	0
x_bmm_a	11	0.461538462	0.246153846	0.292307692	0	0.292307692
x_bmm_a	12	0.369230769	0.353846154	0.276923077	0	0.276923077
x_bmm_a	13	0.553846154	0.169230769	0	0.276923077	0.276923077
x_bmm_a	14	0.538461538	0.184615385	0.276923077	0	0.276923077
x_bmm_a	15	0.769230769	0.230769231	0	0	0
x_bmm_a	16	0.323076923	0.369230769	0.307692308	0	0.307692308
x_bmm_a	17	0.907692308	0.092307692	0	0	0
x_bmm_a	18	0.630769231	0.369230769	0	0	0
x_bmm_a	19	0.707692308	0.292307692	0	0	0
x_bmm_a	20	0.169230769	0.2	0.307692308	0.323076923	0.630769231
x_bmm_a	21	0.630769231	0.369230769	0	0	0
x_bmm_a	22	0.476923077	0.2	0	0.323076923	0.323076923
x_bmm_a	23	0.523076923	0.169230769	0.307692308	0	0.307692308
x_bmm_a	24	0.784615385	0.215384615	0	0	0
x_bmm_a	25	0.369230769	0.246153846	0.384615385	0	0.384615385
x_bmm_a	26	0.446153846	0.261538462	0.292307692	0	0.292307692
x_bmm_a	27	0.615384615	0.384615385	0	0	0
x_bmm_a	28	0.369230769	0.276923077	0.353846154	0	0.353846154
x_bmm_a	29	0.523076923	0.184615385	0	0.292307692	0.292307692
x_bmm_a	30	0.707692308	0.292307692	0	0	0
x_bmm_a	31	0.707692308	0.292307692	0	0	0
x_bmm_a	32	0.415384615	0.276923077	0.307692308	0	0.307692308

};
\addlegendentry{7-bit}
\end{axis}
\end{tikzpicture}
        \caption{$\bf{A_i}$ in Line 8, \Cref{alg:transformer}. }
    \end{subfigure}

    \caption{The searched bit-width distribution of OPT-2.7B. Notably, some layers are frequently assigned relatively high precision, indicating these layers are less tolerant to quantisation.}

    \label{fig:appendix:bartchart_1}
\end{figure}
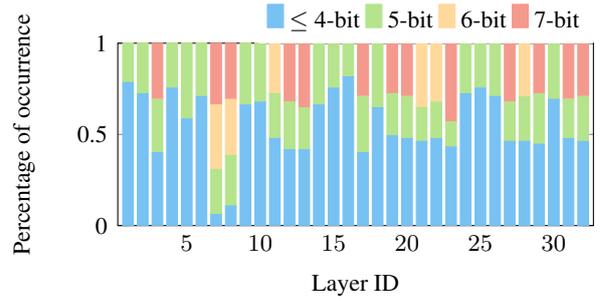
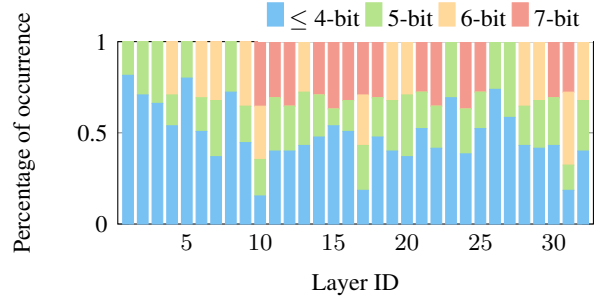
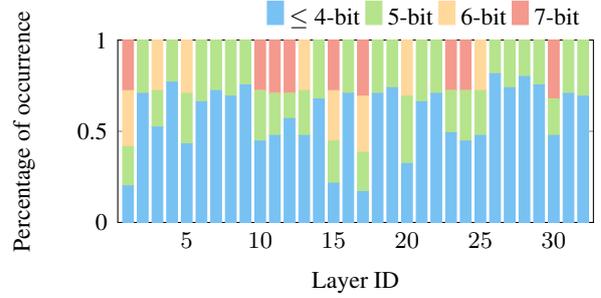
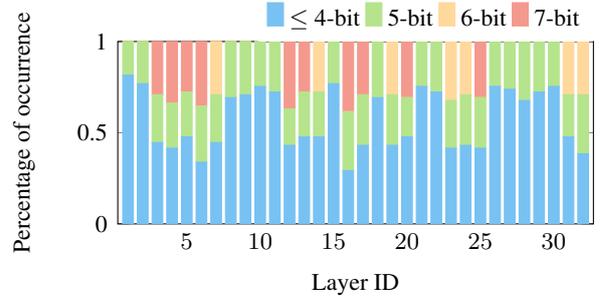
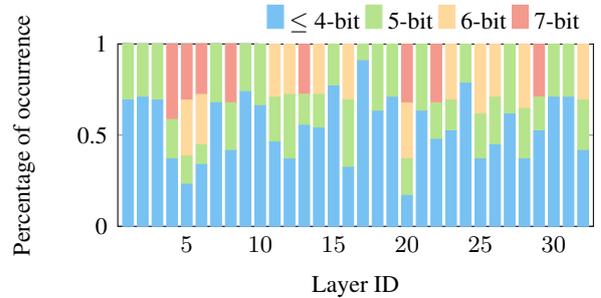

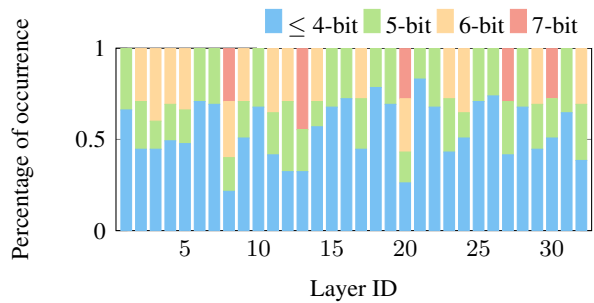
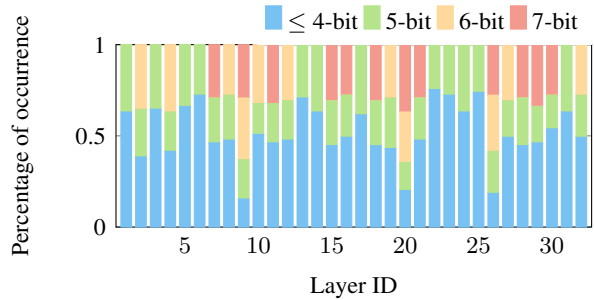
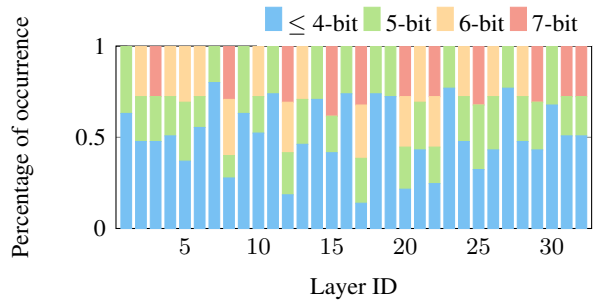
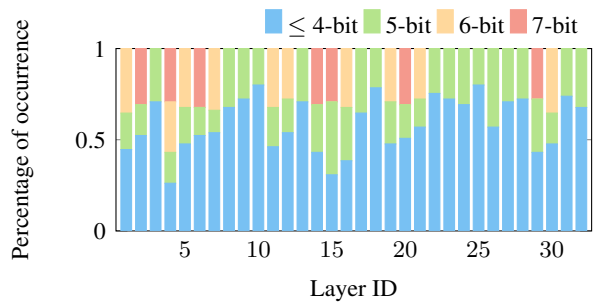
\begin{figure}[t]
    \centering

    \begin{subfigure}[b]{0.49\textwidth}
        \begin{tikzpicture}
\pgfplotsset{every tick label/.append style={font=\footnotesize}}
\begin{axis}[
    width=\textwidth,
    ybar stacked, height=4cm,
    legend style={at={(1.0,1.0)},anchor=south east, draw=none},
    ymin=0, ymax=1,
    ytick={0,0.5,1},
    yticklabels={0,0.5, 1},
    xmin=0.3, xmax=32.7,
    ylabel={\footnotesize{Percentage of occurrence}},
    xlabel={\footnotesize{Layer ID}},
    legend style={font=\footnotesize},
    legend columns = 4,
    bar width=4,
    ]

\addplot[czbluel, draw=czbluel, fill=czbluel] table
[x=layers, y=freq4, col sep=space]
{
tensor_name	layers	freq4	freq5	freq6	freq7	high
x_bmm_v	1	0.661538462	0.338461538	0	0	0
x_bmm_v	2	0.446153846	0.261538462	0.292307692	0	0.292307692
x_bmm_v	3	0.446153846	0.153846154	0.4	0	0.4
x_bmm_v	4	0.492307692	0.2	0.307692308	0	0.307692308
x_bmm_v	5	0.476923077	0.184615385	0.338461538	0	0.338461538
x_bmm_v	6	0.707692308	0.292307692	0	0	0
x_bmm_v	7	0.692307692	0.307692308	0	0	0
x_bmm_v	8	0.215384615	0.184615385	0.307692308	0.292307692	0.6
x_bmm_v	9	0.507692308	0.2	0.292307692	0	0.292307692
x_bmm_v	10	0.676923077	0.323076923	0	0	0
x_bmm_v	11	0.415384615	0.230769231	0.353846154	0	0.353846154
x_bmm_v	12	0.323076923	0.384615385	0.292307692	0	0.292307692
x_bmm_v	13	0.323076923	0.230769231	0	0.446153846	0.446153846
x_bmm_v	14	0.569230769	0.138461538	0.292307692	0	0.292307692
x_bmm_v	15	0.676923077	0.323076923	0	0	0
x_bmm_v	16	0.723076923	0.276923077	0	0	0
x_bmm_v	17	0.446153846	0.276923077	0.276923077	0	0.276923077
x_bmm_v	18	0.784615385	0.215384615	0	0	0
x_bmm_v	19	0.692307692	0.307692308	0	0	0
x_bmm_v	20	0.261538462	0.169230769	0.292307692	0.276923077	0.569230769
x_bmm_v	21	0.830769231	0.169230769	0	0	0
x_bmm_v	22	0.676923077	0.323076923	0	0	0
x_bmm_v	23	0.430769231	0.292307692	0.276923077	0	0.276923077
x_bmm_v	24	0.507692308	0.138461538	0.353846154	0	0.353846154
x_bmm_v	25	0.707692308	0.292307692	0	0	0
x_bmm_v	26	0.738461538	0.261538462	0	0	0
x_bmm_v	27	0.415384615	0.292307692	0	0.292307692	0.292307692
x_bmm_v	28	0.676923077	0.323076923	0	0	0
x_bmm_v	29	0.446153846	0.246153846	0.307692308	0	0.307692308
x_bmm_v	30	0.507692308	0.215384615	0	0.276923077	0.276923077
x_bmm_v	31	0.646153846	0.353846154	0	0	0
x_bmm_v	32	0.384615385	0.307692308	0.307692308	0	0.307692308

};
\addlegendentry{$\leq$ 4-bit}

\addplot[czgreenl, draw=czgreenl, fill=czgreenl] table
[x=layers, y=freq5, col sep=space] {
tensor_name	layers	freq4	freq5	freq6	freq7	high
x_bmm_v	1	0.661538462	0.338461538	0	0	0
x_bmm_v	2	0.446153846	0.261538462	0.292307692	0	0.292307692
x_bmm_v	3	0.446153846	0.153846154	0.4	0	0.4
x_bmm_v	4	0.492307692	0.2	0.307692308	0	0.307692308
x_bmm_v	5	0.476923077	0.184615385	0.338461538	0	0.338461538
x_bmm_v	6	0.707692308	0.292307692	0	0	0
x_bmm_v	7	0.692307692	0.307692308	0	0	0
x_bmm_v	8	0.215384615	0.184615385	0.307692308	0.292307692	0.6
x_bmm_v	9	0.507692308	0.2	0.292307692	0	0.292307692
x_bmm_v	10	0.676923077	0.323076923	0	0	0
x_bmm_v	11	0.415384615	0.230769231	0.353846154	0	0.353846154
x_bmm_v	12	0.323076923	0.384615385	0.292307692	0	0.292307692
x_bmm_v	13	0.323076923	0.230769231	0	0.446153846	0.446153846
x_bmm_v	14	0.569230769	0.138461538	0.292307692	0	0.292307692
x_bmm_v	15	0.676923077	0.323076923	0	0	0
x_bmm_v	16	0.723076923	0.276923077	0	0	0
x_bmm_v	17	0.446153846	0.276923077	0.276923077	0	0.276923077
x_bmm_v	18	0.784615385	0.215384615	0	0	0
x_bmm_v	19	0.692307692	0.307692308	0	0	0
x_bmm_v	20	0.261538462	0.169230769	0.292307692	0.276923077	0.569230769
x_bmm_v	21	0.830769231	0.169230769	0	0	0
x_bmm_v	22	0.676923077	0.323076923	0	0	0
x_bmm_v	23	0.430769231	0.292307692	0.276923077	0	0.276923077
x_bmm_v	24	0.507692308	0.138461538	0.353846154	0	0.353846154
x_bmm_v	25	0.707692308	0.292307692	0	0	0
x_bmm_v	26	0.738461538	0.261538462	0	0	0
x_bmm_v	27	0.415384615	0.292307692	0	0.292307692	0.292307692
x_bmm_v	28	0.676923077	0.323076923	0	0	0
x_bmm_v	29	0.446153846	0.246153846	0.307692308	0	0.307692308
x_bmm_v	30	0.507692308	0.215384615	0	0.276923077	0.276923077
x_bmm_v	31	0.646153846	0.353846154	0	0	0
x_bmm_v	32	0.384615385	0.307692308	0.307692308	0	0.307692308

};
\addlegendentry{5-bit}

\addplot[czorangel, draw=czorangel, fill=czorangel] table
[x=layers, y=freq6, col sep=space] {
tensor_name	layers	freq4	freq5	freq6	freq7	high
x_bmm_v	1	0.661538462	0.338461538	0	0	0
x_bmm_v	2	0.446153846	0.261538462	0.292307692	0	0.292307692
x_bmm_v	3	0.446153846	0.153846154	0.4	0	0.4
x_bmm_v	4	0.492307692	0.2	0.307692308	0	0.307692308
x_bmm_v	5	0.476923077	0.184615385	0.338461538	0	0.338461538
x_bmm_v	6	0.707692308	0.292307692	0	0	0
x_bmm_v	7	0.692307692	0.307692308	0	0	0
x_bmm_v	8	0.215384615	0.184615385	0.307692308	0.292307692	0.6
x_bmm_v	9	0.507692308	0.2	0.292307692	0	0.292307692
x_bmm_v	10	0.676923077	0.323076923	0	0	0
x_bmm_v	11	0.415384615	0.230769231	0.353846154	0	0.353846154
x_bmm_v	12	0.323076923	0.384615385	0.292307692	0	0.292307692
x_bmm_v	13	0.323076923	0.230769231	0	0.446153846	0.446153846
x_bmm_v	14	0.569230769	0.138461538	0.292307692	0	0.292307692
x_bmm_v	15	0.676923077	0.323076923	0	0	0
x_bmm_v	16	0.723076923	0.276923077	0	0	0
x_bmm_v	17	0.446153846	0.276923077	0.276923077	0	0.276923077
x_bmm_v	18	0.784615385	0.215384615	0	0	0
x_bmm_v	19	0.692307692	0.307692308	0	0	0
x_bmm_v	20	0.261538462	0.169230769	0.292307692	0.276923077	0.569230769
x_bmm_v	21	0.830769231	0.169230769	0	0	0
x_bmm_v	22	0.676923077	0.323076923	0	0	0
x_bmm_v	23	0.430769231	0.292307692	0.276923077	0	0.276923077
x_bmm_v	24	0.507692308	0.138461538	0.353846154	0	0.353846154
x_bmm_v	25	0.707692308	0.292307692	0	0	0
x_bmm_v	26	0.738461538	0.261538462	0	0	0
x_bmm_v	27	0.415384615	0.292307692	0	0.292307692	0.292307692
x_bmm_v	28	0.676923077	0.323076923	0	0	0
x_bmm_v	29	0.446153846	0.246153846	0.307692308	0	0.307692308
x_bmm_v	30	0.507692308	0.215384615	0	0.276923077	0.276923077
x_bmm_v	31	0.646153846	0.353846154	0	0	0
x_bmm_v	32	0.384615385	0.307692308	0.307692308	0	0.307692308

};
\addlegendentry{6-bit}

\addplot[czredl, draw=czredl, fill=czredl] table
[x=layers, y=freq7, col sep=space] {
tensor_name	layers	freq4	freq5	freq6	freq7	high
x_bmm_v	1	0.661538462	0.338461538	0	0	0
x_bmm_v	2	0.446153846	0.261538462	0.292307692	0	0.292307692
x_bmm_v	3	0.446153846	0.153846154	0.4	0	0.4
x_bmm_v	4	0.492307692	0.2	0.307692308	0	0.307692308
x_bmm_v	5	0.476923077	0.184615385	0.338461538	0	0.338461538
x_bmm_v	6	0.707692308	0.292307692	0	0	0
x_bmm_v	7	0.692307692	0.307692308	0	0	0
x_bmm_v	8	0.215384615	0.184615385	0.307692308	0.292307692	0.6
x_bmm_v	9	0.507692308	0.2	0.292307692	0	0.292307692
x_bmm_v	10	0.676923077	0.323076923	0	0	0
x_bmm_v	11	0.415384615	0.230769231	0.353846154	0	0.353846154
x_bmm_v	12	0.323076923	0.384615385	0.292307692	0	0.292307692
x_bmm_v	13	0.323076923	0.230769231	0	0.446153846	0.446153846
x_bmm_v	14	0.569230769	0.138461538	0.292307692	0	0.292307692
x_bmm_v	15	0.676923077	0.323076923	0	0	0
x_bmm_v	16	0.723076923	0.276923077	0	0	0
x_bmm_v	17	0.446153846	0.276923077	0.276923077	0	0.276923077
x_bmm_v	18	0.784615385	0.215384615	0	0	0
x_bmm_v	19	0.692307692	0.307692308	0	0	0
x_bmm_v	20	0.261538462	0.169230769	0.292307692	0.276923077	0.569230769
x_bmm_v	21	0.830769231	0.169230769	0	0	0
x_bmm_v	22	0.676923077	0.323076923	0	0	0
x_bmm_v	23	0.430769231	0.292307692	0.276923077	0	0.276923077
x_bmm_v	24	0.507692308	0.138461538	0.353846154	0	0.353846154
x_bmm_v	25	0.707692308	0.292307692	0	0	0
x_bmm_v	26	0.738461538	0.261538462	0	0	0
x_bmm_v	27	0.415384615	0.292307692	0	0.292307692	0.292307692
x_bmm_v	28	0.676923077	0.323076923	0	0	0
x_bmm_v	29	0.446153846	0.246153846	0.307692308	0	0.307692308
x_bmm_v	30	0.507692308	0.215384615	0	0.276923077	0.276923077
x_bmm_v	31	0.646153846	0.353846154	0	0	0
x_bmm_v	32	0.384615385	0.307692308	0.307692308	0	0.307692308

};
\addlegendentry{7-bit}
\end{axis}
\end{tikzpicture}
        \caption{$\bf{V_i}$ in Line 8, \Cref{alg:transformer}. }
    \end{subfigure}
    \begin{subfigure}[b]{0.49\textwidth}
        \begin{tikzpicture}
\pgfplotsset{every tick label/.append style={font=\footnotesize}}
\begin{axis}[
    width=\textwidth,
    ybar stacked, height=4cm,
    legend style={at={(1.0,1.0)},anchor=south east, draw=none},
    ymin=0, ymax=1,
    ytick={0,0.5,1},
    yticklabels={0,0.5, 1},
    xmin=0.3, xmax=32.7,
    ylabel={\footnotesize{Percentage of occurrence}},
    xlabel={\footnotesize{Layer ID}},
    legend style={font=\footnotesize},
    legend columns = 4,
    bar width=4,
    ]

\addplot[czbluel, draw=czbluel, fill=czbluel] table
[x=layers, y=freq4, col sep=space]
{
tensor_name	layers	freq4	freq5	freq6	freq7	high
x_out_proj	1	0.630769231	0.369230769	0	0	0
x_out_proj	2	0.384615385	0.261538462	0.353846154	0	0.353846154
x_out_proj	3	0.646153846	0.353846154	0	0	0
x_out_proj	4	0.415384615	0.215384615	0.369230769	0	0.369230769
x_out_proj	5	0.661538462	0.338461538	0	0	0
x_out_proj	6	0.723076923	0.276923077	0	0	0
x_out_proj	7	0.461538462	0.246153846	0	0.292307692	0.292307692
x_out_proj	8	0.476923077	0.246153846	0.276923077	0	0.276923077
x_out_proj	9	0.153846154	0.215384615	0.338461538	0.292307692	0.630769231
x_out_proj	10	0.507692308	0.169230769	0.323076923	0	0.323076923
x_out_proj	11	0.461538462	0.215384615	0	0.323076923	0.323076923
x_out_proj	12	0.476923077	0.215384615	0.307692308	0	0.307692308
x_out_proj	13	0.707692308	0.292307692	0	0	0
x_out_proj	14	0.630769231	0.369230769	0	0	0
x_out_proj	15	0.446153846	0.246153846	0	0.307692308	0.307692308
x_out_proj	16	0.492307692	0.230769231	0	0.276923077	0.276923077
x_out_proj	17	0.615384615	0.384615385	0	0	0
x_out_proj	18	0.446153846	0.246153846	0	0.307692308	0.307692308
x_out_proj	19	0.430769231	0.276923077	0.292307692	0	0.292307692
x_out_proj	20	0.2	0.153846154	0.276923077	0.369230769	0.646153846
x_out_proj	21	0.476923077	0.230769231	0	0.292307692	0.292307692
x_out_proj	22	0.753846154	0.246153846	0	0	0
x_out_proj	23	0.723076923	0.276923077	0	0	0
x_out_proj	24	0.630769231	0.369230769	0	0	0
x_out_proj	25	0.738461538	0.261538462	0	0	0
x_out_proj	26	0.184615385	0.230769231	0.307692308	0.276923077	0.584615385
x_out_proj	27	0.492307692	0.2	0.307692308	0	0.307692308
x_out_proj	28	0.446153846	0.261538462	0	0.292307692	0.292307692
x_out_proj	29	0.461538462	0.2	0	0.338461538	0.338461538
x_out_proj	30	0.538461538	0.184615385	0	0.276923077	0.276923077
x_out_proj	31	0.630769231	0.369230769	0	0	0
x_out_proj	32	0.492307692	0.230769231	0.276923077	0	0.276923077

};
\addlegendentry{$\leq$ 4-bit}

\addplot[czgreenl, draw=czgreenl, fill=czgreenl] table
[x=layers, y=freq5, col sep=space] {
tensor_name	layers	freq4	freq5	freq6	freq7	high
x_out_proj	1	0.630769231	0.369230769	0	0	0
x_out_proj	2	0.384615385	0.261538462	0.353846154	0	0.353846154
x_out_proj	3	0.646153846	0.353846154	0	0	0
x_out_proj	4	0.415384615	0.215384615	0.369230769	0	0.369230769
x_out_proj	5	0.661538462	0.338461538	0	0	0
x_out_proj	6	0.723076923	0.276923077	0	0	0
x_out_proj	7	0.461538462	0.246153846	0	0.292307692	0.292307692
x_out_proj	8	0.476923077	0.246153846	0.276923077	0	0.276923077
x_out_proj	9	0.153846154	0.215384615	0.338461538	0.292307692	0.630769231
x_out_proj	10	0.507692308	0.169230769	0.323076923	0	0.323076923
x_out_proj	11	0.461538462	0.215384615	0	0.323076923	0.323076923
x_out_proj	12	0.476923077	0.215384615	0.307692308	0	0.307692308
x_out_proj	13	0.707692308	0.292307692	0	0	0
x_out_proj	14	0.630769231	0.369230769	0	0	0
x_out_proj	15	0.446153846	0.246153846	0	0.307692308	0.307692308
x_out_proj	16	0.492307692	0.230769231	0	0.276923077	0.276923077
x_out_proj	17	0.615384615	0.384615385	0	0	0
x_out_proj	18	0.446153846	0.246153846	0	0.307692308	0.307692308
x_out_proj	19	0.430769231	0.276923077	0.292307692	0	0.292307692
x_out_proj	20	0.2	0.153846154	0.276923077	0.369230769	0.646153846
x_out_proj	21	0.476923077	0.230769231	0	0.292307692	0.292307692
x_out_proj	22	0.753846154	0.246153846	0	0	0
x_out_proj	23	0.723076923	0.276923077	0	0	0
x_out_proj	24	0.630769231	0.369230769	0	0	0
x_out_proj	25	0.738461538	0.261538462	0	0	0
x_out_proj	26	0.184615385	0.230769231	0.307692308	0.276923077	0.584615385
x_out_proj	27	0.492307692	0.2	0.307692308	0	0.307692308
x_out_proj	28	0.446153846	0.261538462	0	0.292307692	0.292307692
x_out_proj	29	0.461538462	0.2	0	0.338461538	0.338461538
x_out_proj	30	0.538461538	0.184615385	0	0.276923077	0.276923077
x_out_proj	31	0.630769231	0.369230769	0	0	0
x_out_proj	32	0.492307692	0.230769231	0.276923077	0	0.276923077

};
\addlegendentry{5-bit}

\addplot[czorangel, draw=czorangel, fill=czorangel] table
[x=layers, y=freq6, col sep=space] {
tensor_name	layers	freq4	freq5	freq6	freq7	high
x_out_proj	1	0.630769231	0.369230769	0	0	0
x_out_proj	2	0.384615385	0.261538462	0.353846154	0	0.353846154
x_out_proj	3	0.646153846	0.353846154	0	0	0
x_out_proj	4	0.415384615	0.215384615	0.369230769	0	0.369230769
x_out_proj	5	0.661538462	0.338461538	0	0	0
x_out_proj	6	0.723076923	0.276923077	0	0	0
x_out_proj	7	0.461538462	0.246153846	0	0.292307692	0.292307692
x_out_proj	8	0.476923077	0.246153846	0.276923077	0	0.276923077
x_out_proj	9	0.153846154	0.215384615	0.338461538	0.292307692	0.630769231
x_out_proj	10	0.507692308	0.169230769	0.323076923	0	0.323076923
x_out_proj	11	0.461538462	0.215384615	0	0.323076923	0.323076923
x_out_proj	12	0.476923077	0.215384615	0.307692308	0	0.307692308
x_out_proj	13	0.707692308	0.292307692	0	0	0
x_out_proj	14	0.630769231	0.369230769	0	0	0
x_out_proj	15	0.446153846	0.246153846	0	0.307692308	0.307692308
x_out_proj	16	0.492307692	0.230769231	0	0.276923077	0.276923077
x_out_proj	17	0.615384615	0.384615385	0	0	0
x_out_proj	18	0.446153846	0.246153846	0	0.307692308	0.307692308
x_out_proj	19	0.430769231	0.276923077	0.292307692	0	0.292307692
x_out_proj	20	0.2	0.153846154	0.276923077	0.369230769	0.646153846
x_out_proj	21	0.476923077	0.230769231	0	0.292307692	0.292307692
x_out_proj	22	0.753846154	0.246153846	0	0	0
x_out_proj	23	0.723076923	0.276923077	0	0	0
x_out_proj	24	0.630769231	0.369230769	0	0	0
x_out_proj	25	0.738461538	0.261538462	0	0	0
x_out_proj	26	0.184615385	0.230769231	0.307692308	0.276923077	0.584615385
x_out_proj	27	0.492307692	0.2	0.307692308	0	0.307692308
x_out_proj	28	0.446153846	0.261538462	0	0.292307692	0.292307692
x_out_proj	29	0.461538462	0.2	0	0.338461538	0.338461538
x_out_proj	30	0.538461538	0.184615385	0	0.276923077	0.276923077
x_out_proj	31	0.630769231	0.369230769	0	0	0
x_out_proj	32	0.492307692	0.230769231	0.276923077	0	0.276923077

};
\addlegendentry{6-bit}

\addplot[czredl, draw=czredl, fill=czredl] table
[x=layers, y=freq7, col sep=space] {
tensor_name	layers	freq4	freq5	freq6	freq7	high
x_out_proj	1	0.630769231	0.369230769	0	0	0
x_out_proj	2	0.384615385	0.261538462	0.353846154	0	0.353846154
x_out_proj	3	0.646153846	0.353846154	0	0	0
x_out_proj	4	0.415384615	0.215384615	0.369230769	0	0.369230769
x_out_proj	5	0.661538462	0.338461538	0	0	0
x_out_proj	6	0.723076923	0.276923077	0	0	0
x_out_proj	7	0.461538462	0.246153846	0	0.292307692	0.292307692
x_out_proj	8	0.476923077	0.246153846	0.276923077	0	0.276923077
x_out_proj	9	0.153846154	0.215384615	0.338461538	0.292307692	0.630769231
x_out_proj	10	0.507692308	0.169230769	0.323076923	0	0.323076923
x_out_proj	11	0.461538462	0.215384615	0	0.323076923	0.323076923
x_out_proj	12	0.476923077	0.215384615	0.307692308	0	0.307692308
x_out_proj	13	0.707692308	0.292307692	0	0	0
x_out_proj	14	0.630769231	0.369230769	0	0	0
x_out_proj	15	0.446153846	0.246153846	0	0.307692308	0.307692308
x_out_proj	16	0.492307692	0.230769231	0	0.276923077	0.276923077
x_out_proj	17	0.615384615	0.384615385	0	0	0
x_out_proj	18	0.446153846	0.246153846	0	0.307692308	0.307692308
x_out_proj	19	0.430769231	0.276923077	0.292307692	0	0.292307692
x_out_proj	20	0.2	0.153846154	0.276923077	0.369230769	0.646153846
x_out_proj	21	0.476923077	0.230769231	0	0.292307692	0.292307692
x_out_proj	22	0.753846154	0.246153846	0	0	0
x_out_proj	23	0.723076923	0.276923077	0	0	0
x_out_proj	24	0.630769231	0.369230769	0	0	0
x_out_proj	25	0.738461538	0.261538462	0	0	0
x_out_proj	26	0.184615385	0.230769231	0.307692308	0.276923077	0.584615385
x_out_proj	27	0.492307692	0.2	0.307692308	0	0.307692308
x_out_proj	28	0.446153846	0.261538462	0	0.292307692	0.292307692
x_out_proj	29	0.461538462	0.2	0	0.338461538	0.338461538
x_out_proj	30	0.538461538	0.184615385	0	0.276923077	0.276923077
x_out_proj	31	0.630769231	0.369230769	0	0	0
x_out_proj	32	0.492307692	0.230769231	0.276923077	0	0.276923077

};
\addlegendentry{7-bit}
\end{axis}
\end{tikzpicture}
        \caption{$\bf{B_c}$ in Line 11, \Cref{alg:transformer}. }
    \end{subfigure}
    \begin{subfigure}[b]{0.49\textwidth}
        \begin{tikzpicture}
\pgfplotsset{every tick label/.append style={font=\footnotesize}}
\begin{axis}[
    width=\textwidth,
    ybar stacked, height=4cm,
    legend style={at={(1.0,1.0)},anchor=south east, draw=none},
    ymin=0, ymax=1,
    ytick={0,0.5,1},
    yticklabels={0,0.5, 1},
    xmin=0.3, xmax=32.7,
    ylabel={\footnotesize{Percentage of occurrence}},
    xlabel={\footnotesize{Layer ID}},
    legend style={font=\footnotesize},
    legend columns = 4,
    bar width=4,
    ]

\addplot[czbluel, draw=czbluel, fill=czbluel] table
[x=layers, y=freq4, col sep=space]
{
tensor_name	layers	freq4	freq5	freq6	freq7	high
x_fc1	1	0.630769231	0.369230769	0	0	0
x_fc1	2	0.476923077	0.246153846	0.276923077	0	0.276923077
x_fc1	3	0.476923077	0.246153846	0	0.276923077	0.276923077
x_fc1	4	0.507692308	0.215384615	0.276923077	0	0.276923077
x_fc1	5	0.369230769	0.323076923	0.307692308	0	0.307692308
x_fc1	6	0.553846154	0.169230769	0.276923077	0	0.276923077
x_fc1	7	0.8	0.2	0	0	0
x_fc1	8	0.276923077	0.123076923	0.307692308	0.292307692	0.6
x_fc1	9	0.630769231	0.369230769	0	0	0
x_fc1	10	0.523076923	0.2	0.276923077	0	0.276923077
x_fc1	11	0.738461538	0.261538462	0	0	0
x_fc1	12	0.184615385	0.230769231	0.276923077	0.307692308	0.584615385
x_fc1	13	0.461538462	0.246153846	0.292307692	0	0.292307692
x_fc1	14	0.707692308	0.292307692	0	0	0
x_fc1	15	0.415384615	0.2	0	0.384615385	0.384615385
x_fc1	16	0.738461538	0.261538462	0	0	0
x_fc1	17	0.138461538	0.246153846	0.292307692	0.323076923	0.615384615
x_fc1	18	0.738461538	0.261538462	0	0	0
x_fc1	19	0.723076923	0.276923077	0	0	0
x_fc1	20	0.215384615	0.230769231	0.276923077	0.276923077	0.553846154
x_fc1	21	0.430769231	0.261538462	0.307692308	0	0.307692308
x_fc1	22	0.246153846	0.2	0.276923077	0.276923077	0.553846154
x_fc1	23	0.769230769	0.230769231	0	0	0
x_fc1	24	0.476923077	0.246153846	0.276923077	0	0.276923077
x_fc1	25	0.323076923	0.353846154	0	0.323076923	0.323076923
x_fc1	26	0.430769231	0.292307692	0.276923077	0	0.276923077
x_fc1	27	0.769230769	0.230769231	0	0	0
x_fc1	28	0.476923077	0.246153846	0.276923077	0	0.276923077
x_fc1	29	0.430769231	0.261538462	0	0.307692308	0.307692308
x_fc1	30	0.676923077	0.323076923	0	0	0
x_fc1	31	0.507692308	0.215384615	0	0.276923077	0.276923077
x_fc1	32	0.507692308	0.215384615	0	0.276923077	0.276923077

};
\addlegendentry{$\leq$ 4-bit}

\addplot[czgreenl, draw=czgreenl, fill=czgreenl] table
[x=layers, y=freq5, col sep=space] {
tensor_name	layers	freq4	freq5	freq6	freq7	high
x_fc1	1	0.630769231	0.369230769	0	0	0
x_fc1	2	0.476923077	0.246153846	0.276923077	0	0.276923077
x_fc1	3	0.476923077	0.246153846	0	0.276923077	0.276923077
x_fc1	4	0.507692308	0.215384615	0.276923077	0	0.276923077
x_fc1	5	0.369230769	0.323076923	0.307692308	0	0.307692308
x_fc1	6	0.553846154	0.169230769	0.276923077	0	0.276923077
x_fc1	7	0.8	0.2	0	0	0
x_fc1	8	0.276923077	0.123076923	0.307692308	0.292307692	0.6
x_fc1	9	0.630769231	0.369230769	0	0	0
x_fc1	10	0.523076923	0.2	0.276923077	0	0.276923077
x_fc1	11	0.738461538	0.261538462	0	0	0
x_fc1	12	0.184615385	0.230769231	0.276923077	0.307692308	0.584615385
x_fc1	13	0.461538462	0.246153846	0.292307692	0	0.292307692
x_fc1	14	0.707692308	0.292307692	0	0	0
x_fc1	15	0.415384615	0.2	0	0.384615385	0.384615385
x_fc1	16	0.738461538	0.261538462	0	0	0
x_fc1	17	0.138461538	0.246153846	0.292307692	0.323076923	0.615384615
x_fc1	18	0.738461538	0.261538462	0	0	0
x_fc1	19	0.723076923	0.276923077	0	0	0
x_fc1	20	0.215384615	0.230769231	0.276923077	0.276923077	0.553846154
x_fc1	21	0.430769231	0.261538462	0.307692308	0	0.307692308
x_fc1	22	0.246153846	0.2	0.276923077	0.276923077	0.553846154
x_fc1	23	0.769230769	0.230769231	0	0	0
x_fc1	24	0.476923077	0.246153846	0.276923077	0	0.276923077
x_fc1	25	0.323076923	0.353846154	0	0.323076923	0.323076923
x_fc1	26	0.430769231	0.292307692	0.276923077	0	0.276923077
x_fc1	27	0.769230769	0.230769231	0	0	0
x_fc1	28	0.476923077	0.246153846	0.276923077	0	0.276923077
x_fc1	29	0.430769231	0.261538462	0	0.307692308	0.307692308
x_fc1	30	0.676923077	0.323076923	0	0	0
x_fc1	31	0.507692308	0.215384615	0	0.276923077	0.276923077
x_fc1	32	0.507692308	0.215384615	0	0.276923077	0.276923077

};
\addlegendentry{5-bit}

\addplot[czorangel, draw=czorangel, fill=czorangel] table
[x=layers, y=freq6, col sep=space] {
tensor_name	layers	freq4	freq5	freq6	freq7	high
x_fc1	1	0.630769231	0.369230769	0	0	0
x_fc1	2	0.476923077	0.246153846	0.276923077	0	0.276923077
x_fc1	3	0.476923077	0.246153846	0	0.276923077	0.276923077
x_fc1	4	0.507692308	0.215384615	0.276923077	0	0.276923077
x_fc1	5	0.369230769	0.323076923	0.307692308	0	0.307692308
x_fc1	6	0.553846154	0.169230769	0.276923077	0	0.276923077
x_fc1	7	0.8	0.2	0	0	0
x_fc1	8	0.276923077	0.123076923	0.307692308	0.292307692	0.6
x_fc1	9	0.630769231	0.369230769	0	0	0
x_fc1	10	0.523076923	0.2	0.276923077	0	0.276923077
x_fc1	11	0.738461538	0.261538462	0	0	0
x_fc1	12	0.184615385	0.230769231	0.276923077	0.307692308	0.584615385
x_fc1	13	0.461538462	0.246153846	0.292307692	0	0.292307692
x_fc1	14	0.707692308	0.292307692	0	0	0
x_fc1	15	0.415384615	0.2	0	0.384615385	0.384615385
x_fc1	16	0.738461538	0.261538462	0	0	0
x_fc1	17	0.138461538	0.246153846	0.292307692	0.323076923	0.615384615
x_fc1	18	0.738461538	0.261538462	0	0	0
x_fc1	19	0.723076923	0.276923077	0	0	0
x_fc1	20	0.215384615	0.230769231	0.276923077	0.276923077	0.553846154
x_fc1	21	0.430769231	0.261538462	0.307692308	0	0.307692308
x_fc1	22	0.246153846	0.2	0.276923077	0.276923077	0.553846154
x_fc1	23	0.769230769	0.230769231	0	0	0
x_fc1	24	0.476923077	0.246153846	0.276923077	0	0.276923077
x_fc1	25	0.323076923	0.353846154	0	0.323076923	0.323076923
x_fc1	26	0.430769231	0.292307692	0.276923077	0	0.276923077
x_fc1	27	0.769230769	0.230769231	0	0	0
x_fc1	28	0.476923077	0.246153846	0.276923077	0	0.276923077
x_fc1	29	0.430769231	0.261538462	0	0.307692308	0.307692308
x_fc1	30	0.676923077	0.323076923	0	0	0
x_fc1	31	0.507692308	0.215384615	0	0.276923077	0.276923077
x_fc1	32	0.507692308	0.215384615	0	0.276923077	0.276923077

};
\addlegendentry{6-bit}

\addplot[czredl, draw=czredl, fill=czredl] table
[x=layers, y=freq7, col sep=space] {
tensor_name	layers	freq4	freq5	freq6	freq7	high
x_fc1	1	0.630769231	0.369230769	0	0	0
x_fc1	2	0.476923077	0.246153846	0.276923077	0	0.276923077
x_fc1	3	0.476923077	0.246153846	0	0.276923077	0.276923077
x_fc1	4	0.507692308	0.215384615	0.276923077	0	0.276923077
x_fc1	5	0.369230769	0.323076923	0.307692308	0	0.307692308
x_fc1	6	0.553846154	0.169230769	0.276923077	0	0.276923077
x_fc1	7	0.8	0.2	0	0	0
x_fc1	8	0.276923077	0.123076923	0.307692308	0.292307692	0.6
x_fc1	9	0.630769231	0.369230769	0	0	0
x_fc1	10	0.523076923	0.2	0.276923077	0	0.276923077
x_fc1	11	0.738461538	0.261538462	0	0	0
x_fc1	12	0.184615385	0.230769231	0.276923077	0.307692308	0.584615385
x_fc1	13	0.461538462	0.246153846	0.292307692	0	0.292307692
x_fc1	14	0.707692308	0.292307692	0	0	0
x_fc1	15	0.415384615	0.2	0	0.384615385	0.384615385
x_fc1	16	0.738461538	0.261538462	0	0	0
x_fc1	17	0.138461538	0.246153846	0.292307692	0.323076923	0.615384615
x_fc1	18	0.738461538	0.261538462	0	0	0
x_fc1	19	0.723076923	0.276923077	0	0	0
x_fc1	20	0.215384615	0.230769231	0.276923077	0.276923077	0.553846154
x_fc1	21	0.430769231	0.261538462	0.307692308	0	0.307692308
x_fc1	22	0.246153846	0.2	0.276923077	0.276923077	0.553846154
x_fc1	23	0.769230769	0.230769231	0	0	0
x_fc1	24	0.476923077	0.246153846	0.276923077	0	0.276923077
x_fc1	25	0.323076923	0.353846154	0	0.323076923	0.323076923
x_fc1	26	0.430769231	0.292307692	0.276923077	0	0.276923077
x_fc1	27	0.769230769	0.230769231	0	0	0
x_fc1	28	0.476923077	0.246153846	0.276923077	0	0.276923077
x_fc1	29	0.430769231	0.261538462	0	0.307692308	0.307692308
x_fc1	30	0.676923077	0.323076923	0	0	0
x_fc1	31	0.507692308	0.215384615	0	0.276923077	0.276923077
x_fc1	32	0.507692308	0.215384615	0	0.276923077	0.276923077

};
\addlegendentry{7-bit}
\end{axis}
\end{tikzpicture}
        \caption{$\bf{B_n}$ in Line 13, \Cref{alg:transformer}. }
    \end{subfigure}

    \begin{subfigure}[b]{0.49\textwidth}
        \begin{tikzpicture}
\pgfplotsset{every tick label/.append style={font=\footnotesize}}
\begin{axis}[
    width=\textwidth,
    ybar stacked, height=4cm,
    legend style={at={(1.0,1.0)},anchor=south east, draw=none},
    ymin=0, ymax=1,
    ytick={0,0.5,1},
    yticklabels={0,0.5, 1},
    xmin=0.3, xmax=32.7,
    ylabel={\footnotesize{Percentage of occurrence}},
    xlabel={\footnotesize{Layer ID}},
    legend style={font=\footnotesize},
    legend columns = 4,
    bar width=4,
    ]

\addplot[czbluel, draw=czbluel, fill=czbluel] table
[x=layers, y=freq4, col sep=space]
{
tensor_name	layers	freq4	freq5	freq6	freq7	high
x_fc2	1	0.446153846	0.2	0.353846154	0	0.353846154
x_fc2	2	0.523076923	0.169230769	0	0.307692308	0.307692308
x_fc2	3	0.707692308	0.292307692	0	0	0
x_fc2	4	0.261538462	0.169230769	0.276923077	0.292307692	0.569230769
x_fc2	5	0.476923077	0.2	0.323076923	0	0.323076923
x_fc2	6	0.523076923	0.153846154	0	0.323076923	0.323076923
x_fc2	7	0.538461538	0.123076923	0.338461538	0	0.338461538
x_fc2	8	0.676923077	0.323076923	0	0	0
x_fc2	9	0.723076923	0.276923077	0	0	0
x_fc2	10	0.8	0.2	0	0	0
x_fc2	11	0.461538462	0.215384615	0.323076923	0	0.323076923
x_fc2	12	0.538461538	0.184615385	0.276923077	0	0.276923077
x_fc2	13	0.707692308	0.292307692	0	0	0
x_fc2	14	0.430769231	0.261538462	0	0.307692308	0.307692308
x_fc2	15	0.307692308	0.4	0	0.292307692	0.292307692
x_fc2	16	0.384615385	0.292307692	0.323076923	0	0.323076923
x_fc2	17	0.646153846	0.353846154	0	0	0
x_fc2	18	0.784615385	0.215384615	0	0	0
x_fc2	19	0.476923077	0.230769231	0.292307692	0	0.292307692
x_fc2	20	0.507692308	0.184615385	0	0.307692308	0.307692308
x_fc2	21	0.569230769	0.153846154	0.276923077	0	0.276923077
x_fc2	22	0.753846154	0.246153846	0	0	0
x_fc2	23	0.723076923	0.276923077	0	0	0
x_fc2	24	0.692307692	0.307692308	0	0	0
x_fc2	25	0.8	0.2	0	0	0
x_fc2	26	0.569230769	0.430769231	0	0	0
x_fc2	27	0.707692308	0.292307692	0	0	0
x_fc2	28	0.723076923	0.276923077	0	0	0
x_fc2	29	0.430769231	0.292307692	0	0.276923077	0.276923077
x_fc2	30	0.476923077	0.169230769	0.353846154	0	0.353846154
x_fc2	31	0.738461538	0.261538462	0	0	0
x_fc2	32	0.676923077	0.323076923	0	0	0

};
\addlegendentry{$\leq$ 4-bit}

\addplot[czgreenl, draw=czgreenl, fill=czgreenl] table
[x=layers, y=freq5, col sep=space] {
tensor_name	layers	freq4	freq5	freq6	freq7	high
x_fc2	1	0.446153846	0.2	0.353846154	0	0.353846154
x_fc2	2	0.523076923	0.169230769	0	0.307692308	0.307692308
x_fc2	3	0.707692308	0.292307692	0	0	0
x_fc2	4	0.261538462	0.169230769	0.276923077	0.292307692	0.569230769
x_fc2	5	0.476923077	0.2	0.323076923	0	0.323076923
x_fc2	6	0.523076923	0.153846154	0	0.323076923	0.323076923
x_fc2	7	0.538461538	0.123076923	0.338461538	0	0.338461538
x_fc2	8	0.676923077	0.323076923	0	0	0
x_fc2	9	0.723076923	0.276923077	0	0	0
x_fc2	10	0.8	0.2	0	0	0
x_fc2	11	0.461538462	0.215384615	0.323076923	0	0.323076923
x_fc2	12	0.538461538	0.184615385	0.276923077	0	0.276923077
x_fc2	13	0.707692308	0.292307692	0	0	0
x_fc2	14	0.430769231	0.261538462	0	0.307692308	0.307692308
x_fc2	15	0.307692308	0.4	0	0.292307692	0.292307692
x_fc2	16	0.384615385	0.292307692	0.323076923	0	0.323076923
x_fc2	17	0.646153846	0.353846154	0	0	0
x_fc2	18	0.784615385	0.215384615	0	0	0
x_fc2	19	0.476923077	0.230769231	0.292307692	0	0.292307692
x_fc2	20	0.507692308	0.184615385	0	0.307692308	0.307692308
x_fc2	21	0.569230769	0.153846154	0.276923077	0	0.276923077
x_fc2	22	0.753846154	0.246153846	0	0	0
x_fc2	23	0.723076923	0.276923077	0	0	0
x_fc2	24	0.692307692	0.307692308	0	0	0
x_fc2	25	0.8	0.2	0	0	0
x_fc2	26	0.569230769	0.430769231	0	0	0
x_fc2	27	0.707692308	0.292307692	0	0	0
x_fc2	28	0.723076923	0.276923077	0	0	0
x_fc2	29	0.430769231	0.292307692	0	0.276923077	0.276923077
x_fc2	30	0.476923077	0.169230769	0.353846154	0	0.353846154
x_fc2	31	0.738461538	0.261538462	0	0	0
x_fc2	32	0.676923077	0.323076923	0	0	0

};
\addlegendentry{5-bit}

\addplot[czorangel, draw=czorangel, fill=czorangel] table
[x=layers, y=freq6, col sep=space] {
tensor_name	layers	freq4	freq5	freq6	freq7	high
x_fc2	1	0.446153846	0.2	0.353846154	0	0.353846154
x_fc2	2	0.523076923	0.169230769	0	0.307692308	0.307692308
x_fc2	3	0.707692308	0.292307692	0	0	0
x_fc2	4	0.261538462	0.169230769	0.276923077	0.292307692	0.569230769
x_fc2	5	0.476923077	0.2	0.323076923	0	0.323076923
x_fc2	6	0.523076923	0.153846154	0	0.323076923	0.323076923
x_fc2	7	0.538461538	0.123076923	0.338461538	0	0.338461538
x_fc2	8	0.676923077	0.323076923	0	0	0
x_fc2	9	0.723076923	0.276923077	0	0	0
x_fc2	10	0.8	0.2	0	0	0
x_fc2	11	0.461538462	0.215384615	0.323076923	0	0.323076923
x_fc2	12	0.538461538	0.184615385	0.276923077	0	0.276923077
x_fc2	13	0.707692308	0.292307692	0	0	0
x_fc2	14	0.430769231	0.261538462	0	0.307692308	0.307692308
x_fc2	15	0.307692308	0.4	0	0.292307692	0.292307692
x_fc2	16	0.384615385	0.292307692	0.323076923	0	0.323076923
x_fc2	17	0.646153846	0.353846154	0	0	0
x_fc2	18	0.784615385	0.215384615	0	0	0
x_fc2	19	0.476923077	0.230769231	0.292307692	0	0.292307692
x_fc2	20	0.507692308	0.184615385	0	0.307692308	0.307692308
x_fc2	21	0.569230769	0.153846154	0.276923077	0	0.276923077
x_fc2	22	0.753846154	0.246153846	0	0	0
x_fc2	23	0.723076923	0.276923077	0	0	0
x_fc2	24	0.692307692	0.307692308	0	0	0
x_fc2	25	0.8	0.2	0	0	0
x_fc2	26	0.569230769	0.430769231	0	0	0
x_fc2	27	0.707692308	0.292307692	0	0	0
x_fc2	28	0.723076923	0.276923077	0	0	0
x_fc2	29	0.430769231	0.292307692	0	0.276923077	0.276923077
x_fc2	30	0.476923077	0.169230769	0.353846154	0	0.353846154
x_fc2	31	0.738461538	0.261538462	0	0	0
x_fc2	32	0.676923077	0.323076923	0	0	0

};
\addlegendentry{6-bit}

\addplot[czredl, draw=czredl, fill=czredl] table
[x=layers, y=freq7, col sep=space] {
tensor_name	layers	freq4	freq5	freq6	freq7	high
x_fc2	1	0.446153846	0.2	0.353846154	0	0.353846154
x_fc2	2	0.523076923	0.169230769	0	0.307692308	0.307692308
x_fc2	3	0.707692308	0.292307692	0	0	0
x_fc2	4	0.261538462	0.169230769	0.276923077	0.292307692	0.569230769
x_fc2	5	0.476923077	0.2	0.323076923	0	0.323076923
x_fc2	6	0.523076923	0.153846154	0	0.323076923	0.323076923
x_fc2	7	0.538461538	0.123076923	0.338461538	0	0.338461538
x_fc2	8	0.676923077	0.323076923	0	0	0
x_fc2	9	0.723076923	0.276923077	0	0	0
x_fc2	10	0.8	0.2	0	0	0
x_fc2	11	0.461538462	0.215384615	0.323076923	0	0.323076923
x_fc2	12	0.538461538	0.184615385	0.276923077	0	0.276923077
x_fc2	13	0.707692308	0.292307692	0	0	0
x_fc2	14	0.430769231	0.261538462	0	0.307692308	0.307692308
x_fc2	15	0.307692308	0.4	0	0.292307692	0.292307692
x_fc2	16	0.384615385	0.292307692	0.323076923	0	0.323076923
x_fc2	17	0.646153846	0.353846154	0	0	0
x_fc2	18	0.784615385	0.215384615	0	0	0
x_fc2	19	0.476923077	0.230769231	0.292307692	0	0.292307692
x_fc2	20	0.507692308	0.184615385	0	0.307692308	0.307692308
x_fc2	21	0.569230769	0.153846154	0.276923077	0	0.276923077
x_fc2	22	0.753846154	0.246153846	0	0	0
x_fc2	23	0.723076923	0.276923077	0	0	0
x_fc2	24	0.692307692	0.307692308	0	0	0
x_fc2	25	0.8	0.2	0	0	0
x_fc2	26	0.569230769	0.430769231	0	0	0
x_fc2	27	0.707692308	0.292307692	0	0	0
x_fc2	28	0.723076923	0.276923077	0	0	0
x_fc2	29	0.430769231	0.292307692	0	0.276923077	0.276923077
x_fc2	30	0.476923077	0.169230769	0.353846154	0	0.353846154
x_fc2	31	0.738461538	0.261538462	0	0	0
x_fc2	32	0.676923077	0.323076923	0	0	0

};
\addlegendentry{7-bit}
\end{axis}
\end{tikzpicture}
        \caption{$\bf{B_1}$ in Line 14, \Cref{alg:transformer}. }
    \end{subfigure}
    \caption{The searched bit-width distribution of OPT-2.7B. Notably, some layers are frequently assigned relatively high precision, indicating these layers are less tolerant to quantisation.}
    \label{fig:appendix:bartchart_2}
\end{figure}

Mixed-precision quantisation is also helpful for recovering downstream task accuracy. \Cref{fig:appendix:search_on_lambada} and \Cref{fig:appendix_search_on_arc_easy} depict the performance of 4-bit LLMs on LAMBADA and ARC (easy) as the model scales up. The searched mixed-precision configuration effectively recovers the accuracy. \Cref{fig:appendix:bartchart_1} and~\ref{fig:appendix:bartchart_2} is the activation distribution after searching on LAMBADA 2688 times. Keeping these layers in high precision effectively recovers the accuracy from 36.2\% to 61.3\% without decreasing the memory density, equivalent to a 4.3-bit OPT-2.7B.

\label{appendix:mixed_precision_with_dse}

\begin{figure}[t]
    \centering
    \subcaptionbox{LAMBADA\label{fig:appendix:search_on_lambada}}{
\begin{tikzpicture}
\begin{axis}[
    height=55mm,
    width=75mm,
    grid=both,
    xlabel={Model size},
    ylabel={Mean accuracy (\%)},
    ytick={0.25, 0.35, 0.45, 0.55, 0.65, 0.75},
    yticklabels={25, 35, 45, 55, 65, 75},
    xtick={1, 2, 3, 4, 5},
    xticklabels={125M, 350M, 1.3B, 2.7B, 6.7B},
    legend columns = 3,
    y label style={at={(axis description cs:0.00,1)}, anchor=north east},
    legend style={at={(0.5,1.05)},anchor=south, draw=none, fill=none, font=\footnotesize},
    ]
\addplot[dashed, czred, draw=czred, mark=*, line width=1pt, mark options={solid, scale=1.0}] table
[x=model_idx, y=fp32, col sep=space] {
    model_idx	fp32
    1	0.379002523
    2	0.451581603
    3	0.579274209
    4	0.635940229
    5	0.677081312
};
\addlegendentry{FP32}

\addplot[dashed, czblue, draw=czblue, mark=star, line width=1pt, mark options={solid}] table
[x=model_idx, y=uniform] {
    model_idx	uniform	  mixed	  fp32
    1	0.2866	0.336114885	0.379002523
    2	0.3866	0.459344071	0.451581603
    3	0.4137	0.555621968	0.579274209
    4	0.3621	0.613	0.635940229
    5	0.5603	0.65189	0.677081312
};
\addlegendentry{4-bit uniform}

\addplot[dashed, czorange, draw=czorange, mark=triangle*, line width=1pt, mark options={solid}] table
[x=model_idx, y=mixed, col sep=space] {
    model_idx	uniform	  mixed	  fp32
    1	0.2866	0.336114885	0.379002523
    2	0.3866	0.459344071	0.451581603
    3	0.4137	0.555621968	0.579274209
    4	0.3621	0.613	0.635940229
    5	0.5603	0.65189	0.677081312

};
\addlegendentry{{4-bit mixed}}

\end{axis}
\end{tikzpicture}
    }
    \subcaptionbox{ARC (easy)\label{fig:appendix_search_on_arc_easy}}{
\begin{tikzpicture}
\begin{axis}[
    height=55mm,
    width=75mm,
    grid=both,
    xlabel={Model size},
    ylabel={Mean accuracy (\%)},
    xtick={1, 2, 3, 4, 5},
    xticklabels={125M, 350M, 1.3B, 2.7B, 6.7B},
    legend columns = 3,
    y label style={at={(axis description cs:0.00,1)}, anchor=north east},
    legend style={at={(0.5,1.05)},anchor=south, draw=none, fill=none, font=\footnotesize},
    ]
\addplot[dashed, czred, draw=czred, mark=*, line width=1pt, mark options={solid, scale=1.0}] table
[x=model_idx, y=fp32, col sep=space] {
model_idx	mixed	fp32	uniform
1	40.5	43.5	37.0
2	42.5	44.0	39.9
3	56.1	57.0	50.0
4	58.7	60.8	52.4
5	64.9	65.6	61.5
};
\addlegendentry{FP32}

\addplot[dashed, czblue, draw=czblue, mark=star, line width=1pt, mark options={solid}] table
[x=model_idx, y=uniform] {
model_idx	mixed	fp32	uniform
1	40.5	43.5	37.0
2	42.5	44.0	39.9
3	56.1	57.0	50.0
4	58.7	60.8	52.4
5	64.9	65.6	61.5
};
\addlegendentry{4-bit uniform}

\addplot[dashed, czorange, draw=czorange, mark=triangle*, line width=1pt, mark options={solid}] table
[x=model_idx, y=mixed, col sep=space] {
model_idx	mixed	fp32	uniform
1	40.5	43.5	37.0
2	42.5	44.0	39.9
3	56.1	57.0	50.0
4	58.7	60.8	52.4
5	64.9	65.6	61.5
};
\addlegendentry{{4-bit mixed}}

\end{axis}
\end{tikzpicture}
    }
    \caption{The accuracy of FP32 model, 4-bit uniform BFP, and 4-bit mixed-precision on LAMBADA and ARCH (easy) in the zero-shot prompting context. Searched mixed-precision quantisation configuration captures the distribution inherent in LLMs, effectively recovering the accuracy. }
    \label{fig:appendix:mixed_precision_vs_uniform}
\end{figure}

\section{Mixed-precision with hardware model}

\begin{figure}[t]
    \centering
        \begin{tikzpicture}[thick,scale=0.95, every node/.style={scale=0.95}]
        \pgfplotsset{every x tick label/.append style={font=\footnotesize}, compat=1.3}
        \begin{semilogyaxis}[
            height=0.45\textwidth,
            width=0.50\textwidth,
            grid=both,
            xlabel={Time (s)},
            ylabel={$O_f$},
            xmin=20,
            xtick={20, 100, 500, 1000, 3600},
            ymode=log, log ticks with fixed point, xmode=log,
            legend cell align={left},
            legend style={at={(0.25,0.25)},anchor=north west, draw=none, fill=none},
            ]
        \addplot[czred, draw=czred, mark=*, line width=1pt, mark options={scale=1.0}] table
        [x=t, y=c, col sep=space] {alg_dse.dat};
        \addlegendentry{\small{hardware-aware}}

        \addplot[czblue, draw=czblue, mark=*, line width=1pt, mark options={scale=1.0}] table
        [x=t, y=c, col sep=space] {alg_base.dat};
        \addlegendentry{\small{software-only}}
        \end{semilogyaxis}
        \end{tikzpicture}
    \caption{A comparison between the new hardware-aware search algorithm and the previous one that only considers accuracy and memory density. We introduce two additional hardware metrics, Token Per Second (TPS) and TPS Per LUT (TPS/LUT), such that the search algorithm can be aware of the hardware efficiency. The new objective function is $O_f = acc + \alpha_1\cdot mem + \alpha_2 \cdot tps + \alpha_3 \cdot tpl$, where $tps$ and $tpl$ denote TPS and TPS/LUT respectively. We plot the traces of both searches versus search time for BERT-base. We observe that the hardware-aware curve is higher than the original one, proving our new objective function guides the search for better hardware efficiency in terms of speedup and circuit area.}
    \label{fig:appendix:bert_search_dse}
\end{figure}
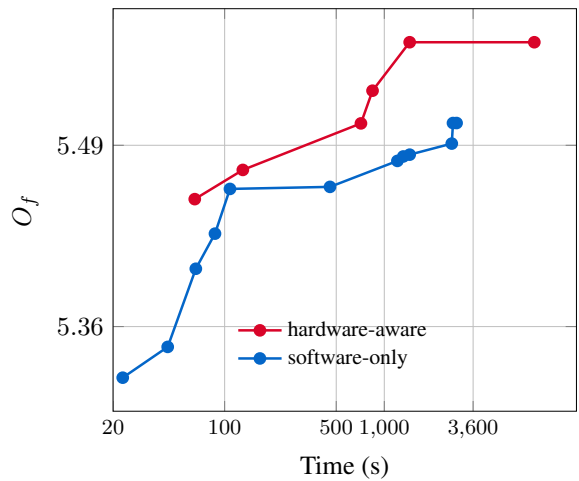

{We additionally performed a hardware-aware quantization search on BERT-Base~\cite{devlin2018bert}. We implemented the actual BFP hardware on FPGAs via high-level synthesis and model the hardware cost using Token per second (TPS) for speedup and TPS per LUT (TPS/LUT) for circuit area efficiency. Our hardware-aware search takes accuracy, memory bitwidth, and hardware cost as feedback. We compare the search traces of hardware-aware search and software-only search in~\Cref{fig:appendix:bert_search_dse}, and observe that the hardware-aware search curve is higher than the original one, proving our new objective function guides the search for better hardware efficiency. We will scale up experiments in future work.}

\end{document}